\documentclass{article}

\usepackage{PRIMEarxiv}

\usepackage[utf8]{inputenc} % allow utf-8 input
\usepackage[T1]{fontenc}    % use 8-bit T1 fonts
\usepackage{hyperref}       % hyperlinks
\usepackage{url}            % simple URL typesetting
\usepackage{booktabs}       % professional-quality tables
\usepackage{amsfonts}       % blackboard math symbols
\usepackage{nicefrac}       % compact symbols for 1/2, etc.
\usepackage{microtype}      % microtypography
\usepackage{lipsum}
\usepackage{fancyhdr}       % header
\usepackage{graphicx}       % graphics
\graphicspath{{media/}}     % organize your images and other figures under media/ folder
\usepackage{booktabs}       % professional-quality tables
\usepackage{tabularx}       % adjustable column width tables 
\usepackage{multirow}
\usepackage{adjustbox}
\usepackage{array}
\usepackage{booktabs}  % Add this line for improved table formatting
\usepackage{amsmath}
\usepackage{caption}
\usepackage{subcaption}
\usepackage{graphicx}
\usepackage{cleveref}
\usepackage{placeins}

\usepackage{amsfonts}    
\usepackage{colortbl}
\usepackage{makecell}
\usepackage{pifont}
\usepackage{caption}
\usepackage{longtable}
\usepackage{changepage}
\usepackage{xcolor}
\usepackage{listings} 
\usepackage{mdframed}

\definecolor{lightblue}{RGB}{206, 242, 245}
\definecolor{lightblue2}{RGB}{206, 245, 230}
\definecolor{lightblue3}{RGB}{234, 234, 251}
\definecolor{newgreen}{HTML}{dbf2d0}
\definecolor{newpurple}{HTML}{d8c2dc}
\definecolor{newblue}{HTML}{a6c9ec}
\definecolor{newpink}{HTML}{f2dadd}

\title{\texorpdfstring{$\mu$-Bench:}  A Vision-Language Benchmark for Microscopy Understanding}

\author{%
  \textbf{Alejandro Lozano} * \\
  Department of Biomedical Data Science\\
  Stanford University\\
  Stanford, CA 94305 \\
  \texttt{lozanoe@stanford.edu} \\
  \And
  \textbf{Jeffrey Nirschl} * \\
  Department of Pathology\\
  Stanford University\\
  Stanford, CA 94305 \\
  \texttt{jnirschl@stanford.edu} \\
    \And
  James Burgess \\
 ICME \\
  Stanford University\\
  Stanford, CA 94305 \\
  \texttt{jmhb@stanford.edu} \\
  \And
  Sanket Rajan Gupte \\
 Department of Computer Science \\
  Stanford University\\
  Stanford, CA 94305 \\
  \texttt{sanketg@stanford.edu} \\
  \And
   Yuhui Zhang \\
 Department of Computer Science \\
  Stanford University\\
  Stanford, CA 94305 \\
  \texttt{yuhuiz@stanford.edu} \\
    \And
   Alyssa Unell \\
 Department of Computer Science \\
  Stanford University\\
  Stanford, CA 94305 \\
  \texttt{aunell@stanford.edu} \\
  \And
  Serena Yeung-Levy \\
  Department of Biomedical Data Science\\
  Stanford University\\
  Stanford, CA 94305 \\
  \texttt{syyeung@stanford.edu} \\
}

\newcommand{\GPT}[2]{\cellcolor{newgreen} GPT-4o & \cellcolor{newgreen!50} #1 ($\pm$ #2)}
\newcommand{\COGVLM}[2]{\cellcolor{newgreen} CogVLM & \cellcolor{newgreen!50} #1 ($\pm$ #2)}
\newcommand{\QWENVLM}[2]{\cellcolor{newgreen} QwenVLM & \cellcolor{newgreen!50} #1 ($\pm$ #2)}
\newcommand{\PALIGEMMA}[2]{\cellcolor{newgreen} PaliGemma & \cellcolor{newgreen!50} #1 ($\pm$ #2)}

\newcommand{\ALIGN}[2]{\cellcolor{newblue} ALIGN & \cellcolor{newblue!50} #1 ($\pm$ #2)}
\newcommand{\OPENCLIP}[2]{\cellcolor{newblue} OpenCLIP & \cellcolor{newblue!50} #1 ($\pm$ #2)}
\newcommand{\CLIP}[2]{\cellcolor{newblue} CLIP & \cellcolor{newblue!50} #1 ($\pm$ #2)}

\newcommand{\PLIP}[2]{\cellcolor{newpink} PLIP & \cellcolor{newpink!50} #1 ($\pm$ #2)}
\newcommand{\CONCH}[2]{\cellcolor{newpink} CONCH & \cellcolor{newpink!50} #1 ($\pm$ #2)}
\newcommand{\QUILTCLIP}[2]{\cellcolor{newpink} QuiltNet & \cellcolor{newpink!50} #1 ($\pm$ #2)}

\newcommand{\RQUILTCLIP}[2]{\cellcolor{newpink} M-QuiltNet* & \cellcolor{newpink!50} #1 ($\pm$ #2)}

\newcommand{\RPLIP}[2]{\cellcolor{newpink} M-PLIP* & \cellcolor{newpink!50} #1 ($\pm$ #2)}

\newcommand{\BIOMEDCLIP}[2]{\cellcolor{newpurple} BiomedCLIP & \cellcolor{newpurple!50} #1 ($\pm$ #2)}

\newcommand{\RANDOM}[2]{\cellcolor{gray!35} Random & \cellcolor{gray!15} #1 ($\pm$ #2)}
\newcommand{\dataset}{$\mu$-Bench}

\newcommand{\numtasks}{22}
\newcommand{\numdatasets}{24}

\newcommand{\cognitionimages}{54}
\newcommand{\perceptionuniquecellmicro}{91}

\newcommand{\numsubmodalities}{8}

\newcommand{\numstaining}{24}

\newcommand{\cognitionquestions}{121}
\newcommand{\cognitionusers}{6}
\newcommand{\cognitioninstitutions}{5}

\begin{document}
\maketitle

\begin{abstract}
Recent advances in microscopy have enabled the rapid generation of terabytes of image data in cell biology and biomedical research. Vision-language models (VLMs) offer a promising solution for large-scale biological image analysis, enhancing researchers’ efficiency, identifying new image biomarkers, and accelerating hypothesis generation and scientific discovery. However, there is a lack of standardized, diverse, and large-scale vision-language benchmarks to evaluate VLMs’ perception and cognition capabilities in biological image understanding. To address this gap, we introduce \dataset, an expert-curated benchmark encompassing \numtasks\ biomedical tasks across various scientific disciplines (biology, pathology), microscopy modalities (electron, fluorescence, light), scales (subcellular, cellular, tissue), and organisms in both normal and abnormal states. We evaluate state-of-the-art biomedical, pathology, and general VLMs on \dataset\ and find that: 
i) current models struggle on all categories, even for basic tasks such as distinguishing microscopy modalities; 
ii) current specialist models fine-tuned on biomedical data often perform worse than generalist models;
iii) fine-tuning in specific microscopy domains can cause catastrophic forgetting, eroding prior biomedical knowledge encoded in their base model. % what abot before further training 
iv) weight interpolation between fine-tuned and pre-trained models offers one solution to forgetting and improves general performance across biomedical tasks.
We release \dataset\ under a permissive license to accelerate the research and development of microscopy foundation models.
\end{abstract}

%%%%%%%%%%%%%%%%%%%%%%%%%%%%%%%%%%%
%%%%%%%%%% Introduction %%%%%%%%%%%
%%%%%%%%%%%%%%%%%%%%%%%%%%%%%%%%%%%
\section{Introduction}
\begin{figure}
    \centering
    \includegraphics[width=1\linewidth]{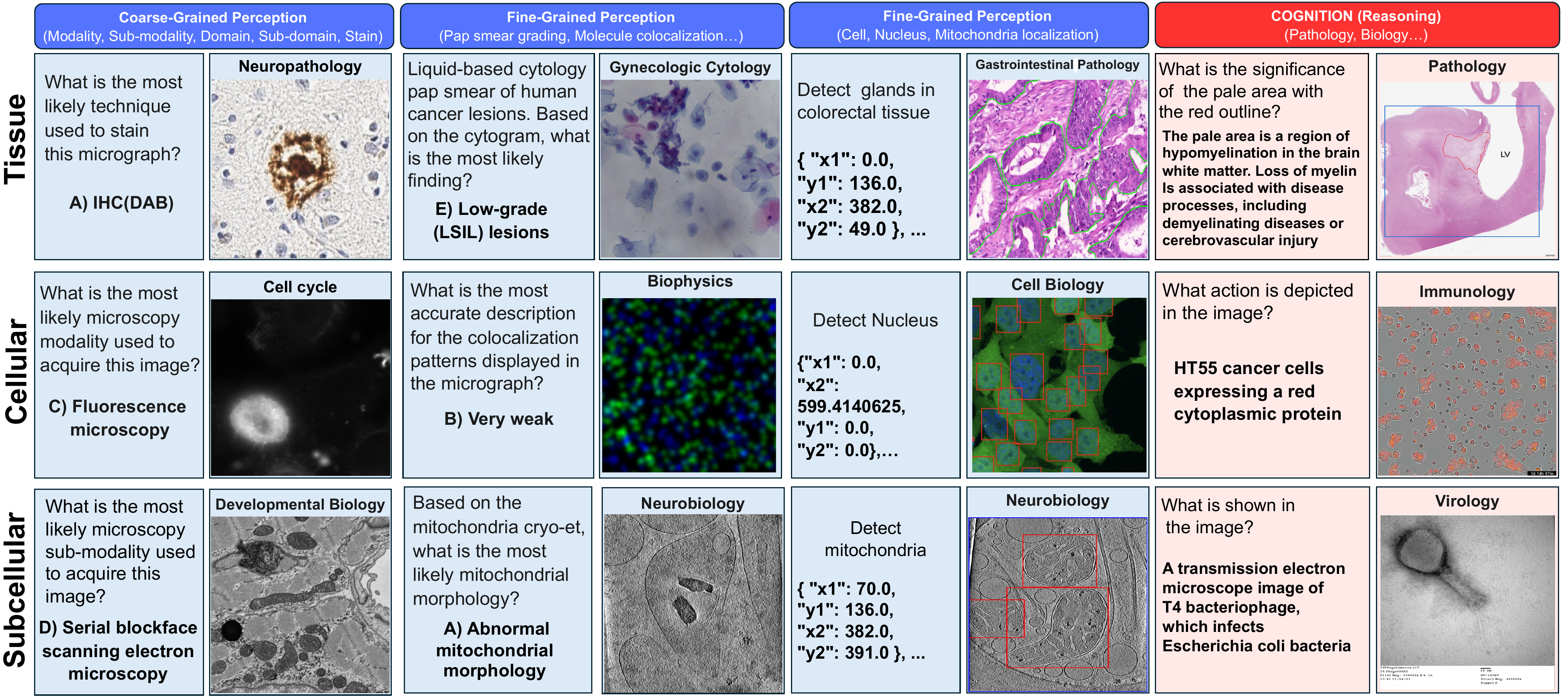}
    \caption{Data samples from \dataset, covering perception (\textcolor{blue}{left}) and cognition  (\textcolor{red}{right}) tasks across subcellular, cellular, and tissue levels tasks across electron, fluorescence, and light microscopy. 
    %\SY{please see my big comment in the data section, and this part should be made sure to be consistent. the content from my big comment should also be referenced in the intro, you don't talk about these different tasks so fig 1 becomes a bit disconnected. ideally make text a little bit bigger, and be clear of the output format for each task (could be in the caption if no space in the figure}
    }
    \label{fig:enter-label}
\end{figure}

Microscopy is a cornerstone of biomedical research \cite{merkle2013ascent}, enabling detailed study of biological structures at multiple scales \cite{weber2021multidisciplinarity}. Advances in cryo-electron microscopy, high-throughput fluorescence microscopy, and whole-slide imaging allow the rapid generation of terabytes of image data, which are essential for fields such as cell biology, biomedical research, and pathology \cite{zhang2023biomedclip}. These data span multiple length scales, allowing researchers to examine atomic/molecular, subcellular/cellular, and cell/tissue-level structures with high precision  \cite{foss2002end}. A crucial first step in microscopy analysis is interpreting and reasoning about the significance of image findings \cite{tu2024towards}. However, this requires domain expertise and comprehensive knowledge of biology, normal/abnormal states, and the capabilities and limitations of microscopy techniques. The increased volume and velocity of microscopy data compound the challenge of manual microscopy image interpretation. %Human-interactive AI systems could augment researchers to make them more efficient by synthesizing findings in large, suggesting next steps, and generally facilitating expert human review.

Text is an intuitive interface for interactive analysis, and thus, vision-language models (VLMs) are one promising approach to assist with image interpretation. Auto-regressive VLMs can follow instructions and respond to questions using free text, which is intuitive for users without a computational background. A biomedical VLM  that has distilled knowledge from diverse microscopy images with input from multiple domain experts could also democratize image interpretation for specialized techniques. In addition, they could relate image findings to recent literature, relevant genes, small molecule therapeutics, and diseases and potentially accelerate hypothesis generation and discovery \cite{gao2024empowering,wang2023scientific,schwartz2023scaling,malacrida2023phasor,nogare2023using,carpenter2023smart,li2023challenges,li2023challenges,lambert2023towards, hein2023global}.

Foundation VLMs trained on large corpora of image-text data have revolutionized many areas of artificial intelligence (AI), excelling in a variety of natural language processing (NLP) \cite{brown2020language,vaswani2017attention}, computer-vision (CV) \cite{kirillov2023segment, caron2021emerging}, and vision-language \cite{radford2021learning,alayrac2022flamingo,bai2023qwen} tasks. 
% A key component to the success of these models is the large-scale datasets of natural images and texts sourced from the internet\cite{schuhmann2022laion}. 
It is well known that biomedical data are significantly under-represented in internet-based image datasets, and a critical first question is how well generalist VLMs trained on natural image-text data perform on out-of-domain biomedical vision-language tasks.
% \JB{motivation talks about generalist VLMs but not specialist biomedical VLMs}

However, there are no diverse, large-scale vision-language benchmarks to evaluate generalist or specialist VLMs in microscopy image interpretation. Whereas segmentation benchmarks are common\cite{ma2024multimodality}, and recent efforts have curated impressive image-drug phenotype multimodal datasets\cite{Chandrasekaran2024-jump1}, the current state of microscopy vision-language benchmarks is limited. Existing microscopy vision-language benchmarks often focus on single-domain diagnostic capabilities (e.g., histopathology \cite{he2020pathvqa, Huang2023-plip}) rather than describing and understanding \cite{he2020pathvqa}. Although these are significant contributions, they lack the diversity of images and content needed to evaluate the performance of VLMS across tasks, scales, microscopy modalities, organisms, and biological processes. Lastly, in contrast to the general vision-language domain, current biomedical evaluations lack comprehensive characterization across both perception (recognizing specific objects, counting, localization, and color) and cognition abilities (integrating knowledge and perceptual to deduce more complex answers) \cite{fu2023mme}. This gap hinders the progress of developing robust VLMs tailored for biomedical research.

To address the need for robust VLMs tailored for biomedical research, we present two contributions:

\begin{enumerate}

\item  \textbf{\dataset\:} A high-quality vision-language benchmark of 17,235 microscopy images from a diverse collection of unpublished and published datasets with newly added expert annotations and a permissive license. \dataset\ spans  \numtasks\  perception and cognition tasks across light, fluorescence, and electron microscopy. It includes closed VQA, captioning, object detection, and segmentation tasks across \numsubmodalities\ microscopy sub-modalities and \numstaining\ staining techniques, representative of  12 scientific domains.

\item  \textbf{Characterization}  We leverage \dataset\ to characterize state-of-the-art generalist and domain-specific biomedical VLMs. We show several findings: 
even the best-performing VLMs have high error rates across all microscopy tasks (do not generalize well); 
specialist biomedical VLMs often underperform general VLMs; specialist fine-tuning in specific domains can cause catastrophic forgetting of biological knowledge that existed in the base model; and a simple weight ensembling strategy can mitigate the forgetting problem. 
\end{enumerate}

%%%%%%%%%%%%%%%%%%%%%%%%%%%%%%%%%%%
%%%%%%%%%% Related Work %%%%%%%%%%%
%%%%%%%%%%%%%%%%%%%%%%%%%%%%%%%%%%%

\section{Related Work}

\textbf{Biomedical Vision-Language Benchmarks.} While previous works have developed various biomedical vision-language benchmarks that have been instrumental in advancing diagnostic capabilities, they present three main problems: 1) Task simplicity: Most biomedical computer vision benchmarks predominantly focus on supervised tasks such as classification and segmentation \cite{caicedo2019nucleus, saltz2018spatial, antonelli2022medical} rather than vision-language tasks. As shown in Table \ref{tab:lit_rev}, biomedical vision-language benchmarks are less common. 2) Lack of diversity: existing vision-language datasets are usually limited to diagnostic imaging such as radiology or pathology \cite{rajpurkar2017chexnet, esteva2017dermatologist}; there is a lack of benchmarks for basic research microscopy. 3) Limited accessibility: PMC-15M is a large image-text dataset used to train BiomedCLIP \cite{zhang2023biomedclip}, which includes diverse microscopy images, domains, and organisms but is not publicly accessible for training or evaluation.

\textbf{Vision-Language Models in Biomedicine.} Vision-language models (VLMs) can be generally categorized into two types: 1) Contrastive models, such as CLIP \cite{radford2021learning} and ALIGN \cite{jia2021scaling}, which use contrastive learning to create shared image-text embeddings, facilitating tasks like zero-shot classification and text-image retrieval; and 2) Auto-regressive models, such as Flamingo \cite{alayrac2022flamingo} and GPT-4 \cite{achiam2023gpt}, which integrate image embeddings with large language models (LLMs) to perform zero-shot tasks, follow instructions, and reason about content. These VLMs have significant potential to advance biomedicine \cite{moor2023foundation, tu2024towards, gao2024empowering, wang2023scientific, m2024augmenting}.

However, these models are primarily trained on general datasets with limited biomedical coverage, leading to suboptimal performance on biomedical tasks \cite{schuhmann2022laion, zhao2023clip}. To address this gap, specialized VLMs have been developed by fine-tuning generalist models on biomedical data. Notable examples include BiomedCLIP \cite{zhang2023biomedclip} train on images from PubMed, and histopathology vision-language models such as PLIP \cite{huang2023visual} and CONCH \cite{yu2022coca}, which were trained on Twitter or various pathology sources. Despite the increase in vision-language foundation models for microscopy, there are no diverse benchmarks to evaluate the performance of image-based perception and reasoning about microscopy images across diverse scales and modalities. Our work on \dataset\ addresses this issue by providing a comprehensive benchmark that includes a variety of important and diverse biological processes, organisms, microscopy modalities, domains, and tasks to support the evaluation of microscopy foundation models.
%%%%%%%%%%%%%%%%%%%%%%%%%%%%%%%%%%%
%%%%%%%% Dataset Curation %%%%%%%%%
%%%%%%%%%%%%%%%%%%%%%%%%%%%%%%%%%%%

% # # # # # # # # # # # # # #
% [START] FIGURE 1: WORKFLOW
% # # # # # # # # # # # # # #

\begin{figure*}[!t]
\includegraphics[width=1.0\textwidth]{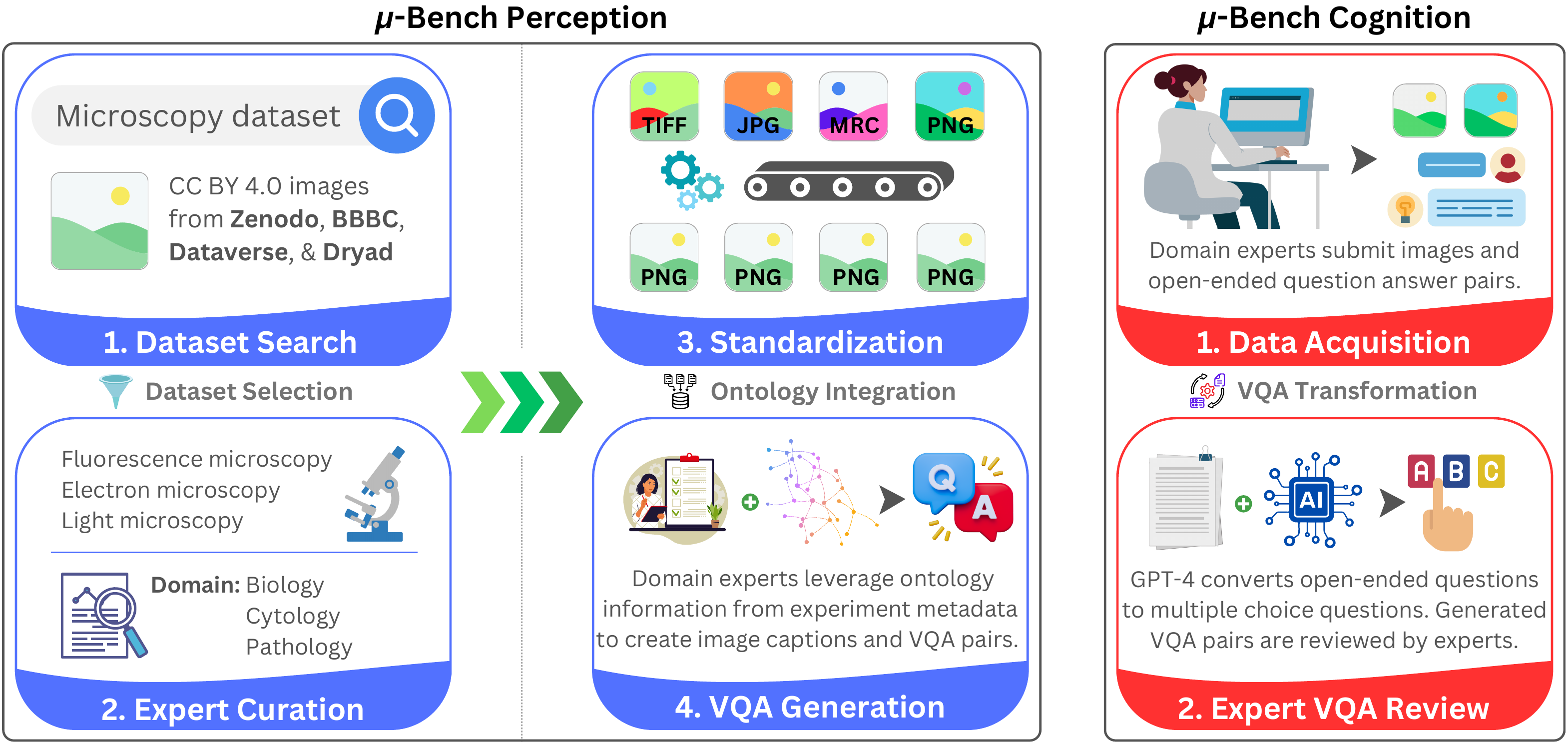}
\centering
\caption{\dataset \ construction protocol. \textbf{Perception dataset (\textcolor{blue}{left}):} first taxonomize use cases across subcellular, cellular, and tissue-level applications and collect representative datasets spanning multiple imaging modalities to test those scenarios. Next, datasets are converted to a common format, and the ontological information extracted from their metadata is standardized. Aided by this information, experts synthesize VQA pairs designed to test perception ability. \textbf{Cognition dataset (\textcolor{red}{right}):} First, domain experts use an interactive web application to upload their images and corresponding open-ended VQA pairs. Next, GPT-4 transforms the VQA pairs into a close-ended multiple-choice format. All GPT-4 generations are reviewed by experts before being incorporated into the cognition dataset.}
% \SY{describe subparts better, for example step 3 and ontology integration aren't very clear. importantly, also revise caption to be clear that cognition is something different, right now it reads like right side is the next step after left side in th same pipeline.}

\label{fig:collection-protocol}
\end{figure*}
\section{Dataset collection methodology}

% relevaring publiuc avilables to allow redistirbutioo, 
%this datasets exists 
% value our putting together value: Standraiztaion, we annoted

Recognizing the need for an expert-level benchmark in microscopy for comprehensive biological and biomedical understanding, we developed a benchmark to assess the perception and cognition capabilities of VLMs in microscopy image analysis following the methodology shown in Figure \ref{fig:collection-protocol}. At a high level, the pipeline consists of two main components:  (i) A biomedical expert categorized potential tasks and collected diverse microscopy datasets across multiple scientific domains, focusing on evaluating perception capabilities. (ii) We then complement \dataset\ by crowdsourcing questions from a larger group of microscopists using a web application.

%\cite{liang2022holistic} to asses both perception and cognition capabilities. 

 %such as its existence, count, position, and color.
 %The latter refers to compositing the perception information and the knowledge in LLM to deduce more complex answers.

%create a way to test the foundational understanding of microscopy 

\subsection{Perception Dataset Curation}

%\SY{BIG COMMENT: you need to be really clear about the dataset structure. Fig 1 says closed VQA, Captioning, localization, etc. You don't have anywhere where you, in your text, explicitly write out this separation of types of tasks, define each one, how many questions of each type there are, and define evaluation metrics for each task. For example, I'm not sure captioning is even the best phrasing for the name of the open vqa task here, since you are evaluating as accuracy later. and you don't explicitly discuss localization in the results (or closed vqa vs captioning vs localization perforamnce, etc.}

% I don't see a section on instances or segmentation. Object/instance detection is standard so we don't need to justify that. I can mention that it is important for bioimage analysis if I know where it should go.

\textbf{Dataset Review and Selection} Open data repositories, including Zenodo, Dataverse, Dryad, and BBBC, among others, were searched for microscopy biomedical image datasets. Data with permissive licenses (CC BY 4.0) allowing derivatives and redistribution were prioritized. A cell biologist and pathologist reviewed the images to ensure high quality (e.g. absence of artifacts or distortion). Diverse datasets were selected to include important biological processes (e.g., cell cycle), organelles (mitochondria, nucleus), and cell/tissue types.  Efforts were made to include diverse biological structures, microscopy modalities, and fields of study, however, the field of basic microscopy research is broad with future efforts intended to fill in gaps in coverage.
%\textbf{We recognize that while this method does not provide coverage throughout all microscopy,  it provides gold standards across the collected tasks.}

\textbf{Standardization} 
The original datasets had different organizational structures and file formats and often very little metadata. Information regarding the scientific discipline (domain), microscopy method, staining, pixel calibration, and the organism was determined by expert review or by consulting the original publication. The base experimental metadata was supplemented with manual annotation of multiple bio-ontology identifiers (SNOMED, BTO, FMA, LOINC, UBERON, etc.) to connect the image data with rich biology concepts and relationships knowledge graphs in the future. All images were converted into lossless PNG files at their original resolution with metadata in a paired json file. An MD5 checksum was computed for the image data, and each image was assigned a 128-bit unique identifier.
The image-json pairs were converted into an Apache Arrow file for public distribution and ease of use through Hugging Face datasets \cite{lhoest-etal-2021-datasets}.

\textbf{VQA task generation} We used the standardized metadata to create closed VQA questions that test capabilities at different levels: easier \textit{coarse-grained} perception and challenging \textit{fine-grained} perception (examples are shown in Figure \ref{fig:enter-label}).

The coarse-grained perception split tests basic image properties: the broad category of scientific discipline, the type of microscope, or the stain/contrast agent. These groups are visually distinct (e.g. fluorescence vs. electron microscopy) and relatively straightforward even for non-biologists, but provide important context to biology image interpretation. Although easier, these tasks are important to assess whether VLMs have commonsense knowledge of biology and microscopy.

The fine-grained perception split is more challenging. Within each category of scientific discipline or microscopy modality, there are image classes or features that need to be recognised to perform image interpretation. Dataset-specific tasks include the identification of cell type, subcellular organelles, cell cycle phase, and other biological processes that are visually distinct and important for reasoning about biological images. Solving fine-grained perception relies on finer-grained visual features, and is more challenging for humans. 

We formulate both coarse-grained and fine-grained perception as closed visual question answering (VQA). We chose this over open VQA as it's simpler to analyze, and doesn't rely on LLMs for automatic evaluation. To generate VQA options in coarse-grained perception, we designed a tree encompassing microscopy modalities, scientific domains, and staining techniques, which enables sampling fine-grained options within concepts  (e.g., selecting IHC(DAB) and IHC(RED) as likely stain options for question regarding light microscopy).

\begin{figure*}[!t]
\includegraphics[width=1.0\textwidth]{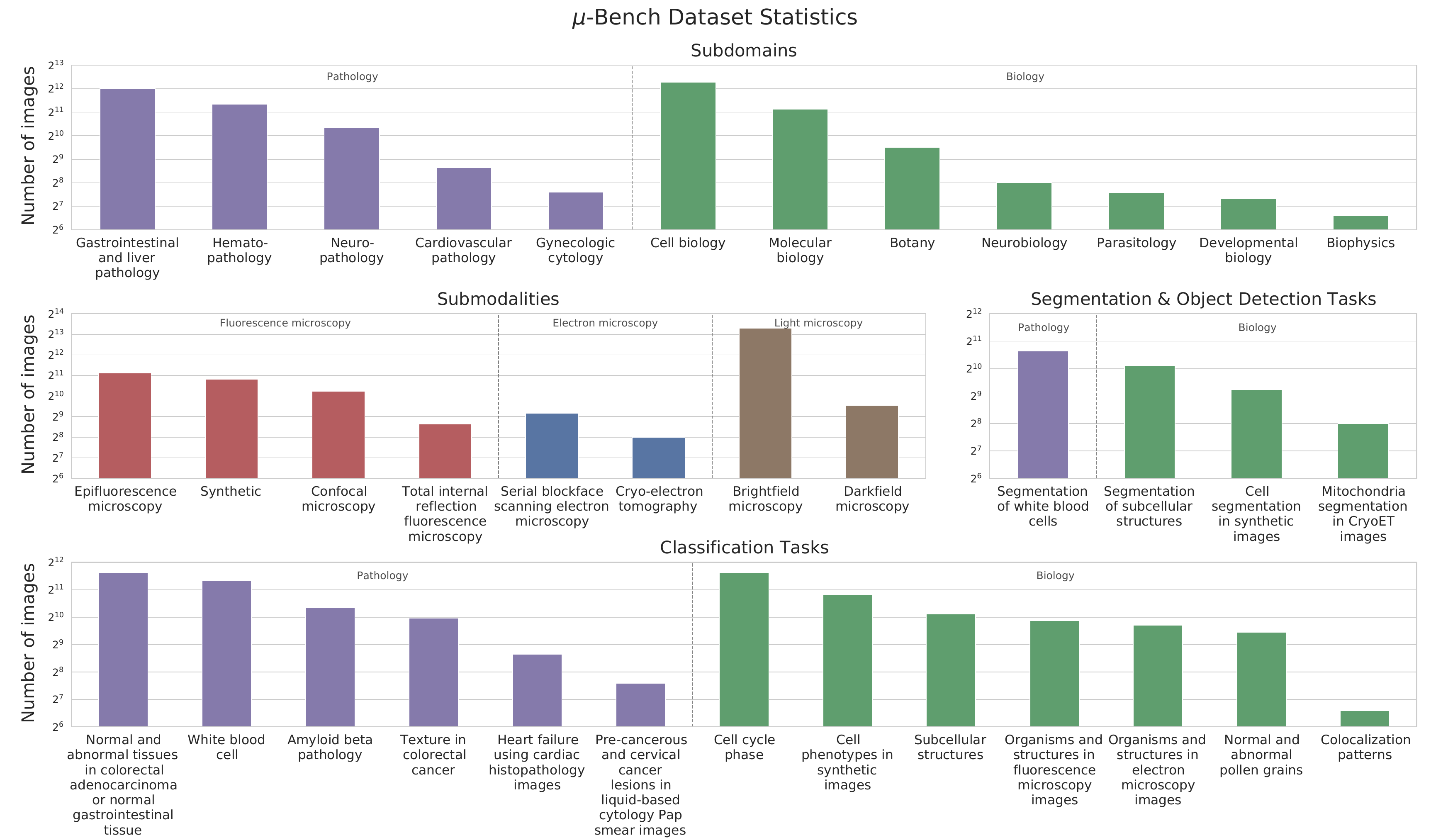}
\centering
\caption{\dataset\ Perception dataset statistics. The Perception benchmark consists of microscopy images from 12 subdomains in Biology and Pathology, obtained using 8 different imaging techniques, including light, fluorescence, and electron microscopy. It includes 17 perception fine-grained tasks: 13 for classification and 4 for segmentation or object detection.}
\label{fig:dataset_stats}
\end{figure*}

\textbf{Localization task generation} We also generate a spatial localization benchmark split, which requires predicting the bounding box or segmentation mask for a cell, nucleus or organelle (examples are in Tables \ref{fig:enter-label}). 
Understanding position and layout enables modeling spatial relationships and context, and is fundamental to image understanding. Datasets with segmentation were converted to allow instance segmentation, semantic segmentation, and object detection (bounding box and centroid).

\textbf{Quality control}
Throughout all processing, we validate the schema of each data instance to ensure a consistent format and catch/fix errors before adding them to \dataset\. The schema includes: \textit{modality}: identification of the microscopy modality (BF, EF, or EM); \textit{submodality}: identification of the microscopy sub-modality (e.g., confocal, phase contrast, or scanning electron microscopy); \textit{domain}: determination of the field of study (e.g., cell biology, histology, or pathology); \textit{subdomain}: identification of the sub-field (e.g., cancer research, neurobiology, or infectious diseases); \textit{staining}: recognition of the staining technique (e.g., H\&E, DAPI, or IHC).

\subsection{Cognitive Dataset Curation}
%\paragraph{Motivation} 
While perception datasets evaluate the fundamental capabilities of VLMs for microscopy image analysis, they fall short in assessing their ability to use perception to reason about objects. We curated a cognitive dataset to evaluate more advanced aspects like knowledge and reasoning. The cognitive dataset includes questions related to gene pathways, metabolic pathways, cell signaling and signal transduction, cell physiology and function, protein-protein interactions, cell-cell interactions, unique properties of the cell of origin or cell type in the image, cytoskeleton and cell structure or morphology, and drug mechanisms of action. These categories cover fundamental biological concepts and cellular processes to more deeply evaluate VLMs’ knowledge in understanding microscopy images.

\paragraph{Cognition Dataset Collection} We began by providing detailed guidelines for question creation to experts (see \cref{crowsoucing}), which ensured consistency and quality across the dataset. Using an internal chat-like web interface, we asked domain experts to submit questions reflecting their daily research activities. We encouraged a focus on questions that required challenging image-based reasoning, domain expertise, interpretation of experimental results, or hypothesis generation. %rather than straightforward tasks like classification or counting.

In addition to submitting questions, experts provided crucial context regarding experimental details, image acquisition, organisms, treatments, and image descriptions. With this comprehensive information, GPT-4V generated answers to the submitted questions. These answers were subsequently reviewed by experts, who evaluated the accuracy and interpretation of the responses. %Overall, we collected 121 pairs, each consisting of an image, a question, a GPT-4V answer, and expert feedback.

\textbf{Multiple-Choice Question Transformation} The collected pairs (image, question, GPT-4V answer, feedback) were transformed into multiple-choice questions using GPT-4. This transformation was guided by a carefully designed prompt ( ~\cref{sec:todo2}), verified by a cell biologist and a pathologist, to ensure the questions are challenging and reflective of real-world scenarios faced by biomedical researchers. Each transformed question includes an image, a question, and six candidate choices. One choice is correct, while the other five are distractors generated by GPT-4, where one choice is ``None of the above.'' Domain experts verified the validity of the generated questions and manually corrected a small number of questions. Finally, we ensured that correct answers were uniformly distributed among answer choices A to F and formalized the entire dataset.

%%%%%%%%%%%%%%%%%%%%%%%%%%%%%%%%%%%
%%%%%%%% Dataset Description %%%%%%
%%%%%%%%%%%%%%%%%%%%%%%%%%%%%%%%%%%
\section{Dataset Description}
% \input{sections/dataset_description}

%\JB{Can we group this with the previous section on dataset? Also, since we have a figure for statistics, we can have less text here.}

\textbf{Perception Dataset Statistics} For our perception benchmark, we collected a total of 17,235 microscopy images from \numdatasets\ distinct public datasets (see Table \ref{table:sources}) with permissive licensing, prioritizing open CC-BY licenses. To the best of our knowledge, \dataset \ Perception is the most diverse microscopy vision-language benchmark, spanning light (LM), fluorescence (FM), and electron microscopy (EM), covering \numsubmodalities\ 
 microscopy sub-modalities (see Figure \ref{fig:dataset_stats}), \perceptionuniquecellmicro\ unique cells, tissues, and structures over \numstaining\ unique staining techniques (see Figure \ref{repalebystaining}). The perception benchmark subset spans this diversity through closed VQA, object detection, and segmentation.

\textbf{Cognition Dataset Statistics} For our cognition benchmark, we collected \cognitionimages\ microscopy images and \cognitionquestions\ questions from experts in the field. Entries were received from \cognitionusers\ users across \cognitioninstitutions\ different institutions. The \dataset \ Cognition dataset encompasses 3 modalities (fluorescence, electron, light) with 12 sub-modalities, 2 domains (pathology and biology) with 14 sub-domains, and 3 scales (nano, micro, macro), covering a diverse range of topics such as pathology, immunology, and virology. Distributions are shown in Appendix Table \ref{tbl:uBenchCognitionDomainStatistics}.

%%%%%%%%%%%%%%%%%%%%%%%%%%%%%%%%%%%
%%%%%%%%%% Benchmarking %%%%%%%%%%%
%%%%%%%%%%%%%%%%%%%%%%%%%%%%%%%%%%%

\section{VLM benchmarking and results}
\subsection{Benchmarking approach}

Data artifacts like \dataset\ enable studying model behavior within specialist domains. Since our benchmark covers a wide range of biomedical tasks, we can, for the first time, compare biomedical perception and cognition capabilities across microscopy imagining modalities. In this section, we show the utility of \dataset\ by reporting empirical findings on a range of VLMs.

 \begin{table}[ht]
    \centering
    \caption{ Macro-average accuracy (with bootstrap confidence interval) for coarse-grained and fine-grained perception and cognition  (reasoning) in \dataset\ . }
    \begin{adjustbox}{width=\textwidth}
    \begin{tabular}{c c c c c c}
        \toprule 
        \rowcolor{gray!10} \multicolumn{6}{c}{$\mu$-Bench} \\
        \midrule
        \midrule
        \multicolumn{2}{c}{Perception (Coarse-Grained)} & \multicolumn{2}{c}{ Perception (Fine-Grained)} & \multicolumn{2}{c}{Cognition (Reasoning) } \\
        \midrule
        Model & Accuracy ($\pm$ CI) & Model & Accuracy ($\pm$ CI) & Model & Accuracy ($\pm$ CI) \\
        \midrule 
        \midrule
        \GPT{62.68}{0.35} & \GPT{51.73}{0.82} & \GPT{62.00}{9.00} \\
        \COGVLM{52.05}{0.35} & \BIOMEDCLIP{34.65}{0.75} & \QWENVLM{41.00}{10.00} \\
        \QWENVLM{49.85}{0.35} & \CONCH{33.64}{0.72} & \COGVLM{41.00}{10.00} \\
        \BIOMEDCLIP{47.57}{0.34} & \ALIGN{31.9}{0.72} & \OPENCLIP{38.33}{8.33} \\
        \ALIGN{40.7}{0.34} & \CLIP{30.09}{0.71} & \ALIGN{31.00}{9.00} \\
        \OPENCLIP{36.34}{0.33} & \OPENCLIP{29.36}{0.69} & \CLIP{28.00}{9.00} \\
        \PALIGEMMA{36.29}{0.33} & \COGVLM{28.18}{0.70} & \PALIGEMMA{25.00}{8.00} \\
        \CLIP{35.41}{0.34} & \QUILTCLIP{27.85}{0.69} & \BIOMEDCLIP{25.00}{8.00} \\
        \PLIP{31.11}{0.32} & \QWENVLM{27.81}{0.70} & \CONCH{18.00}{7.00} \\
        \CONCH{27.84}{0.31} & \PLIP{25.49}{0.68} & \RANDOM{17.00}{7.00} \\
        
        \QUILTCLIP{26.58}{0.31} & \PALIGEMMA{21.29}{0.64} & \PLIP{17.00}{7.00} \\
        \RANDOM{18.34}{0.27} & \RANDOM{19.13}{0.60} & \QUILTCLIP{13.00}{6.00} \\
        \bottomrule
    \end{tabular}
    \end{adjustbox}
    \small{\textsuperscript{ \ding{59}} \colorbox{newgreen}{ } General autoregressive VLMs \colorbox{newblue}{}  General contrastive VLMS   \colorbox{newpink}{} Pathology contrastive VLMS  \\ \colorbox{newpurple}{}  Biomedical contrastive VLMS.}
\label{tbl:uBenchPerformance}
\end{table}

%Story:
%  Benchmark allows characterize model behanvor that was not previously observed due to the lack of holistic benchmarks
% we leverage dataset to study these behavior and understand current limitations if any
% we find that chagpt is the best model for microscopy, surpassing even fine-tuned contrastive models . However this in not true for all applications

First, we categorized VLMs into two groups: generalist models trained on natural images and language, and `specialist' models, fine-tuned on biomedical data. Within generalist models, we also distinguish between contrastive and auto-regressive models.

\paragraph{Generalist Contrastive (GC) VLMs}
We evaluate ALIGN \cite{jia2021scaling}, OpenCLIP \cite{cherti2023reproducible}, and CLIP \cite{radford2021learning} as the canonical contrastive models for natural images. Notably, OpenCLIP and CLIP serve as foundational models for numerous specialist biomedical VLMs.

\paragraph{Generalist autoregressive (GA) VLMs}
We evaluate with GPT-4o \cite{achiam2023gpt}, a state-of-the-art enterprise VLM. 
For open source models, we test CogVLM \cite{wang2023cogvlm}, QwenVLM \cite{bai2023qwen}, and PaliGemma \cite{team2024gemma} for their strong performance on general domain tasks, instruction-following capabilities, and, for QwenVLM and PaliGemma only, their ability to perform object detection. 

\paragraph{Specialist contrastive (SC) VLMs}
Our specialist model selection had two constraints: choosing the best-performing models and preferring minimal architectural changes from their base generalist versions, allowing performance analysis based on variations in training mixtures. We selected  BiomedCLIP \cite{zhang2023biomedclip} since it is a strong model that covers all biomedical imaging modalities in our benchmark (trained on 15 million image-text pairs collected from PubMedCentral).
Additionally, we included three pathology VLMs: PLIP (CLIP trained on H\&E) \cite{huang2023visual}, QuiltNet (CLIP trained on H\&E and IHC) \cite{ikezogwo2024quilt}, and CONCH (CoCa trained on H\&E and IHC) \cite{lu2023towards}, with training dataset sizes of 208k, 1 million, and 1.2 million, respectively. While CONCH and BiomedCLIP are based on OpenCLIP and CoCa \cite{yu2022coca} respectively, they modify the architecture or training strategy. 

\paragraph{Evaluation} The Closed VQA component of \dataset\ was evaluated with accuracy, generating confidence intervals (CI)  via bootstrap \cite{efron1994introduction}   (\cref{sec:cofidence_intervals}). 
Object detection was evaluated for models with object detection capabilities (PaliGemma and QwenVLM) in open VQA format using the GRIT localization metric \cite{gupta2022grit} as adopted by prior works \cite{bai2023qwen}.

%%%%%%%%%%%%%%%%%%%%%%%%%%%%%%%%%%%
%%%%%%%%%% Resulsts %%%%%%%%%%%%%%%
%%%%%%%%%%%%%%%%%%%%%%%%%%%%%%%%%%%

\subsection{Results}

%\begin{figure*}
    
%\end{figure*}

\begin{figure*}[ht!]
\centering
% \subfloat[]{
\includegraphics[width=0.54\linewidth, page=11, trim={0 1000 540 0},clip]{images/figs-robustness.pdf}
% } 
\hspace{-2mm}
% \hfill
%%% General Models (Contrastive)
\subfloat[Perception (coarse-grained)]{\includegraphics[width=0.38\columnwidth]{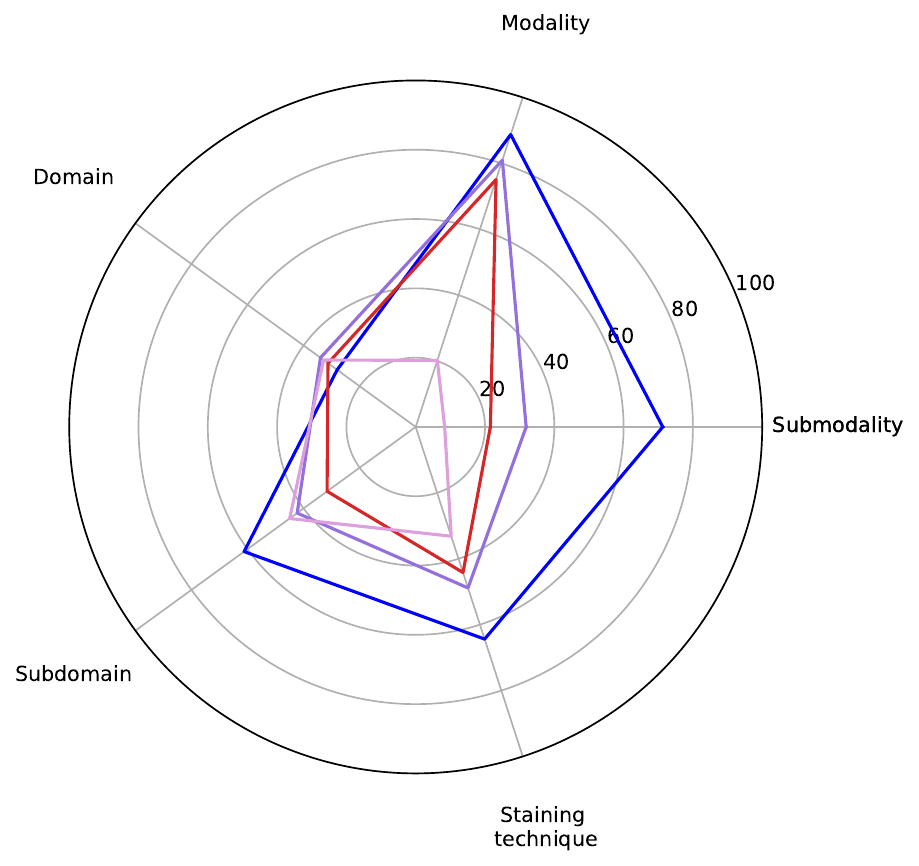}}
\subfloat[Perception (fine-grained)]{\includegraphics[width=0.38\columnwidth]{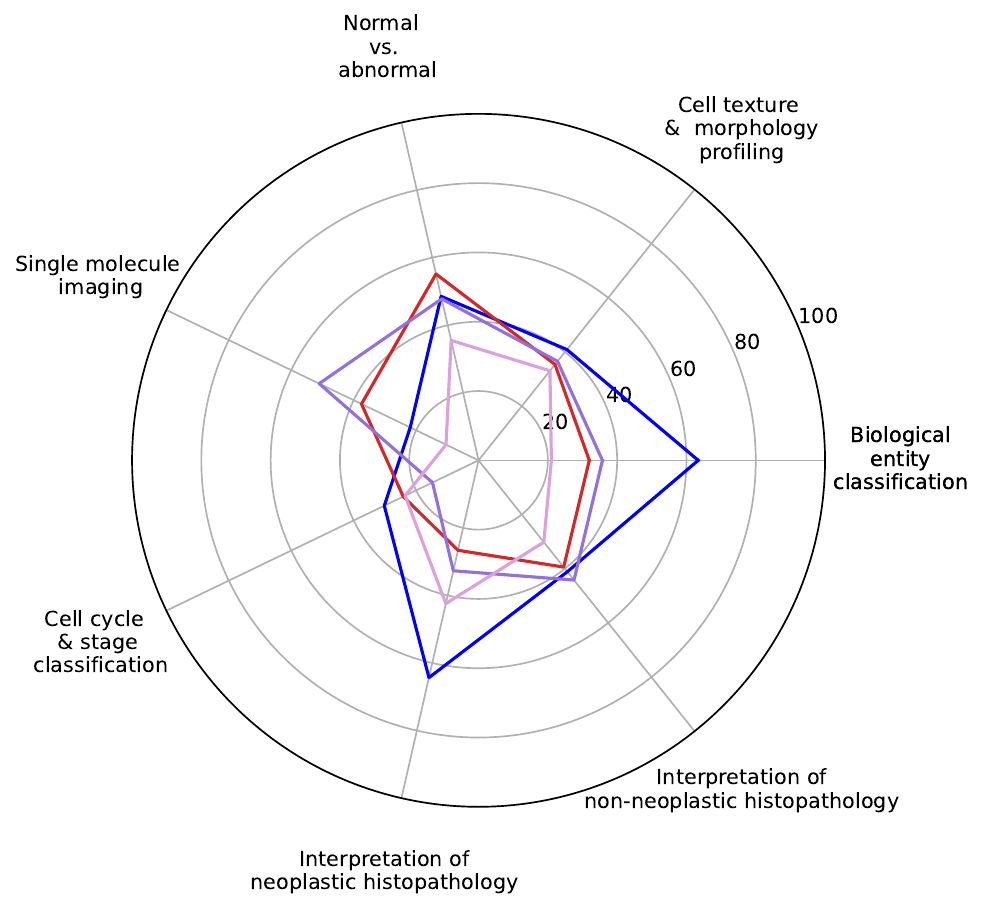}}  \\
\subfloat[ Pathology only Perception \\(coarse-grained) ]{\includegraphics[width=0.38\columnwidth]
{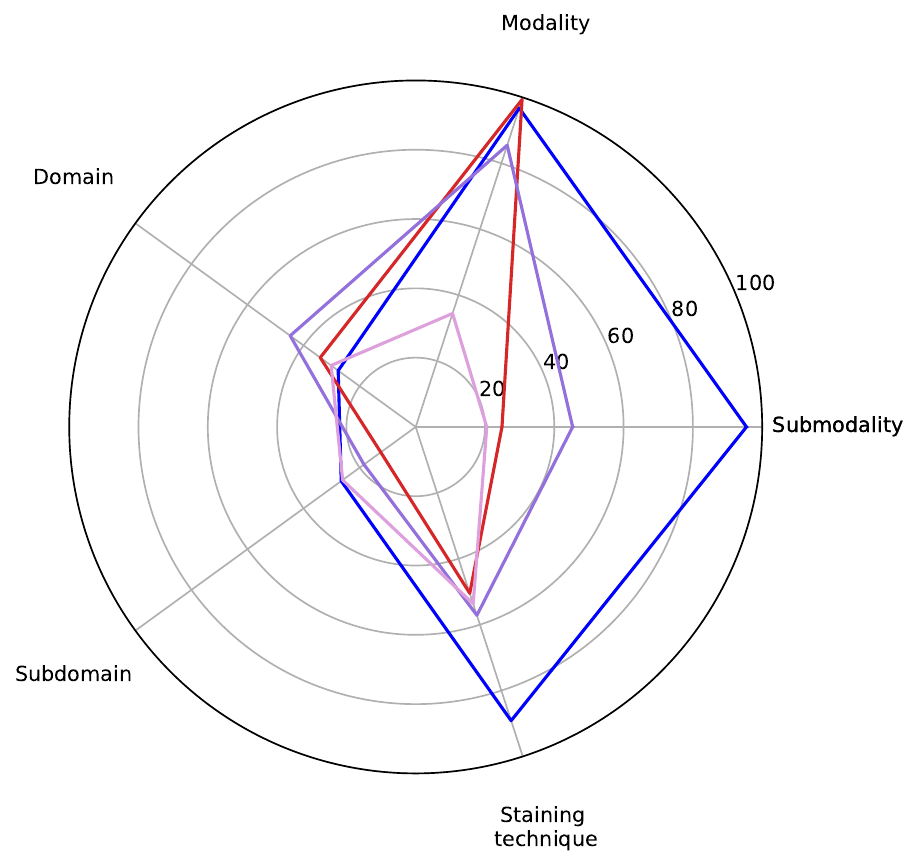}} 
\subfloat[ Pathology only Perception \\  (fine-grained)]{\includegraphics[width=0.38\columnwidth]{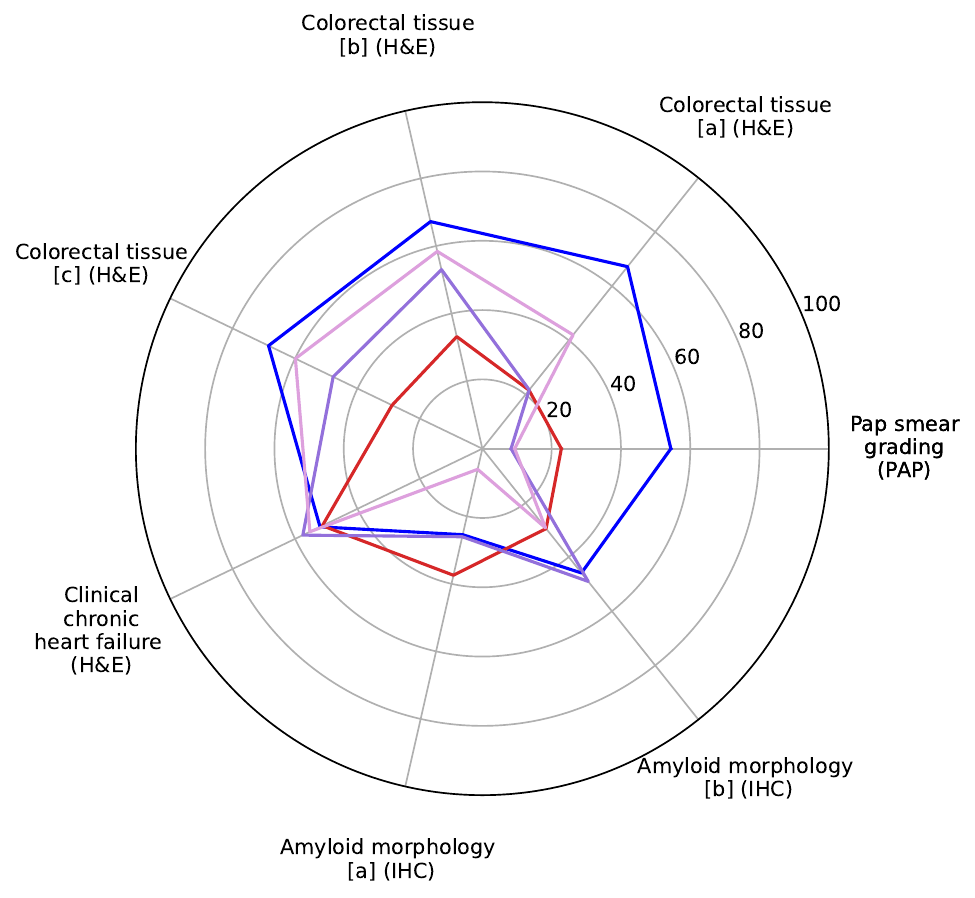}}

\caption{Performance comparison on the perception benchmark for the best-performing general domain auto-regressive model (GPT-4o), contrastive model (ALIGN), specialist biomedical contrastive model (BiomedCLIP), and specialist pathology contrastive model (CONCH). The top row shows performance in  all  of the \dataset\, while the bottom row shows pathology-only samples.
% \SY{text too small, check all figures for text size}
} 
 \label{radial-path-finegrained}
\end{figure*}

 \paragraph{All models have high error rates}
 %\paragrapAh{GPT-4o performs best, but all models show high error rates} 
Table \ref{tbl:uBenchPerformance} shows the accuracy performance across perception and cognition. Even the top-performing model (GPT-4o) has high error rates, with an accuracy of 62.6\% on coarse-grained perception, 51.7\%  on fine-grained perception, and 62.0\%  on cognition tasks.  \textit{On average} GPT-4o  also outperforms BiomedCLIP (the best biomedical SC VLM) by a minimum of 15\%  in all evaluation dimensions and  CONCH (the best pathology SC VLM) in pathology-specific perception tasks, showing a difference of 39.37\% on coarse-grained and 19.40\% in fined-grained tasks (as illustrated in Table \ref{tbl:uBenchPerformance}. However, finer subgroup analysis (Figure \ref{fig:main_results})  shows that GP-4o does not excel in all perception tasks, including domain identification (coarse-grained) and single-molecule imaging,  normal vs abnormal classification, and non-neoplastic histopathology interpretation (fine-grained). The model architecture and training data for GPT-4 are closed source, making it challenging to conclude from these results. However, GPT-4o's high error rates, its substantial gap compared to SC models, and performance variation across task subgroups highlight that $\mu$-Bench is challenging and is not saturated by state-of-the-art general, biomedical, and pathology models.

\textbf{Specialist biomedical models are often worse than non-specialist models}
While specialist models are explicitly developed for the biomedical domain, they often underperform non-specialized open-source models. For example, in both coarse-grained perception and cognition tasks (Table \ref{tbl:uBenchPerformance}), GA models (CogVLM and QwenVLM) outperform the best SC model (BiomedCLIP) by  4.4\% and  16.0\% margins respectively. While GA models have a different training objective, larger training mixture, and more model parameters, a similar trend is observed with   GC models (ALIGN, OpenCLIP, and CLIP)  as they outperform all pathology VLMs in the same tasks by at least 9.5\%  (PLIP- ALIGN) and 20.3\% (CONCH - OpenCLIP) respectively. This ranking is reversed in fine-grained perception tasks, where  BiomedCLIP and CONCH perform best. Indeed, fine-grained perception closely resembles the data mixture used to fine-tune contrastive specialist models \cite{Huang2023-plip}. This characterization shows weakness in current microscopy biomedical model development, which we investigate next.

\textbf{Specialist training can cause catastrophic forgetting}
Generalist contrastive models like (OpenCLIP and CLIP) surprisingly outperform their fine-tuned counterparts (PILP and QuiltNet) in coarse-grained perception and cognition
% , even across domains covered by the fine-tuned mixture 
(Table \ref{tbl:uBenchPerformance}).
Specifically, PILP and QuiltNet are fine-tuned directly from OpenCLIP and CLIP using only pathology data closest to \dataset\  fine-grained perception tasks. Although it improves performance on pathology-specific fine-grained tasks (Figure \ref{radial-path-finegrained}), it degrades performance on all other tasks (Table \ref{tbl:uBenchPerformance}).

\begin{figure}
\label{fig:ensembling}
\subfloat[Perception (fine-grained)]{
    \centering
    \includegraphics[width=0.54\linewidth, page=7, trim={0 255 0 0},clip]{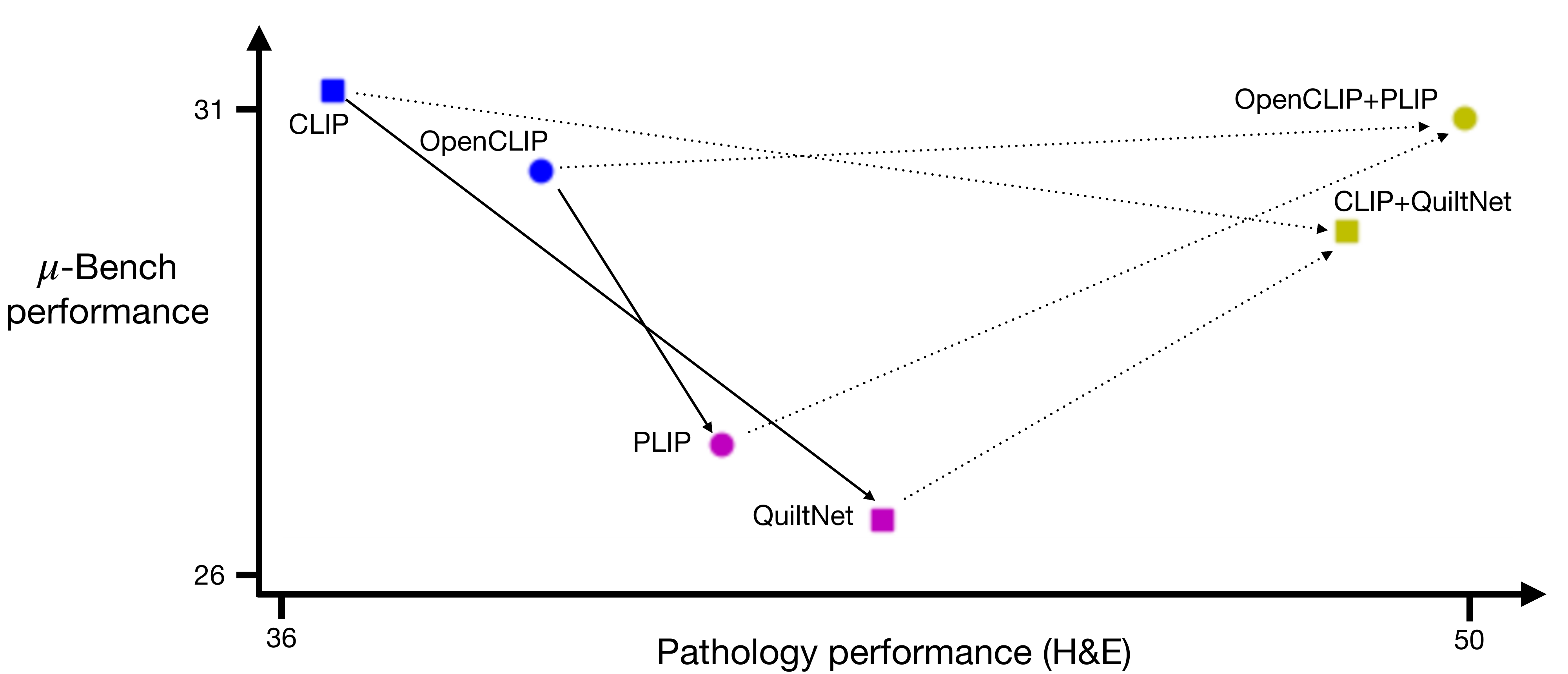}
    }
\hfill
\subfloat[Perception (coarse-grained)]{
    \centering
    \includegraphics[width=0.40\linewidth, page=9, trim={0 255 640 0},clip]{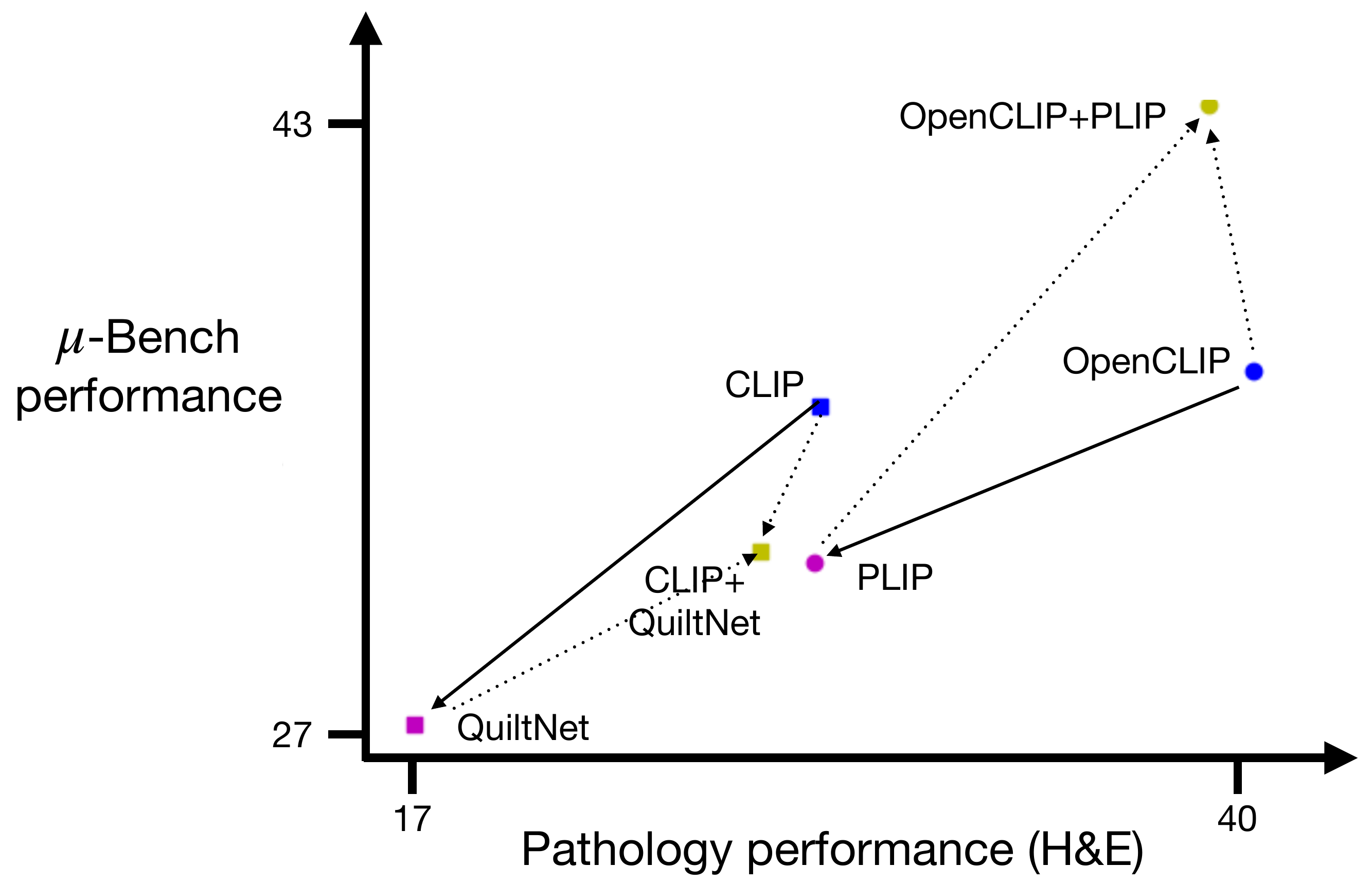}
    }

\caption{
Fine-tuning and microscopy perception generalization on \dataset\ .  Base CLIP models (\textcolor{blue}{blue}) are fine-tuned to PLIP and QuiltNet using pathology data mixtures (\textcolor{violet}{pink}). Weight-merging base models with their corresponding fine-tuned models (\textcolor{olive}{olive}) improves specialist zero-shot performance on \dataset\ coarse-grained (\textbf{Left})  and fined-grained (\textbf{Right}) perception. }

\label{fig:robust}
\end{figure}

\textbf{\dataset\ characterization drives robust model development} 
\label{sec:results-model_ensembling}
To address catastrophic forgetting identified in our multi-level evaluation, we ensemble base model weights (OpenCLIP / CLIP) with fine-tuned model weights (PLIP/QuiltNet) to create merged models (PLIP+OpenCLIP / QuiltNet+CLIP), as suggested in \cite{wortsman2022robust}. As shown in Figure \ref{fig:robust}, when comparing merged models to their fine-tuned counterparts, perception performance increases across all of  \dataset\ (y-axis), including pathology-specific tasks (x-axis).

\textbf{\dataset\ supports probing design decisions for biomedical VLMs}
We have shown that \dataset\ offers valuable insights into microscopy biomedical VLMs and hope it encourages further evaluations of design choices.
Data diversity is one factor: Table \ref{tbl:uBenchPerformance} illustrates that BiomedCLIP, trained across all microscopy modalities in \dataset, surpasses specialist models, albeit with a smaller margin for fine-grained tasks compared to CONCH, which uses pathology data.
Regarding model architecture and training strategy, generalist autoregressive models (CogVLM and QwenVLM) outperform contrastive models (ALIGN and CLIP) in coarse-grained perception, but the opposite is true for fine-grained perception.
For object localization, PaliGemma outperformed QwenVLM (Table \ref{tab:localization}) on \dataset, though both performed poorly, and no specialist models support detection.
Future research could explore prompting strategies, data curation, and new methods to mitigate catastrophic forgetting.

\section{Conclusion}

Benchmarks drive advancements in machine learning by providing a standard to measure progress and allowing researchers to identify weaknesses in current approaches. Thus, the lack of biomedical vision-language benchmarks limits the ability to develop and evaluate specialist VLMs. We address this gap in microscopy by introducing the most extensive collection of vision-language tasks spanning perception and cognition. We use \dataset \ to establish, for the first time, the performance of some of the most capable VLMs available and find high error rates of 30\%, highlighting room for improvement. We demonstrate how \dataset\ can be leveraged to generate new insights. Lastly, we share \dataset \ to enable researchers to measure progress in microscopy foundation models.

%Bibliography
\bibliographystyle{unsrt}  
\bibliography{ubench}

\begin{thebibliography}{10}

\bibitem{merkle2013ascent}
Arno~P Merkle and Jeff Gelb.
\newblock The ascent of 3d x-ray microscopy in the laboratory.
\newblock {\em Microscopy Today}, 21(2):10--15, 2013.

\bibitem{weber2021multidisciplinarity}
Michael Weber and Jan Huisken.
\newblock Multidisciplinarity is critical to unlock the full potential of modern light microscopy.
\newblock {\em Frontiers in Cell and Developmental Biology}, 9:739015, 2021.

\bibitem{zhang2023biomedclip}
Sheng Zhang, Yanbo Xu, Naoto Usuyama, Hanwen Xu, Jaspreet Bagga, Robert Tinn, Sam Preston, Rajesh Rao, Mu~Wei, Naveen Valluri, et~al.
\newblock Biomedclip: a multimodal biomedical foundation model pretrained from fifteen million scientific image-text pairs.
\newblock {\em arXiv preprint arXiv:2303.00915}, 2023.

\bibitem{foss2002end}
Laurence Foss.
\newblock {\em The end of modern medicine: Biomedical science under a microscope}.
\newblock SUNY Press, 2002.

\bibitem{tu2024towards}
Tao Tu, Shekoofeh Azizi, Danny Driess, Mike Schaekermann, Mohamed Amin, Pi-Chuan Chang, Andrew Carroll, Charles Lau, Ryutaro Tanno, Ira Ktena, et~al.
\newblock Towards generalist biomedical ai.
\newblock {\em NEJM AI}, 1(3):AIoa2300138, 2024.

\bibitem{gao2024empowering}
Shanghua Gao, Ada Fang, Yepeng Huang, Valentina Giunchiglia, Ayush Noori, Jonathan~Richard Schwarz, Yasha Ektefaie, Jovana Kondic, and Marinka Zitnik.
\newblock Empowering biomedical discovery with ai agents.
\newblock {\em arXiv preprint arXiv:2404.02831}, 2024.

\bibitem{wang2023scientific}
Hanchen Wang, Tianfan Fu, Yuanqi Du, Wenhao Gao, Kexin Huang, Ziming Liu, Payal Chandak, Shengchao Liu, Peter Van~Katwyk, Andreea Deac, et~al.
\newblock Scientific discovery in the age of artificial intelligence.
\newblock {\em Nature}, 620(7972):47--60, 2023.

\bibitem{schwartz2023scaling}
Morgan Schwartz, Uriah Israel, Xuefei Wang, Emily Laubscher, Changhua Yu, Rohit Dilip, Qilin Li, Joud Mari, Johnathon Soro, Kevin Yu, et~al.
\newblock Scaling biological discovery at the interface of deep learning and cellular imaging.
\newblock {\em Nature Methods}, 20(7):956--957, 2023.

\bibitem{malacrida2023phasor}
Leonel Malacrida.
\newblock Phasor plots and the future of spectral and lifetime imaging.
\newblock {\em Nature Methods}, 20(7):965--967, 2023.

\bibitem{nogare2023using}
Damian~Dalle Nogare, Matthew Hartley, Joran Deschamps, Jan Ellenberg, and Florian Jug.
\newblock Using ai in bioimage analysis to elevate the rate of scientific discovery as a community.
\newblock {\em Nature methods}, 20(7):973--975, 2023.

\bibitem{carpenter2023smart}
Anne~E Carpenter, Beth~A Cimini, and Kevin~W Eliceiri.
\newblock Smart microscopes of the future.
\newblock {\em Nature methods}, 20(7):962--964, 2023.

\bibitem{li2023challenges}
Xinyang Li, Yuanlong Zhang, Jiamin Wu, and Qionghai Dai.
\newblock Challenges and opportunities in bioimage analysis.
\newblock {\em Nature Methods}, 20(7):958--961, 2023.

\bibitem{lambert2023towards}
Talley Lambert and Jennifer Waters.
\newblock Towards effective adoption of novel image analysis methods.
\newblock {\em Nature Methods}, 20(7):971--972, 2023.

\bibitem{hein2023global}
Marco~Y Hein, Duo Peng, Verina Todorova, Frank McCarthy, Kibeom Kim, Chad Liu, Laura Savy, Camille Januel, Rodrigo Baltazar-Nunez, Sophie Bax, et~al.
\newblock Global organelle profiling reveals subcellular localization and remodeling at proteome scale.
\newblock {\em bioRxiv}, pages 2023--12, 2023.

\bibitem{brown2020language}
Tom Brown, Benjamin Mann, Nick Ryder, Melanie Subbiah, Jared~D Kaplan, Prafulla Dhariwal, Arvind Neelakantan, Pranav Shyam, Girish Sastry, Amanda Askell, et~al.
\newblock Language models are few-shot learners.
\newblock {\em Advances in neural information processing systems}, 33:1877--1901, 2020.

\bibitem{vaswani2017attention}
Ashish Vaswani, Noam Shazeer, Niki Parmar, Jakob Uszkoreit, Llion Jones, Aidan~N Gomez, {\L}ukasz Kaiser, and Illia Polosukhin.
\newblock Attention is all you need.
\newblock {\em Advances in neural information processing systems}, 30, 2017.

\bibitem{kirillov2023segment}
Alexander Kirillov, Eric Mintun, Nikhila Ravi, Hanzi Mao, Chloe Rolland, Laura Gustafson, Tete Xiao, Spencer Whitehead, Alexander~C Berg, Wan-Yen Lo, et~al.
\newblock Segment anything.
\newblock In {\em Proceedings of the IEEE/CVF International Conference on Computer Vision}, pages 4015--4026, 2023.

\bibitem{caron2021emerging}
Mathilde Caron, Hugo Touvron, Ishan Misra, Herv{\'e} J{\'e}gou, Julien Mairal, Piotr Bojanowski, and Armand Joulin.
\newblock Emerging properties in self-supervised vision transformers.
\newblock In {\em Proceedings of the IEEE/CVF international conference on computer vision}, pages 9650--9660, 2021.

\bibitem{radford2021learning}
Alec Radford, Jong~Wook Kim, Chris Hallacy, Aditya Ramesh, Gabriel Goh, Sandhini Agarwal, Girish Sastry, Amanda Askell, Pamela Mishkin, Jack Clark, et~al.
\newblock Learning transferable visual models from natural language supervision.
\newblock In {\em International conference on machine learning}, pages 8748--8763. PMLR, 2021.

\bibitem{alayrac2022flamingo}
Jean-Baptiste Alayrac, Jeff Donahue, Pauline Luc, Antoine Miech, Iain Barr, Yana Hasson, Karel Lenc, Arthur Mensch, Katherine Millican, Malcolm Reynolds, et~al.
\newblock Flamingo: a visual language model for few-shot learning.
\newblock {\em Advances in neural information processing systems}, 35:23716--23736, 2022.

\bibitem{bai2023qwen}
Jinze Bai, Shuai Bai, Shusheng Yang, Shijie Wang, Sinan Tan, Peng Wang, Junyang Lin, Chang Zhou, and Jingren Zhou.
\newblock Qwen-vl: A frontier large vision-language model with versatile abilities.
\newblock {\em arXiv preprint arXiv:2308.12966}, 2023.

\bibitem{ma2024multimodality}
Jun Ma, Ronald Xie, Shamini Ayyadhury, Cheng Ge, Anubha Gupta, Ritu Gupta, Song Gu, Yao Zhang, Gihun Lee, Joonkee Kim, et~al.
\newblock The multimodality cell segmentation challenge: toward universal solutions.
\newblock {\em Nature methods}, pages 1--11, 2024.

\bibitem{Chandrasekaran2024-jump1}
Srinivas~Niranj Chandrasekaran, Beth~A Cimini, Amy Goodale, Lisa Miller, Maria Kost-Alimova, Nasim Jamali, John~G Doench, Briana Fritchman, Adam Skepner, Michelle Melanson, et~al.
\newblock Three million images and morphological profiles of cells treated with matched chemical and genetic perturbations.
\newblock {\em Nature Methods}, pages 1--8, 2024.

\bibitem{he2020pathvqa}
Xuehai He, Yichen Zhang, Luntian Mou, Eric Xing, and Pengtao Xie.
\newblock Pathvqa: 30000+ questions for medical visual question answering.
\newblock {\em arXiv preprint arXiv:2003.10286}, 2020.

\bibitem{Huang2023-plip}
Zhi Huang, Federico Bianchi, Mert Yuksekgonul, Thomas~J Montine, and James Zou.
\newblock A visual--language foundation model for pathology image analysis using medical twitter.
\newblock {\em Nature medicine}, 29(9):2307--2316, 2023.

\bibitem{fu2023mme}
Chaoyou Fu, Peixian Chen, Yunhang Shen, Yulei Qin, Mengdan Zhang, Xu~Lin, Jinrui Yang, Xiawu Zheng, Ke~Li, Xing Sun, et~al.
\newblock Mme: A comprehensive evaluation benchmark for multimodal large language models.
\newblock {\em arXiv preprint arXiv:2306.13394}, 2023.

\bibitem{caicedo2019nucleus}
Juan~C Caicedo, Allen Goodman, Kyle~W Karhohs, Beth~A Cimini, Jeanelle Ackerman, Marzieh Haghighi, CherKeng Heng, Tim Becker, Minh Doan, Claire McQuin, et~al.
\newblock Nucleus segmentation across imaging experiments: the 2018 data science bowl.
\newblock {\em Nature methods}, 16(12):1247--1253, 2019.

\bibitem{saltz2018spatial}
Joel Saltz, Rajarsi Gupta, Le~Hou, Tahsin Kurc, Pankaj Singh, Vu~Nguyen, Dimitris Samaras, Kenneth~R Shroyer, Tianhao Zhao, Rebecca Batiste, et~al.
\newblock Spatial organization and molecular correlation of tumor-infiltrating lymphocytes using deep learning on pathology images.
\newblock {\em Cell reports}, 23(1):181--193, 2018.

\bibitem{antonelli2022medical}
Michela Antonelli, Annika Reinke, Spyridon Bakas, Keyvan Farahani, Annette Kopp-Schneider, Bennett~A Landman, Geert Litjens, Bjoern Menze, Olaf Ronneberger, Ronald~M Summers, et~al.
\newblock The medical segmentation decathlon.
\newblock {\em Nature communications}, 13(1):4128, 2022.

\bibitem{rajpurkar2017chexnet}
Pranav Rajpurkar, Jeremy Irvin, Kaylie Zhu, Brandon Yang, Hershel Mehta, Tony Duan, Daisy Ding, Aarti Bagul, Curtis Langlotz, Katie Shpanskaya, et~al.
\newblock Chexnet: Radiologist-level pneumonia detection on chest x-rays with deep learning.
\newblock {\em arXiv preprint arXiv:1711.05225}, 2017.

\bibitem{esteva2017dermatologist}
Andre Esteva, Brett Kuprel, Roberto~A Novoa, Justin Ko, Susan~M Swetter, Helen~M Blau, and Sebastian Thrun.
\newblock Dermatologist-level classification of skin cancer with deep neural networks.
\newblock {\em nature}, 542(7639):115--118, 2017.

\bibitem{jia2021scaling}
Chao Jia, Yinfei Yang, Ye~Xia, Yi-Ting Chen, Zarana Parekh, Hieu Pham, Quoc Le, Yun-Hsuan Sung, Zhen Li, and Tom Duerig.
\newblock Scaling up visual and vision-language representation learning with noisy text supervision.
\newblock In {\em International conference on machine learning}, pages 4904--4916. PMLR, 2021.

\bibitem{achiam2023gpt}
Josh Achiam, Steven Adler, Sandhini Agarwal, Lama Ahmad, Ilge Akkaya, Florencia~Leoni Aleman, Diogo Almeida, Janko Altenschmidt, Sam Altman, Shyamal Anadkat, et~al.
\newblock Gpt-4 technical report.
\newblock {\em arXiv preprint arXiv:2303.08774}, 2023.

\bibitem{moor2023foundation}
Michael Moor, Oishi Banerjee, Zahra Shakeri~Hossein Abad, Harlan~M Krumholz, Jure Leskovec, Eric~J Topol, and Pranav Rajpurkar.
\newblock Foundation models for generalist medical artificial intelligence.
\newblock {\em Nature}, 616(7956):259--265, 2023.

\bibitem{m2024augmenting}
Andres M.~Bran, Sam Cox, Oliver Schilter, Carlo Baldassari, Andrew~D White, and Philippe Schwaller.
\newblock Augmenting large language models with chemistry tools.
\newblock {\em Nature Machine Intelligence}, pages 1--11, 2024.

\bibitem{schuhmann2022laion}
Christoph Schuhmann, Romain Beaumont, Richard Vencu, Cade Gordon, Ross Wightman, Mehdi Cherti, Theo Coombes, Aarush Katta, Clayton Mullis, Mitchell Wortsman, et~al.
\newblock Laion-5b: An open large-scale dataset for training next generation image-text models.
\newblock {\em Advances in Neural Information Processing Systems}, 35:25278--25294, 2022.

\bibitem{zhao2023clip}
Zihao Zhao, Yuxiao Liu, Han Wu, Yonghao Li, Sheng Wang, Lin Teng, Disheng Liu, Xiang Li, Zhiming Cui, Qian Wang, et~al.
\newblock Clip in medical imaging: A comprehensive survey.
\newblock {\em arXiv preprint arXiv:2312.07353}, 2023.

\bibitem{huang2023visual}
Zhi Huang, Federico Bianchi, Mert Yuksekgonul, Thomas~J Montine, and James Zou.
\newblock A visual--language foundation model for pathology image analysis using medical twitter.
\newblock {\em Nature medicine}, 29(9):2307--2316, 2023.

\bibitem{yu2022coca}
Jiahui Yu, Zirui Wang, Vijay Vasudevan, Legg Yeung, Mojtaba Seyedhosseini, and Yonghui Wu.
\newblock Coca: Contrastive captioners are image-text foundation models.
\newblock {\em arXiv preprint arXiv:2205.01917}, 2022.

\bibitem{lhoest-etal-2021-datasets}
Quentin Lhoest, Albert Villanova~del Moral, Yacine Jernite, Abhishek Thakur, Patrick von Platen, Suraj Patil, Julien Chaumond, Mariama Drame, Julien Plu, Lewis Tunstall, Joe Davison, Mario {\v{S}}a{\v{s}}ko, Gunjan Chhablani, Bhavitvya Malik, Simon Brandeis, Teven Le~Scao, Victor Sanh, Canwen Xu, Nicolas Patry, Angelina McMillan-Major, Philipp Schmid, Sylvain Gugger, Cl{\'e}ment Delangue, Th{\'e}o Matussi{\`e}re, Lysandre Debut, Stas Bekman, Pierric Cistac, Thibault Goehringer, Victor Mustar, Fran{\c{c}}ois Lagunas, Alexander Rush, and Thomas Wolf.
\newblock Datasets: A community library for natural language processing.
\newblock In {\em Proceedings of the 2021 Conference on Empirical Methods in Natural Language Processing: System Demonstrations}, pages 175--184, Online and Punta Cana, Dominican Republic, November 2021. Association for Computational Linguistics.

\bibitem{cherti2023reproducible}
Mehdi Cherti, Romain Beaumont, Ross Wightman, Mitchell Wortsman, Gabriel Ilharco, Cade Gordon, Christoph Schuhmann, Ludwig Schmidt, and Jenia Jitsev.
\newblock Reproducible scaling laws for contrastive language-image learning.
\newblock In {\em Proceedings of the IEEE/CVF Conference on Computer Vision and Pattern Recognition}, pages 2818--2829, 2023.

\bibitem{wang2023cogvlm}
Weihan Wang, Qingsong Lv, Wenmeng Yu, Wenyi Hong, Ji~Qi, Yan Wang, Junhui Ji, Zhuoyi Yang, Lei Zhao, Xixuan Song, et~al.
\newblock Cogvlm: Visual expert for pretrained language models.
\newblock {\em arXiv preprint arXiv:2311.03079}, 2023.

\bibitem{team2024gemma}
Gemma Team, Thomas Mesnard, Cassidy Hardin, Robert Dadashi, Surya Bhupatiraju, Shreya Pathak, Laurent Sifre, Morgane Rivi{\`e}re, Mihir~Sanjay Kale, Juliette Love, et~al.
\newblock Gemma: Open models based on gemini research and technology.
\newblock {\em arXiv preprint arXiv:2403.08295}, 2024.

\bibitem{ikezogwo2024quilt}
Wisdom Ikezogwo, Saygin Seyfioglu, Fatemeh Ghezloo, Dylan Geva, Fatwir Sheikh~Mohammed, Pavan~Kumar Anand, Ranjay Krishna, and Linda Shapiro.
\newblock Quilt-1m: One million image-text pairs for histopathology.
\newblock {\em Advances in Neural Information Processing Systems}, 36, 2024.

\bibitem{lu2023towards}
Ming~Y Lu, Bowen Chen, Drew~FK Williamson, Richard~J Chen, Ivy Liang, Tong Ding, Guillaume Jaume, Igor Odintsov, Andrew Zhang, Long~Phi Le, et~al.
\newblock Towards a visual-language foundation model for computational pathology.
\newblock {\em arXiv preprint arXiv:2307.12914}, 2023.

\bibitem{efron1994introduction}
Bradley Efron and Robert~J Tibshirani.
\newblock {\em An introduction to the bootstrap}.
\newblock Chapman and Hall/CRC, 1994.

\bibitem{gupta2022grit}
Tanmay Gupta, Ryan Marten, Aniruddha Kembhavi, and Derek Hoiem.
\newblock Grit: General robust image task benchmark.
\newblock {\em arXiv preprint arXiv:2204.13653}, 2022.

\bibitem{wortsman2022robust}
Mitchell Wortsman, Gabriel Ilharco, Jong~Wook Kim, Mike Li, Simon Kornblith, Rebecca Roelofs, Raphael~Gontijo Lopes, Hannaneh Hajishirzi, Ali Farhadi, Hongseok Namkoong, et~al.
\newblock Robust fine-tuning of zero-shot models.
\newblock In {\em Proceedings of the IEEE/CVF conference on computer vision and pattern recognition}, pages 7959--7971, 2022.

\bibitem{Acevedo2020-ja}
Andrea Acevedo, Anna Merino, Santiago Alf{\'e}rez, {\'A}ngel Molina, Laura Bold{\'u}, and Jos{\'e} Rodellar.
\newblock A dataset of microscopic peripheral blood cell images for development of automatic recognition systems.
\newblock {\em Data Brief}, 30(105474):105474, June 2020.

\bibitem{Burgess2024-lg}
James Burgess, Jeffrey~J Nirschl, Maria-Clara Zanellati, Alejandro Lozano, Sarah Cohen, and Serena Yeung-Levy.
\newblock Orientation-invariant autoencoders learn robust representations for shape profiling of cells and organelles.
\newblock {\em Nat. Commun.}, 15(1), February 2024.

\bibitem{Wu2010-dr}
Yong Wu, Mansoureh Eghbali, Jimmy Ou, Rong Lu, Ligia Toro, and Enrico Stefani.
\newblock Quantitative determination of spatial protein-protein correlations in fluorescence confocal microscopy.
\newblock {\em Biophys. J.}, 98(3):493--504, February 2010.

\bibitem{iudin2023empiar}
Andrii Iudin, Paul~K Korir, Sriram Somasundharam, Simone Weyand, Cesare Cattavitello, Neli Fonseca, Osman Salih, Gerard~J Kleywegt, and Ardan Patwardhan.
\newblock Empiar: the electron microscopy public image archive.
\newblock {\em Nucleic Acids Research}, 51(D1):D1503--D1511, 2023.

\bibitem{eulenberg2017reconstructing}
Philipp Eulenberg, Niklas K{\"o}hler, Thomas Blasi, Andrew Filby, Anne~E Carpenter, Paul Rees, Fabian~J Theis, and F~Alexander Wolf.
\newblock Reconstructing cell cycle and disease progression using deep learning.
\newblock {\em Nature communications}, 8(1):463, 2017.

\bibitem{Held2010-zb}
Michael Held, Michael H~A Schmitz, Bernd Fischer, Thomas Walter, Beate Neumann, Michael~H Olma, Matthias Peter, Jan Ellenberg, and Daniel~W Gerlich.
\newblock {CellCognition}: time-resolved phenotype annotation in high-throughput live cell imaging.
\newblock {\em Nat. Methods}, 7(9):747--754, September 2010.

\bibitem{Hussain2020-vq}
Elima Hussain, Lipi~B Mahanta, Himakshi Borah, and Chandana~Ray Das.
\newblock Liquid based-cytology pap smear dataset for automated multi-class diagnosis of pre-cancerous and cervical cancer lesions.
\newblock {\em Data Brief}, 30(105589):105589, June 2020.

\bibitem{Battiato2020POLLEN13KAL}
Sebastiano Battiato, Alessandro Ortis, Francesca Trenta, Lorenzo Ascari, Mara Politi, and Consolata Siniscalco.
\newblock Pollen13k: A large scale microscope pollen grain image dataset.
\newblock {\em 2020 IEEE International Conference on Image Processing (ICIP)}, pages 2456--2460, 2020.

\bibitem{jung2022wbc}
Changhun Jung, Mohammed Abuhamad, David Mohaisen, Kyungja Han, and DaeHun Nyang.
\newblock Wbc image classification and generative models based on convolutional neural network.
\newblock {\em BMC Medical Imaging}, 22(1):94, 2022.

\bibitem{Kather2016-db}
Jakob~Nikolas Kather, Cleo-Aron Weis, Francesco Bianconi, Susanne~M Melchers, Lothar~R Schad, Timo Gaiser, Alexander Marx, and Frank~Gerrit Z{\"o}llner.
\newblock Multi-class texture analysis in colorectal cancer histology.
\newblock {\em Sci. Rep.}, 6:27988, June 2016.

\bibitem{Kather2019-vo}
Jakob~Nikolas Kather, Johannes Krisam, Pornpimol Charoentong, Tom Luedde, Esther Herpel, Cleo-Aron Weis, Timo Gaiser, Alexander Marx, Nektarios~A Valous, Dyke Ferber, Lina Jansen, Constantino~Carlos Reyes-Aldasoro, Inka Z{\"o}rnig, Dirk J{\"a}ger, Hermann Brenner, Jenny Chang-Claude, Michael Hoffmeister, and Niels Halama.
\newblock Predicting survival from colorectal cancer histology slides using deep learning: A retrospective multicenter study.
\newblock {\em PLoS Med.}, 16(1):e1002730, January 2019.

\bibitem{Nirschl2018-pc}
Jeffrey~J Nirschl, Andrew Janowczyk, Eliot~G Peyster, Renee Frank, Kenneth~B Margulies, Michael~D Feldman, and Anant Madabhushi.
\newblock A deep-learning classifier identifies patients with clinical heart failure using whole-slide images of h\&e tissue.
\newblock {\em PLoS One}, 13(4):e0192726, April 2018.

\bibitem{Cho2022-um}
Nathan~H Cho, Keith~C Cheveralls, Andreas-David Brunner, Kibeom Kim, Andr{\'e}~C Michaelis, Preethi Raghavan, Hirofumi Kobayashi, Laura Savy, Jason~Y Li, Hera Canaj, James Y~S Kim, Edna~M Stewart, Christian Gnann, Frank McCarthy, Joana~P Cabrera, Rachel~M Brunetti, Bryant~B Chhun, Greg Dingle, Marco~Y Hein, Bo~Huang, Shalin~B Mehta, Jonathan~S Weissman, Rafael G{\'o}mez-Sj{\"o}berg, Daniel~N Itzhak, Lo{\"\i}c~A Royer, Matthias Mann, and Manuel~D Leonetti.
\newblock {OpenCell}: Endogenous tagging for the cartography of human cellular organization.
\newblock {\em Science}, 375(6585):eabi6983, March 2022.

\bibitem{sirinukunwattana2016gland}
Korsuk Sirinukunwattana, Josien P.~W. Pluim, Hao Chen, Xiaojuan Qi, PhengAnn Heng, Yun~Bo Guo, Li~Yang Wang, Bogdan~J. Matuszewski, Elia Bruni, Urko Sanchez, Anton B¨ohm, Olaf Ronneberger, Bassem~Ben Cheikh, Daniel Racoceanu, Philipp Kainz, Michael Pfeiffer, Martin Urschler, David R.~J. Snead, and Nasir~M. Rajpoot.
\newblock Gland segmentation in colon histology images: The glas challenge contest, 2016.

\bibitem{Tang2019-fy}
Ziqi Tang, Kangway~V Chuang, Charles DeCarli, Lee-Way Jin, Laurel Beckett, Michael~J Keiser, and Brittany~N Dugger.
\newblock Interpretable classification of alzheimer's disease pathologies with a convolutional neural network pipeline.
\newblock {\em Nat. Commun.}, 10(1):2173, May 2019.

\bibitem{Wong2022-nm}
Daniel~R Wong, Ziqi Tang, Nicholas~C Mew, Sakshi Das, Justin Athey, Kirsty~E McAleese, Julia~K Kofler, Margaret~E Flanagan, Ewa Borys, Charles~L White, 3rd, Atul~J Butte, Brittany~N Dugger, and Michael~J Keiser.
\newblock Deep learning from multiple experts improves identification of amyloid neuropathologies.
\newblock {\em Acta Neuropathol. Commun.}, 10(1):66, April 2022.

\bibitem{Wu2023-ht}
Gong-Her Wu, Charlene Smith-Geater, Jes{\'u}s~G Galaz-Montoya, Yingli Gu, Sanket~R Gupte, Ranen Aviner, Patrick~G Mitchell, Joy Hsu, Ricardo Miramontes, Keona~Q Wang, Nicolette~R Geller, Cathy Hou, Cristina Danita, Lydia-Marie Joubert, Michael~F Schmid, Serena Yeung, Judith Frydman, William Mobley, Chengbiao Wu, Leslie~M Thompson, and Wah Chiu.
\newblock {CryoET} reveals organelle phenotypes in huntington disease patient {iPSC-derived} and mouse primary neurons.
\newblock {\em Nat. Commun.}, 14(1):692, February 2023.

\bibitem{hasan2018overview}
Sadid~A Hasan, Yuan Ling, Oladimeji Farri, Joey Liu, Henning M{\"u}ller, and Matthew~P Lungren.
\newblock Overview of imageclef 2018 medical domain visual question answering task.
\newblock In {\em CLEF (Working Notes)}, 2018.

\bibitem{abacha2019vqa}
Asma~Ben Abacha, Sadid~A Hasan, Vivek~V Datla, Joey Liu, Dina Demner-Fushman, and Henning M{\"u}ller.
\newblock Vqa-med: Overview of the medical visual question answering task at imageclef 2019.
\newblock {\em CLEF (working notes)}, 2(6), 2019.

\bibitem{ben2021overview}
Asma Ben~Abacha, Mourad Sarrouti, Dina Demner-Fushman, Sadid~A Hasan, and Henning M{\"u}ller.
\newblock Overview of the vqa-med task at imageclef 2021: Visual question answering and generation in the medical domain.
\newblock In {\em Proceedings of the CLEF 2021 Conference and Labs of the Evaluation Forum-working notes}. 21-24 September 2021, 2021.

\bibitem{sitara2021ssn}
Noor Mohamed~Sheerin Sitara and Kavitha Srinivasan.
\newblock Ssn mlrg at vqa-med 2021: An approach for vqa to solve abnormality related queries using improved datasets.
\newblock In {\em CLEF (working notes)}, pages 1329--1335, 2021.

\bibitem{lau2018dataset}
Jason~J Lau, Soumya Gayen, Asma Ben~Abacha, and Dina Demner-Fushman.
\newblock A dataset of clinically generated visual questions and answers about radiology images.
\newblock {\em Scientific data}, 5(1):1--10, 2018.

\bibitem{kovaleva2020towards}
Olga Kovaleva, Chaitanya Shivade, Satyananda Kashyap, Karina Kanjaria, Joy Wu, Deddeh Ballah, Adam Coy, Alexandros Karargyris, Yufan Guo, David~Beymer Beymer, et~al.
\newblock Towards visual dialog for radiology.
\newblock In {\em Proceedings of the 19th SIGBioMed workshop on biomedical language processing}, pages 60--69, 2020.

\bibitem{huang2022ovqa}
Yefan Huang, Xiaoli Wang, Feiyan Liu, and Guofeng Huang.
\newblock Ovqa: A clinically generated visual question answering dataset.
\newblock In {\em Proceedings of the 45th International ACM SIGIR Conference on Research and Development in Information Retrieval}, pages 2924--2938, 2022.

\bibitem{liu2021slake}
Bo~Liu, Li-Ming Zhan, Li~Xu, Lin Ma, Yan Yang, and Xiao-Ming Wu.
\newblock Slake: A semantically-labeled knowledge-enhanced dataset for medical visual question answering.
\newblock In {\em 2021 IEEE 18th International Symposium on Biomedical Imaging (ISBI)}, pages 1650--1654. IEEE, 2021.

\bibitem{zhang2023pmc}
Xiaoman Zhang, Chaoyi Wu, Ziheng Zhao, Weixiong Lin, Ya~Zhang, Yanfeng Wang, and Weidi Xie.
\newblock Pmc-vqa: Visual instruction tuning for medical visual question answering.
\newblock {\em arXiv preprint arXiv:2305.10415}, 2023.

\bibitem{li2022blip}
Junnan Li, Dongxu Li, Caiming Xiong, and Steven Hoi.
\newblock Blip: Bootstrapping language-image pre-training for unified vision-language understanding and generation.
\newblock In {\em International conference on machine learning}, pages 12888--12900. PMLR, 2022.

\bibitem{chakraborty2020biomedbert}
Souradip Chakraborty, Ekaba Bisong, Shweta Bhatt, Thomas Wagner, Riley Elliott, and Francesco Mosconi.
\newblock Biomedbert: A pre-trained biomedical language model for qa and ir.
\newblock In {\em Proceedings of the 28th international conference on computational linguistics}, pages 669--679, 2020.

\end{thebibliography}

%%%%%%%%%%%%%%%%%%%%%%%%%%%%%%%%%%%
%%%%%%%%%%% Appendix %%%%%%%%%%%%%%
%%%%%%%%%%%%%%%%%%%%%%%%%%%%%%%%%%%
\newpage
\section{Appendix}
\appendix

\section{Limitations}
\label{sec:limitations}
\dataset\ is a diverse benchmark for evaluating vision-language models on microscopy image and text data. The dataset is intended for testing purposes only, not for training. The moderate size will allow many academic researchers to assess the performance of models trained on natural images or other biology datasets.

While \dataset\ covers various biological length scales, microscopy modalities, scientific disciplines, and organisms, not all domains and modalities are equally represented. This partially reflects the usage patterns of the field, with human samples (cell lines or tissues) being more common in biomedical research. Brightfield and fluorescence microscopy images are also more prevalent in \dataset\ compared to electron microscopy. This means that the results on \dataset\ may not apply equally well to all model organisms or electron microscopy images. The VLM performance may be lower in these areas due to the data being rare in both natural image datasets and uncommon in biomedical datasets.

We strive for a high-quality dataset and involve cell biologists and pathologists during dataset creation, quality control, and interpretation of results. However, the field of biology is diverse, and no individual or small group can reliably stay up-to-date with all aspects of biological/biomedical research. For example, we included a botany dataset (ICPR 2020 Pollen) to show a commitment to including diverse scientific disciplines, but we currently do not have expert plant biologist contributors.

\dataset\ represents the first vision-language benchmark for microscopy covering all major biology length scales and modalities. We see \dataset\ as a living benchmark we intend to grow, although we acknowledge the aforementioned limitations. Future versions will prioritize incorporating new data from diverse organisms and microscopy modalities to improve representation across all length scales, microscopy modalities, and organisms. We will also benefit from community engagement by involving domain experts from diverse fields and obtaining data to balance currently under-represented areas.

\section{Ethical Compliance and Acknowledgements}
\subsection{Ethics Statement}
\label{sec:ethics}
\textbf{Safe and ethical use of biomedical data:} We prioritize safe and ethical research practices while creating \dataset\ . All public datasets with patient-derived histopathology images had already been de-identified by the dataset's original authors in compliance with applicable privacy laws and institutional guidelines. The public histopathology image data and metadata were reviewed, and it was determined that it was not possible to identify any individual from the de-identified data. The Stanford Institutional Review Board \href{https://researchcompliance.stanford.edu/panels/hs/for-all-researchers}{guidelines} were reviewed and discussed, and the use of the images was determined not to be human subjects research. 

The \dataset\ cognition images and questions were voluntarily submitted by a small (<10) number of biology/pathology users alpha-testing a free web chat application. There was no intervention, experiment vs. control group, or research question during the alpha testing. There was no greater risk of using the app than other internet apps. At registration, users agreed to the service terms, which included releasing image and text data under CC-BY-SA-4.0. 

\textbf{Consent and data usage:} We thank and respect the original dataset authors and use data according to the original copyright and license. \dataset\ was developed with both academic and commercial research in mind. Many datasets are a version of CC-BY-4.0 to allow both academic and commercial usage. However, some data restricts commercial applications via non-commercial clauses or CC-BY-NC-4.0-related licenses. While creating \dataset\, we significantly improved the data by performing expert review, quality control/standardization, expert labeling with biomedical ontology codes, and creating multiple-choice VQA questions and captions for each image. When the original dataset license is CC-BY-4.0, we release our \dataset\ versions under a permissive CC-BY-SA-4.0 to foster a transparent and collaborative benchmark. For the subset with CC-BY-NC-SA-4.0, we respect the original license and release these data under CC-BY-NC-SA-4.0.

\textbf{Bias and data diversity:}  We recognize that AI models, including VLMs, can perpetuate or exacerbate biases in the training data, and incorporating diverse data may mitigate these biases. Diversity is key to understanding biological processes and how they vary across biological sex, ethnicity, or other factors. When possible, we annotate cell lines with age, sex, ethnicity, and other metadata from Cellosaurus or the Cell Line Ontology. We consciously include diverse microscopy images from multiple institutions across various organisms, modalities, and biological states to ensure \dataset\ provides an accurate and fair performance assessment. However, images of human samples and brightfield/fluorescence microscopy are more common in the field and thus over-represented in \dataset\ .

\textbf{Potential negative societal impacts:}  AI models trained on biomedical data have the potential for far-reaching impacts on society, both positive and negative. Potential negative impacts include biased performance across different demographic groups and reinforcing existing disparities in research and healthcare. We are committed to mitigating these risks by ensuring the dataset's diversity and continuously reviewing the ethical implications of our work. Additionally, we will engage with the broader research community to identify and address any emerging ethical concerns.

We are committed to ongoing improvements of \dataset\ to prioritize diverse and representative microscopy data. We will review and update \dataset\ in response to evolving ethical standards and technological advancements.

\subsection{Conflicts of Interest Disclosures}

The authors have no conflicts of interest to disclose.

\subsection{Funding/Support}

This research was supported by NIH grants (NIH\#P30AG066515 to JJN), the Chan Zuckerberg Initiative Neurodegeneration Challenge Pairs Pilot Project to SYL (2020-221724, 5022), the Wu Tsai Knight Initiative Innovation grant (\#KIG 102) to SYL, Arc Institute to AL, Quad Fellowship to JB, and NSF Graduate Research Fellowship (\#DGE-2146755) and Stanford Graduate Fellowship to AU. SYL is a Chan Zuckerberg Biohub – San Francisco Investigator.

\subsection{Acknowledgments}

We thank all the domain experts for testing the web app and submitting questions that were eventually used in \dataset\ cognition. We highlight the contributions from Pedro Guedes-Dias, Andrew Moore, and Julian Perez. We appreciate feedback from Josiah Aklilu and Orr Zohar on earlier manuscript versions.

\section{Benchmark Details}
\label{sec:appendix-benchmark_details}
\subsection{Instructions for downloading the benchmark}
The benchmark can be downloaded via HuggingFace Datasets at the following \href{https://huggingface.co/datasets/jnirschl/uBench}{url}.

\subsection{\dataset\ overview}

\dataset\ is organized into three main categories:
\begin{enumerate}
    \item \textbf{\dataset\ Perception (Coarse-grained)}: Containing basic questions about the type of biomedical field of study (domain), subdomain, microscopy modality, submodality, and stain.
    \item  \textbf{\dataset\ Perception (Fine-grained)}: Identification or questions regarding a biological cell, cellular process, subcellular or tissue structures, biological state (normal/abnormal), etc.
    \item \textbf{\dataset\ Cognition (Reasoning)}: Expert-generated questions that typically require visual-based reasoning or integrating knowledge about the micrograph's composition and subject to deduce an answer. 
    \item \textbf{\dataset\ Object Detection (Localization)}: Bounding box detection of common biological objects, with an easy and hard data split.
\end{enumerate}

\subsection{\dataset\ statistics}
Table \ref{tbl:uBenchDatasetStatistics} shows descriptive statistics for \dataset\ Perception while  Table \ref{tbl:uBenchCognitionDatasetStatistics} shows statistics for \dataset\ Reasoning.

\subsection{\dataset\ Perception}
\dataset\ Perception contains 17,235 microscopy images collected from  25 unique datasets of 96 different cell and tissue types across light, fluorescence, and electron microscopy. If defined, only images from each dataset's test set are selected; otherwise, a random 15 percent split is created. Each cell/tissue class is sub-sampled to a maximum of 200 sub-classes per class (if the cell type is less than the maximum, the full set is used). Each cell/tissue type is densely annotated (then propagated to each instance of the same type)  by an expert, as shown in Figure \ref{fig:data_point}

Table \ref{tbl:uBenchDomainStatistics} provides summary statistics of domain coverage. Overall, the benchmark covers 8,637 biology images and 8,678 pathology images across 12 subdomains. Similarly, Table \ref{tbl:uBenchModalityStatistics} shows summary statistics of microscopy modalities covered by \dataset\ perception, including 10,864 images for light microscopy, 5,618 for fluorescence microscopy, and 833 images for electron microscopy across 8 microscopy imaging submodalities.

% \subsubsection{\dataset\ Perception (Coarse-grained)}
\textbf{\dataset\ Perception (Coarse-grained)}:
Hierarchical metadata for each of the 17,235 perception images and task-specific templates (shown in Table \ref{tbl:template_coarse_grained}) are used to create 5 coarse-grained questions and captions regarding microscopy modality, submodality, domain, subdomain, and staining technique. The use of hierarchical metadata enables the generation of options within each hierarchical level. For example, for microscopy submodality, we leveraged microscopy modality metadata (shown in Figure\ref{fig:metadata_microscopy}  ) to randomly sample submodality options within a microscopy modality  (e.g. differential interference contrast microscopy within light microscopy). A total of 86,175 (17,235x5) coarse-grained questions are generated using this approach, leveraging domain metadata Figure \ref{fig:metadata_domain} as well as staining metadata  Figure \ref{fig:metadata_stain}). Section \ref{perception_coarse_grained_data_examples} provides 10 random examples of coarse-grained data points (two per type).

\begin{table}[h!]
    \centering
    \caption{\dataset\ Perception dataset statistics summary}
    \begin{tabular}{lr}
    \toprule

    \textbf{Aspect} & \textbf{Count} \\
    \midrule

    Datasets & 25 \\
    Domains & 2 \\
    Subdomains & 12 \\
    Modalities & 3 \\
    Submodalities & 8 \\
    Coarse-grained perception tasks & 5 \\
    Fine-grained perception tasks & 18 \\
    Classification tasks & 13 \\
    Segmentation and Object Detection tasks & 5 \\

    \bottomrule
    \end{tabular}
\label{tbl:uBenchDatasetStatistics}
\end{table}

% \subsubsection{\dataset\ Perception (Fine-grained)}
\textbf{\dataset\ Perception (Fine-grained)}:
Metadata for each of the  25 unique datasets and custom prompts (Table \ref{tbl:uBenchDatasetStatistics}) are used to generate 13 unique tasks, which are 13 (17,235 unique question-image pairs) closed VQA style questions.
% and 5 (3641 images) segmentation/detection tasks Table \ref{tbl:uBenchFineGrainedTaskStatistics} provides a breakdown of the number of images aggregated by class type. Section \ref{perception_fine_grained_data_examples}} shows random examples of
of  VQA fine-grained tasks.

% \subsection{Perception dataset: source details}
\textbf{Perception dataset: source details}:
Table \ref{table:sources} lists the dataset names, licenses, and corresponding DOIs or URLs for dataset used. The majority of these datasets were sourced from open data repositories, such as Zenodo, Dataverse, Dryad, BBBC, EMPIAR, Kaggle, and various project websites.

\textbf{Perception dataset: forming questions for evaluation}
Each sample has a question and set of candidate answers. We describe how to evaluate this VQA task in \cref{sec:appendix_evaldetails}.

\begin{table}[]
\centering
\caption{Provenance of the datasets for the \dataset\ Perception dataset. The dataset name is provided along with the original license and URL to the dataset (or DOI where available).}
\begin{tabular}{@{}p{6cm}p{4cm}p{5cm}@{}}
\toprule
\textbf{Dataset} & \textbf{License}  & \textbf{Link}                                                            \\ \midrule
Acevedo et al 2020 \cite{Acevedo2020-ja}  & CC-BY-4.0                                    & \href{https://doi.org/10.17632/snkd93bnjr.1}{DOI}                                                                                                                                                                                    \\
PCST-Contour \cite{Burgess2024-lg}                                  & CC-BY-4.0 & \href{https://doi.org/10.5281/zenodo.7388245}{DOI}                                                                                                                                                                                   \\
PCST-Eccentricity \cite{Burgess2024-lg}                   & CC-BY-4.0        & \href{https://doi.org/10.5281/zenodo.7388245}{DOI}                                                                                                                                                                                   \\
PCST-Texture \cite{Burgess2024-lg}       & CC-BY-4.0                             & \href{https://doi.org/10.5281/zenodo.7388245}{DOI}                                                                                                                                                                                   \\
Colocalization benchmark \cite{Wu2010-dr}        & CC-BY-NC-SA-3.0                 & \href{https://doi.org/10.1016/j.bpj.2009.10.037}{DOI}                                                                                                                                                                                \\
EMPIAR SBF-SEM \cite{iudin2023empiar}  & CC0 &\href{https://www.ebi.ac.uk/empiar/EMPIAR-10127/}{URL},
\href{https://www.ebi.ac.uk/empiar/EMPIAR-10994/}{URL},
\href{https://www.ebi.ac.uk/empiar/EMPIAR-11464/}{URL},
\href{https://www.ebi.ac.uk/empiar/EMPIAR-11831/}{URL},
\href{https://www.ebi.ac.uk/empiar/EMPIAR-11759/}{URL}  \\

BBBC048 (Brightfield) \cite{eulenberg2017reconstructing}    & CC-BY-NC-SA-3.0        & \href{https://data.broadinstitute.org/bbbc/BBBC048}{URL}   \\
BBBC048 (Darkfield) \cite{eulenberg2017reconstructing}    & CC-BY-NC-SA-3.0        & \href{https://data.broadinstitute.org/bbbc/BBBC048}{URL}                \\
BBBC048 (Epifluorescence) \cite{eulenberg2017reconstructing}    & CC-BY-NC-SA-3.0  & \href{https://data.broadinstitute.org/bbbc/BBBC048}{URL}                                                                                                                                                               \\
CellCognition (Golgi) \cite{Held2010-zb}                     & Attribution  &\href{http://gerlichlab.imba.oeaw.ac.at/data/chromatin\_golgi\_aparatus/classifiers/GalT\_10x\_MD\_exp835.zip}{URL}                                                                                                                  \\
CellCognition (H2B)  \cite{Held2010-zb}  & Attribution                                 & \href{http://gerlichlab.imba.oeaw.ac.at/data/chromatin\_microtubles/classifiers/H2b\_20x\_MD\_exp911.zip}{URL}                                                                                                                       \\
CellCognition (Mt)  \cite{Held2010-zb}      & Attribution                                 & \href{http://gerlichlab.imba.oeaw.ac.at/data/chromatin\_microtubles/classifiers/Tub\_20x\_MD\_exp911.zip}{URL}                                                                                                                       \\
Pap Smear 2019 \cite{Hussain2020-vq}   &   CC-BY-4.0                                  & \href{https://doi.org/10.17632/zddtpgzv63.4}{DOI}                                                                                                                                                                                   \\
ICPR2020 Pollen \cite{Battiato2020POLLEN13KAL}    & Non-commercial                                        & \href{http://dx.doi.org/10.1007/978-3-030-68793-9\_34}{DOI}                                                                                                                                                                         \\
Jung et al 2022 \cite{jung2022wbc}    & CC-BY-NC-4.0                                     & \href{https://drive.usercontent.google.com/download?id=1\_0NknbNGBMZ5aYyHzR5pyDKdp8QYYbNr\&export=download\&authuser=0\&confirm=t\&uuid=26579c04-9c9a-4b0c-b5ea-342bc6c1e29c\&at=APZUnTXsYclVxkIwqSoUnbhmev8r\%3A1713560403139}{URL} \\
Kather et al 2016 \cite{Kather2016-db}    & CC-BY-4.0                                  & \href{https://zenodo.org/record/53169\#.XGZemKwzbmG }{DOI}                                                                                                                                                                           \\
Kather et al 2018  \cite{Kather2019-vo}     & CC-BY-4.0                                      & \href{https://doi.org/10.5281/zenodo.1214456}{DOI}                                                                                                                                                                                   \\
Kather et al 2018 Val7K \cite{Kather2019-vo}     & CC-BY-4.0                     & \href{https://doi.org/10.5281/zenodo.1214456}{DOI}                                                                                                                                                                                   \\
Nirschl et al 2018 \cite{Nirschl2018-pc}     & CC-BY-4.0                                   & \href{https://doi.org/10.17867/10000113}{DOI}                                                                                                                                                                                        \\
OpenCell \cite{Cho2022-um}               & CC-BY-4.0                                               & \href{https://czb-opencell.s3.amazonaws.com/index.html}{URL}                                                                                                                                                           \\
GlaS Challenge 2015 \cite{sirinukunwattana2016gland}   & Non-commercial                & \href{https://www.kaggle.com/datasets/sani84/glasmiccai2015-gland-segmentation}{URL}                                                                                                                                                 \\
Tang et al 2019 \cite{Tang2019-fy}    & CC-BY-4.0                                       & \href{https://doi.org/10.5281/zenodo.1470797}{DOI}                                                                                                                                                                                   \\
Wong et al 2022 \cite{Wong2022-nm}   & CC-BY-4.0                                           & \href{https://doi.org/10.17605/osf.io/XH2JD}{DOI}                                                                                                                                                                                    \\
Wu et al 2023 \cite{Wu2023-ht}      & CC-BY-SA-4.0                                         & \href{https://doi.org/10.1038/s41467-023-36096-w}{DOI}  \\

Fluorescence  Cells \& Structures                  & CC-BY-4.0   &     New Data   
\\ \bottomrule
\label{table:sources}
\end{tabular}
\end{table}

\subsection{\dataset\ Cognition}
\textbf{Cognition dataset: generation details}

\label{crowsoucing}
The cognition dataset was collected during alpha-testing of a web application chat interface. A group of 6 biology and pathology experts were invited to interact with a web application as they wished during their daily routines (invitation is shown in Figure \ref{fig:invitation}). There was no specific research question, intervention, or experimental vs. control group; this was a free service. User registration was voluntary and required reviewing and accepting the terms of service. The terms of service discussed that this was not a research study and that the risk of harm is insignificant and would be similar to any VLM chatbot. The terms of service indicated that uploading images indicated the user had copyright or permission, that images did not contain offensive content, and that the images and text could be released with a CC-BY-SA-4.0 license.

\begin{figure}
    \centering
    \includegraphics[width=1\linewidth]{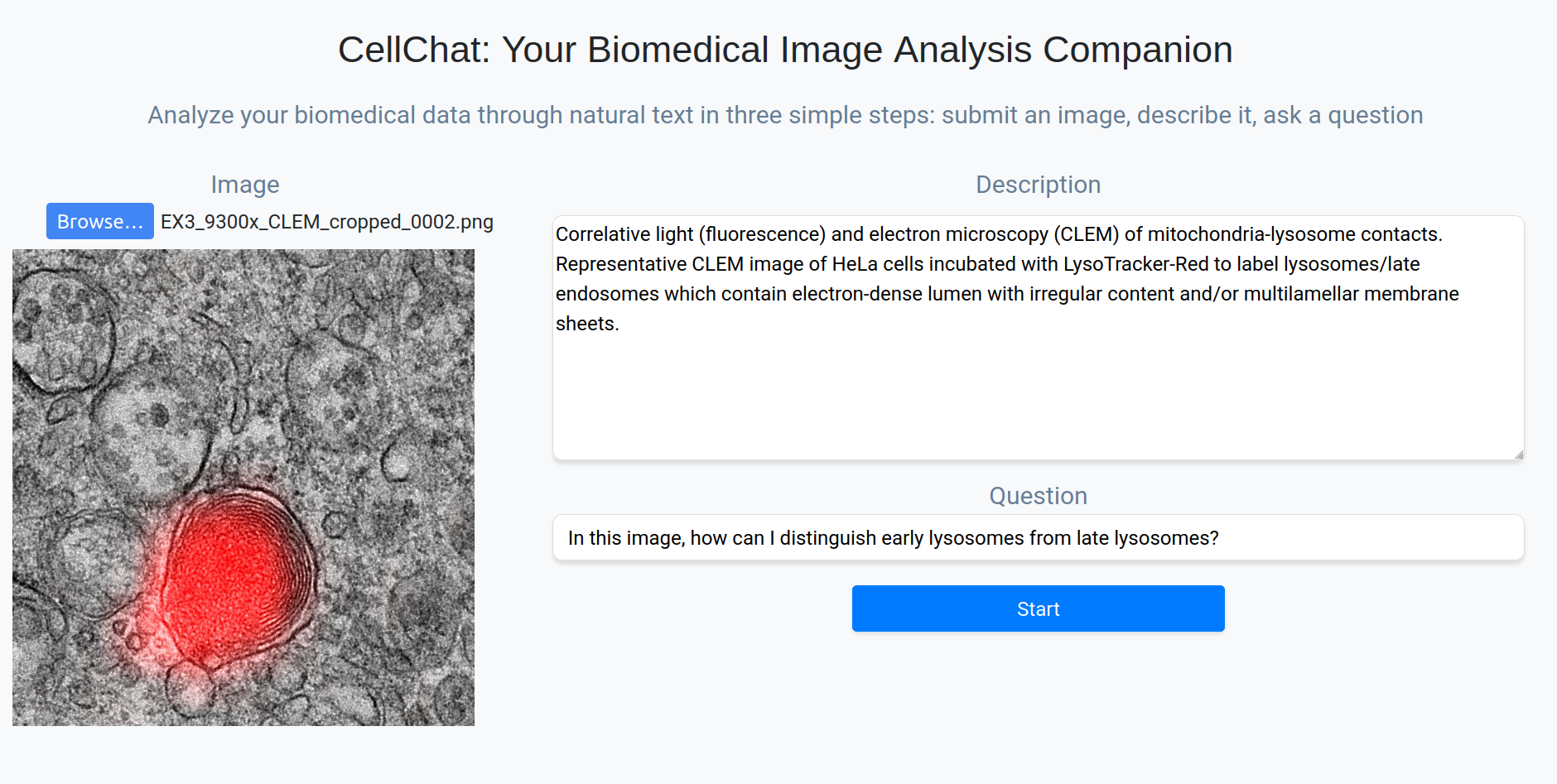}
    \caption{CellChat App: Interactive web app used to elicit  real-world open VQA examples  from expert microscopist.}
    \label{fig:webApp}
\end{figure}

Figure \ref{fig:webApp}) shows a screenshot of the web application interface. A typical usage involved uploading a microscopy image, providing context about the image (experiment details), and asking a question. The web app processed the submission using GPT-4V and provided an answer in real time. Users were encouraged to provide feedback on the answer and a correct answer (if known). Users were encouraged to ask questions that required complex visual reasoning, advanced biomedical knowledge, or could be considered challenging for humans. Random samples for cognition questions are shown in  Section \ref{cognition_data_examples}

\begin{table}[h!]
    \centering
    \caption{\dataset\ Cognition dataset statistics summary}
    \begin{tabular}{lr}
    \toprule

    \textbf{Aspect} & \textbf{Count} \\
    \midrule

    Submitters & 6\\
    Institutions & 5\\
    Domains & 2 \\
    Subdomains & 10 \\
    Modalities & 3 \\
    Submodalities & 12 \\
    \bottomrule
    \end{tabular}
\label{tbl:uBenchCognitionDatasetStatistics}
\end{table}

\subsection{\dataset\ Object detection}
For datasets with segmentation annotations, we copy the segmentation annotations and also convert them to object detection annotations, giving 3641 images \cite{Burgess2024-lg, Held2010-zb, Cho2022-um, Wu2023-ht, sirinukunwattana2016gland}.

We define an easy and hard split. The easy split contains the `cell' class from Burgess et al. \cite{Burgess2024-lg} dataset and `nucleus' class from the Held et al. \cite{Held2010-zb}. Here the image contains only the target object, meaning that simple foreground / background segmentation would work well.  The hard split contains the `nucleus' class from Burgess et al. \cite{Burgess2024-lg}, the `nucleus' class from opencell \cite{Cho2022-um}, the `mitochondria' class from Wu et al. \cite{Wu2023-ht}, and the `gland' class from  Sirinukunwattana et al. \cite{sirinukunwattana2016gland}. Here, the target object must be separated from surrounding visual information, and this is challenging.

\subsection{Comparison with a supervised linear model trained on DINOv2 Features}
Although some datasets have been used in the biomedical computer vision community \cite{Kather2016-db, Kather2019-vo}, others do not have well-established performance baselines. Thus, we evaluate a supervised linear model's baseline classification performance for the fine-grained and coarse-grained perception tasks. This establishes that the questions in our benchmark are solvable using image features. 

The images in \dataset\ represent the test subset of the original data. The original data were split into train, validation, and test subsets according to the published splits. If no published splits were available, which was the case for many previously unused datasets, we created train/val/test splits (0.7/0.1/0.2) stratifying samples by the class label.  We use DINOv2 S14 features (384 dims) because they are robust and can train strong linear models on many natural image classification datasets. We extract DinoV2-S14 features from the train/val subsets and train a logistic regression classifier to determine the performance of a linear classifier for these tasks. We use PCA dimensionality reduction (0.95\% var) and whitening on the features with Optuna for hyperparameter tuning. Baseline performance was determined using the best dummy classifier, predicting based on a random sampling of prior probabilities, the most common class, etc.

\begin{figure}
    \centering
    \includegraphics[width=0.85\linewidth]{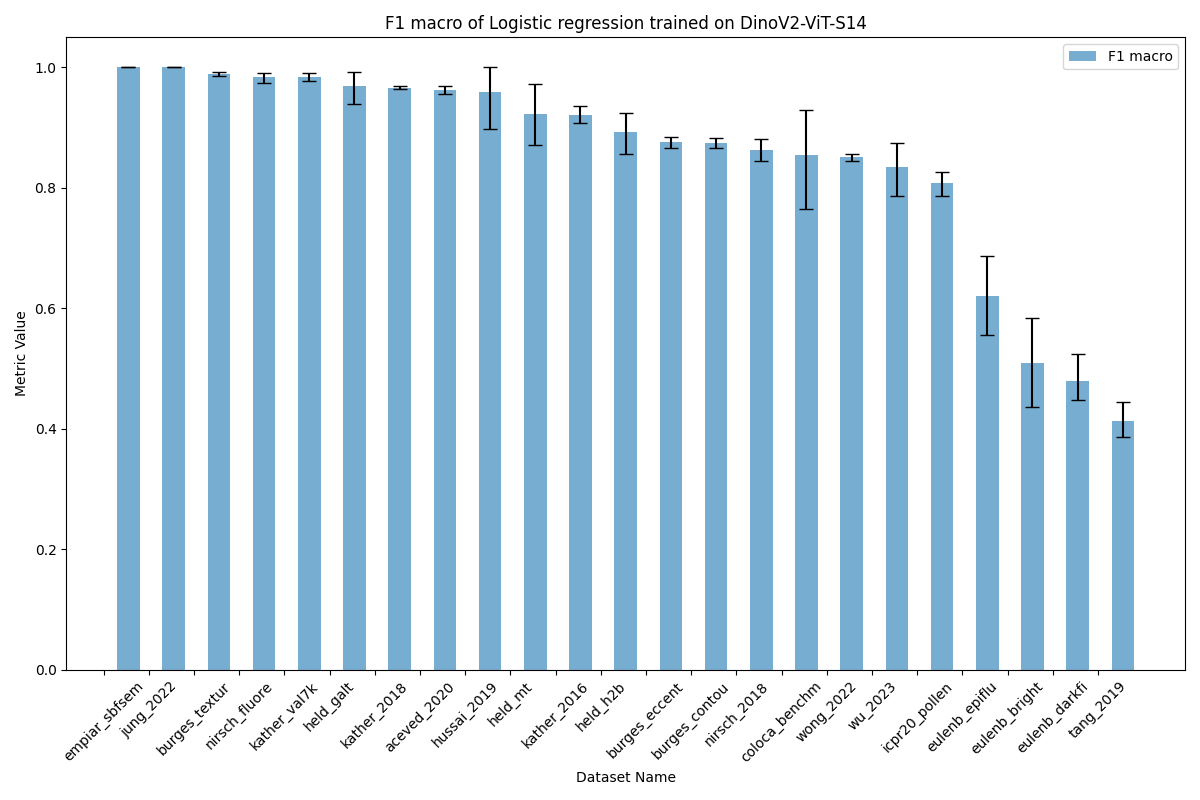}
    \caption{F1 Macro results for logistic regression classifier trained using DinoV2 features.}
    \label{fig:dinov2}
\end{figure}

Figure \ref{fig:dinov2} illustrates that most datasets provide sufficient signal to train a robust linear classifier with DinoV2 feature representations primarily learned from natural images. Our results show a weighted average accuracy of 0.86 across all classification tasks. The lowest performance was observed in the classification of mitosis stages in darkfield microscopy (50.96\%). In contrast, the highest performance was seen in the classification of synthetic white blood cells in bright microscopy and the organisms and structures in electron microscopy, with a weighted average accuracy of 100\%. Lastly, 58.33 \% of the tasks achieved a balanced accuracy greater than 80\%. These results show a weighted average of 86.71\% across all classification tasks with the upper and lower performance range we could expect from a VLM.

\label{sec:todo2} 

\section{Evaluation Details}
\label{sec:appendix_evaldetails}

\subsection{VQA evaluation details}
Three out of four tasks are formulated as multiple-choice visual question answering: coarse-grained perception, fine-grained perception, and cognition. We describe their evaluation here, while the fourth task of object detection is described separately in \cref{sec:appendix_detection}. For each sample, $i$, we have an image, $\textbf{x}_i$ a question string, $q$, and a set of $k$ candidate answer strings $\{a_{ij}\}_{j=1}^k$. 

For the autoregressive models, $f_A$, we first generate a query string, $t$, from the question and candidate answer strings, $t(q,\{a_{ij}\}_{j=1}^k)$, using the template in Figure \ref{fig:autoregressive_promtps}. The prompt text instructs the response to start with a letter for one of the multiple choice answers, (`A', `B', $\dots$). We pass the prompt string and image to the model and decode the output,  $y=f_A(x_i, t)$, where $y$ is the output string. We have two strategies for matching the response string, $y$, to the candidate answers $\{a_{ij}\}_{j=1}^k$. First, for each $j$, we check whether the answer string is in the output: $a_{ij}\in y$ with lowercase string matching. This is important because models do not always follow the instructions to output the multiple choice letter and instead return the answer. If there were no matches, then we extract the first character from $y$. If it is one of (`A', `B', $\dots$) -- as instructed in the prompt -- then that it is assigned to the corresponding answer, $a_{ij}$. Otherwise, we mark the answer as incorrect. 

The contrastive models have a vision encoder $E_x$ and text caption encoder $E_c$. We first compute the image embedding, $z_{x_i}=E_x(\textbf{x}_i)$. Then for each candidate answer, $a_{ij}$, we form a caption, $c$, as $c(a_{ij})$, using a text template that is suitable for CLIP-like models. The templates are in \cref{tbl:template_coarse_grained} for the fine-grained perception task, and \cref{tbl:template_coarse_grained_pathology} for the coarse-grained perception task. 
For the cognition task, each caption is the concatenation of the question string, $q$, and the candidate answer string, $a_{ij}$. 
We get the embedding for each caption, $z_{c_{ij}}=E_c(c_{ij})$ for $j\in[1,k]$, and then compute the cosine similarity score for each caption, $s_{ij}=z_{c_{ij}}\cdot z_{x_i}$ for $j\in[1,k]$. The $j$ with the largest $s_{ij}$ is the final prediction. If argmax($s_{ij}$) has the same index as the correct answer the question is marked as correct, incorrect otherwise.

Code for evaluation is made public through our repository: \href{https://github.com/Ale9806/eVLLM}{eVLLM}.

\subsection{Object detection evaluation details}
We evaluate the two models that support localization: QwenVL \cite{bai2023qwen} and PaliGemma \cite{team2024gemma}, and follow their user guide for prompting. For `QwenVL' the prompt is \texttt{``Detect \{class\_name\}''}. For PaliGemma the prompt is \texttt{``Detect \{class\_name\}; \{class\_name\}''}, where the repeated class name indicates that multiple instances may be predicted. Our early experiments found that PaliGemma would sometimes fail to localize any instances using detection prompting, but would localize them with segmentation prompting. So if PaliGemma does return zero instances, we prompt for segmentation \texttt{``Segment \{class\_name\}; \{class\_name\}''} and extract the bounding box. The Burgess et al. dataset has two classes, so we prompt the model one at a time. Both models output detection predictions as a string with a standardized structure, which we parse using regex. 

We use the GRIT localization metric \cite{gupta2022grit} because it is well-motivated and has previously been used in VLM evaluations by QwenVL \cite{bai2023qwen}. The score is:
\begin{align}
    \sum_{i=1}^M \frac{IoU_i}{P+G_{missed}}
\end{align}
There are $M$ ground truth boxes, and $P$ predicted boxes, which are matched using the Hungarian algorithm on the IoU metric. $G_{missed}$ is the number of predicted boxes not matched to a ground truth box. Intuitively, this metric measures the average IoU for matched boxes, while penalizing making too many predictions using $G_{missed}$ (similar to the precision metric). Note that we cannot use the more typical mAP score from object detection because they depend on a threshold for controlling the false-positive rate, which these VLMs do not support.

%%%%%%%%%%%%%%%%%%%%%%%%%%%%%%%%%%%%%
%%%%%%%%%%%%%% TABLE %%%%%%%%%%%%%%%%
%%%%%%%%%%%%%%%%%%%%%%%%%%%%%%%%%%%%%

\section{Additional Benchmarking Results}
\label{sec:appendix_benchmark}

\subsection{Object Detection Results}
\label{sec:appendix_detection}
\cref{tab:localization} summarizes object detection for the datasets having object localization annotations. 

\begin{table}[]
\centering
\caption{
Object detection results for all datasets with detection annotations and for all models that support object detection. 
}
\begin{tabular}{@{}llrrl@{}}
\toprule
\textbf{Dataset} & \textbf{Class }& \textbf{PaliGemma} & \textbf{QwenVLM}  &  \\ \midrule
\textbf{Easy} &  & \multicolumn{1}{l}{} & \multicolumn{1}{l}{} &  \\
PCST-Contour & cell & 76.5 & 82.1 &  \\
PCST-Eccentricity & cell & 76.6 & 84.7 &  \\
PCST-Texture & cell & 78.0 & 85.9 &  \\
CellCognition (Golgi) & nucleus & 72.4 & 46.7 &  \\
CellCognition (H2B) & nucleus & 80.4 & 68.1 &  \\
CellCognition (Mt)  & nucleus & 72.6 & 51.6 &  \\ \\
\textbf{Hard} &  & \multicolumn{1}{l}{} & \multicolumn{1}{l}{} &  \\
PCST-Contour & nucleus & 31.7 & 6.6 &  \\
PCST-Eccentricity & nucleus & 30.9 & 6.2 &  \\
PCST-Texture & nucleus & 30.2 & 6.4 &  \\
OpenCell & nucleus & 0.0 & 0.2 &  \\
Wu et al 2023  & mitochondria & 22.8 & 30.0 &  \\
GlaS Challenge & gland & 2.7 & 5.8 & 
\\ \bottomrule
\end{tabular}
\label{tab:localization}
\end{table}

Overall, the localization scores are very poor, which is expected since both models are generalist and object localization has received relatively less attention in autoregressive VLMs. Looking at the splits:
\begin{itemize}
    \item \textbf{Easy split.} Although both models have higher scores for the `cell' class in Burgess et al., they still fall below 80, and the task is extremely easy. The story is similar for `nucleus' in Held et al., but for QwenVLM, the scores are even lower.
    \item \textbf{Hard split.} All models perform poorly on the hard split. In Burgess et al `nucleus', PaliGemma scores around 30, however qualitative inspection shows that in most cases, the bounding box predicts the entire cell, which encapsulates the nucleus. Similarly, in Wu et al., the mitochondria class scores more than 20 for both models, but qualitative inspection shows that the prediction is usually a box around the entire image. We find the same pattern in `Sirinukunwattana et al.' for `gland' detection.
\end{itemize}

\subsection{Weight ensembling details}
In the results section \ref{sec:results-model_ensembling}, we consider PLIP, which was fine-tuned from OpenCLIP, and QuiltNet, which was fine-tuned from CLIP, both using pathology data. Since we have benchmark results for all these models, our evaluation can evaluate the impact of fine-tuning on pathology data (moving from OpenCLIP/CLIP to PLIP/QuiltNet). We showed that pathology fine-tuning can improve performance on the pathology subset of \dataset\ (the fine-grained perception), where we filter for all samples from the histopathology or H\&E imaging modality. However, the fine-tuned models have worse overall performance on \dataset 
 (which includes the pathology subset). 

We proposed to create `merged models' by combining the base model (OpenCLIP or CLIP) with their fine-tuned models (PLIP or QuiltNet) with weight merging. Specifically, following \cite{wortsman2022robust}, for a base model with weights $\theta_B$ and tuned model $\theta_T$ (which have the same architecture), the merged model weights $\theta_M$ are :
\begin{align*}
    \theta_M = \alpha \cdot \theta_B + (1-\alpha) \cdot \theta_T
\end{align*}
That is, we are linearly interpolating each model weight independently, with a single fixed constant $\alpha$. We arbitrarily set $\alpha=0.5$ in our experiments, but tuning that constant could lead to better overall results. 

We now show more comprehensive results in table \ref{tbl:uBenchPerformance}, which is the main results table but includes our merged models, M-PLIP and M-QuiltNet; table \ref{tbl:uBenchPerformance_hande} is the same but with the pathology-only split. In all cases (M-PLIP and M-QuiltNet), the merged models outperform the tuned models (PLIP and QuiltNet), while also outperforming the base models (OpenCLIP and CLIP) in almost all cases. For fine-grained perception (the task that is most in distribution for PLIP and QuiltNet training), the merged models become among the strongest performing overall models, and on pathology-specific fine-grained perception, the merged models outperform BiomedCLIP. 

 \begin{table}[h]
    \centering
    \caption{ Macro-average accuracy (with bootstrap confidence interval) for coarse-grained and fine-grained perception and cognition  (reasoning) in \dataset\ . Robust models (byproduct of merging fine-tuned models with their respective base models) are also included.}
    \begin{adjustbox}{width=\textwidth}
    \begin{tabular}{c c c c c c}
        \toprule 
        \rowcolor{gray!10} \multicolumn{6}{c}{$\mu$-Bench} \\
        \midrule
        \midrule
        \multicolumn{2}{c}{Perception (Coarse-Grained)} & \multicolumn{2}{c}{ Perception (Fine-Grained)} & \multicolumn{2}{c}{Cognition (Reasoning) } \\
        \midrule
        Model & Accuracy ($\pm$ CI) & Model & Accuracy ($\pm$ CI) & Model & Accuracy ($\pm$ CI) \\
        \midrule 
        \midrule
        \GPT{62.68}{0.35} & \GPT{51.73}{0.82} & \GPT{62.00}{9.00} \\
        \COGVLM{52.05}{0.35} & \BIOMEDCLIP{34.65}{0.75} & \QWENVLM{41.00}{10.00} \\
        \QWENVLM{49.85}{0.35} & \CONCH{33.64}{0.72} & \COGVLM{41.00}{10.00} \\
        \BIOMEDCLIP{47.57}{0.34} & \RPLIP{32.99}{0.73} & \OPENCLIP{38.33}{8.33} \\
        \RPLIP{43.25}{0.34} &\RQUILTCLIP{32.42}{0.71}& \RPLIP{34.17}{8.33} \\
        \ALIGN{40.7}{0.34} & \ALIGN{31.9}{0.72} & \ALIGN{31.00}{9.00} \\
        \OPENCLIP{36.34}{0.33} & \CLIP{30.09}{0.71} & \CLIP{28.00}{9.00} \\
        \PALIGEMMA{36.29}{0.33} &\OPENCLIP{29.36}{0.69}  & \RQUILTCLIP{25.83}{7.52} \\
        \CLIP{35.41}{0.34} & \COGVLM{28.18}{0.70} & \BIOMEDCLIP{25.00}{8.00} \\
         \RQUILTCLIP{31.26}{0.32} & \QUILTCLIP{27.85}{0.69}& \PALIGEMMA{25.00}{8.00} \\
        \PLIP{31.11}{0.32} & \QWENVLM{27.81}{0.70} & \CONCH{18.00}{7.00} \\
        \CONCH{27.84}{0.31} & \PLIP{25.49}{0.68} & \RANDOM{17.00}{7.00} \\
        
        \QUILTCLIP{26.58}{0.31} & \PALIGEMMA{21.29}{0.64} & \PLIP{17.00}{7.00} \\
        \RANDOM{18.34}{0.27} & \RANDOM{19.13}{0.60} & \QUILTCLIP{13.00}{6.00} \\

        \bottomrule
    \end{tabular}
    \end{adjustbox}
    \small{\textsuperscript{ \ding{59}} \colorbox{newgreen}{ } General autoregressive VLMs \colorbox{newblue}{}  General contrastive VLMS   \colorbox{newpink}{} Pathology contrastive VLMS  \\ \colorbox{newpurple}{}  Biomedical contrastive VLMS.}
\label{tbl:uBenchPerformance_all}
\end{table}

\subsection{Model Performance on Pathology Specific Tasks}

While prior evaluations show that general contrastive VLMs have some biology and pathology knowledge (a finding also reported in \cite{Huang2023-plip} and \cite{ikezogwo2024quilt}), most specialist models analyzed in this work were fine-tuned. To this end, we analyzed the performance on pathology-only tasks and found similar rankings.

 \begin{table}[h]
    \centering
    \caption{ Macro-average accuracy (with bootstrap confidence interval) for coarse-grained and fine-grained perception  (pathology only tasks) and cognition  (reasoning) in \dataset\ . Robust models (byproduct of merging fine-tuned models with their respective base models) are also included.}
    \begin{adjustbox}{width=\textwidth}
    \begin{tabular}{c c c c c c}
        \toprule 
        \rowcolor{gray!10} \multicolumn{6}{c}{$\mu$-Bench} \\
        \midrule
        \midrule
        \multicolumn{2}{c}{Perception (Coarse-Grained)} & \multicolumn{2}{c}{ Perception (Fine-Grained)} & \multicolumn{2}{c}{Cognition (Reasoning) } \\
        \midrule
        Model & Accuracy ($\pm$ CI) & Model & Accuracy ($\pm$ CI) & Model & Accuracy ($\pm$ CI) \\
        \midrule 
        \midrule
        \GPT{71.29}{0.45} & \GPT{61.88}{1.13} & \GPT{62.00}{9.00} \\
        \QWENVLM{67.89}{0.47} & \CONCH{42.44}{1.13} & \QWENVLM{41.00}{10.00} \\
        \COGVLM{59}{0.51} & \RPLIP{39}{1.12} & \COGVLM{41.00}{10.00} \\
        \ALIGN{46.32}{0.50} & \RQUILTCLIP{36.95}{1.22} & \OPENCLIP{38.33}{8.33} \\
        \BIOMEDCLIP{45.5}{0.51} & \BIOMEDCLIP{35.29}{1.09}& \RPLIP{34.17}{8.33} \\
        \PALIGEMMA{43.34}{0.51} & \QUILTCLIP{33.28}{1.08} & \ALIGN{31.00}{9.00} \\
        \RPLIP{42.27}{0.51} & \OPENCLIP{32.35}{1.09} & \CLIP{28.00}{9.00} \\
        \RQUILTCLIP{37.68}{0.50} &\PLIP{32.02}{1.06}  & \RQUILTCLIP{25.83}{7.52} \\
        \OPENCLIP{37.59}{0.50} & \CLIP{28.63}{1.01} & \BIOMEDCLIP{25.00}{8.00} \\
         \CLIP{31.92}{0.47} & \ALIGN{27.51}{0.99}& \PALIGEMMA{25.00}{8.00} \\
        \CONCH{31.11}{0.32} & \QWENVLM{24.62}{0.97} & \CONCH{18.00}{7.00} \\
        \QUILTCLIP{29.18}{0.45} & \COGVLM{23.8}{0.98} & \RANDOM{17.00}{7.00} \\
        \PLIP{22.72}{0.42} & \PALIGEMMA{22.77}{0.97} & \PLIP{17.00}{7.00} \\
        \RANDOM{18.15}{0.39} & \RANDOM{17.74}{0.86} & \QUILTCLIP{13.00}{6.00} \\

        \bottomrule
    \end{tabular}
    \end{adjustbox}
    \small{\textsuperscript{ \ding{59}} \colorbox{newgreen}{ } General autoregressive VLMs \colorbox{newblue}{}  General contrastive VLMS   \colorbox{newpink}{} Pathology contrastive VLMS  \\ \colorbox{newpurple}{}  Biomedical contrastive VLMS.}
\label{tbl:uBenchPerformance_hande}
\end{table}

%%%%%%%%%%%%%%%%%%%%%
%%%%%% Figure %%%%%%%%
%%%%%%%%%%%%%%%%%%%%%

\begin{figure*}[ht!]
\centering
% \subfloat[]{
\includegraphics[width=0.54\linewidth, page=11, trim={0 1000 540 0},clip]{images/figs-robustness.pdf} \label{legend}
% } 

%%% General Models (Contrastive)
\subfloat[ Perception coarse-grained) \\
(Pathology only)]{\includegraphics[width=0.40\columnwidth]{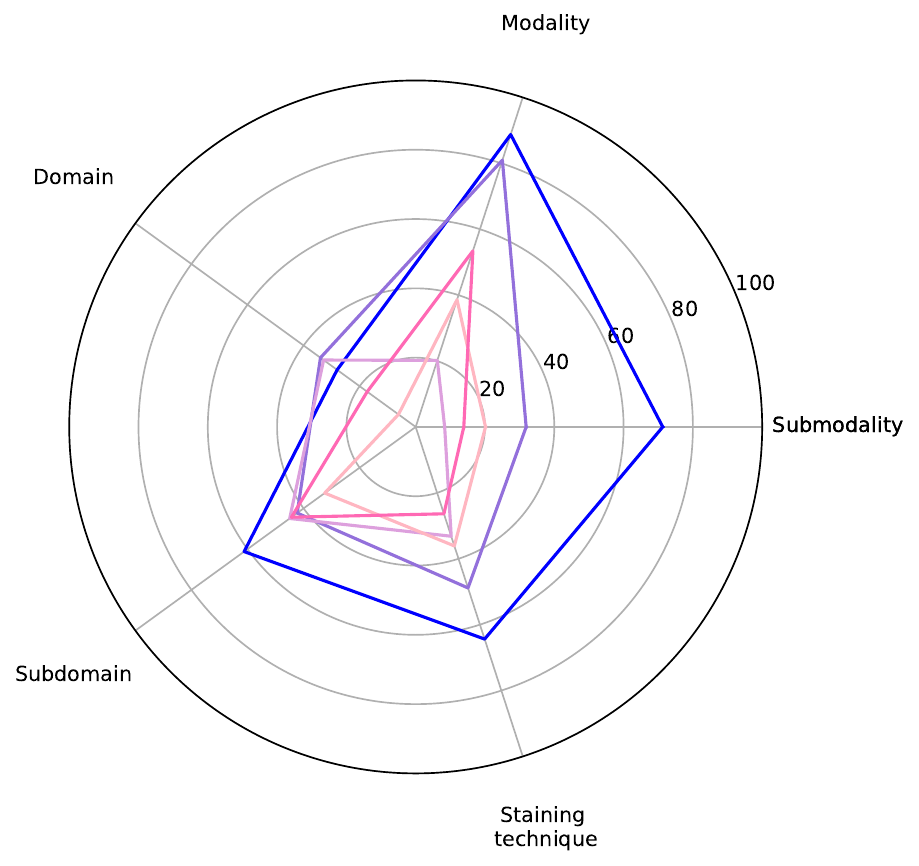}
   }  \label{fig5a1} \hspace{-3mm}
\subfloat[Perception fine-grained \\ (Pathology only) ]{\includegraphics[width=0.40\columnwidth]{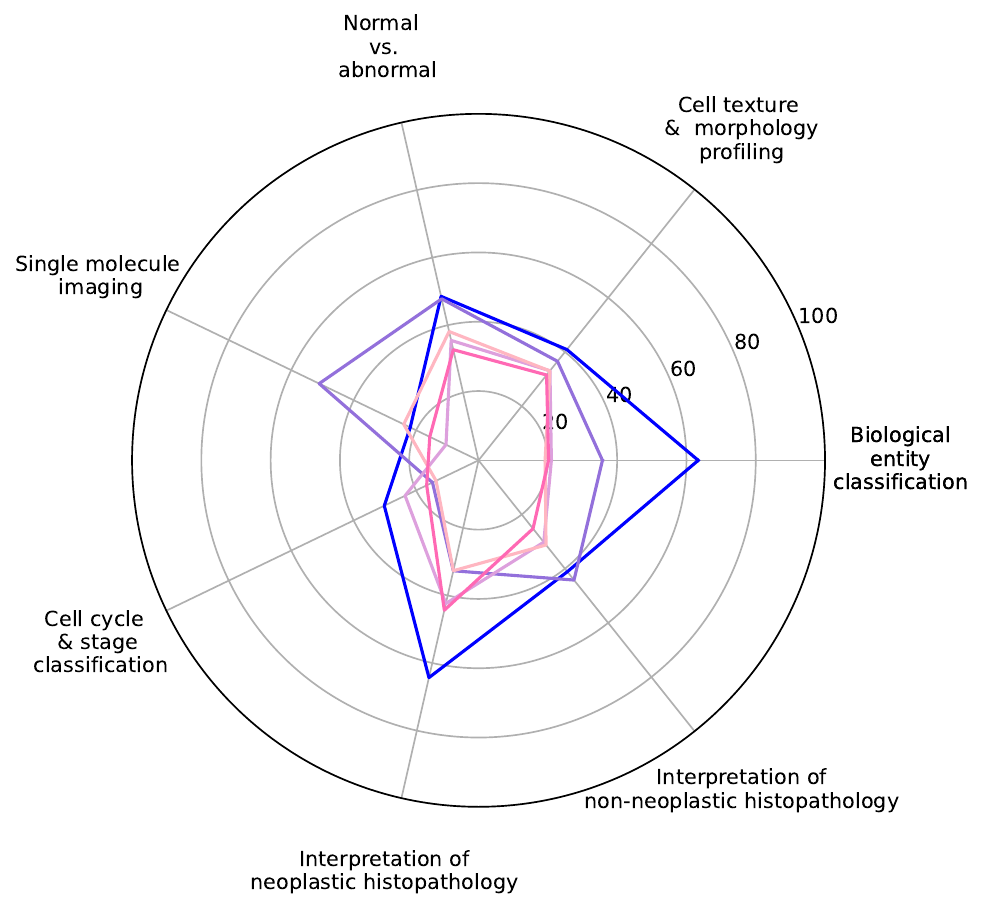}
    }\label{fig5b1}  \\
\subfloat[Understanding (Pathology)]{\includegraphics[width=0.40\columnwidth]
{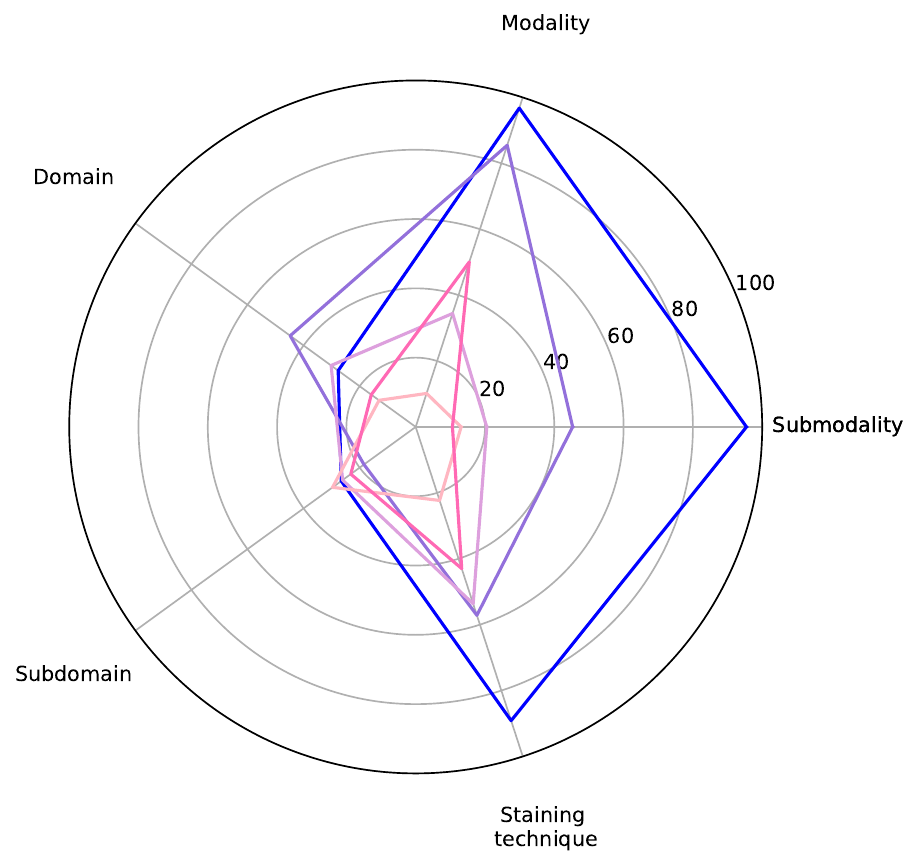}
    } \label{fig5c1} \hspace{-3mm}
\subfloat[Perception (fine-grained)]{\includegraphics[width=0.40\columnwidth]{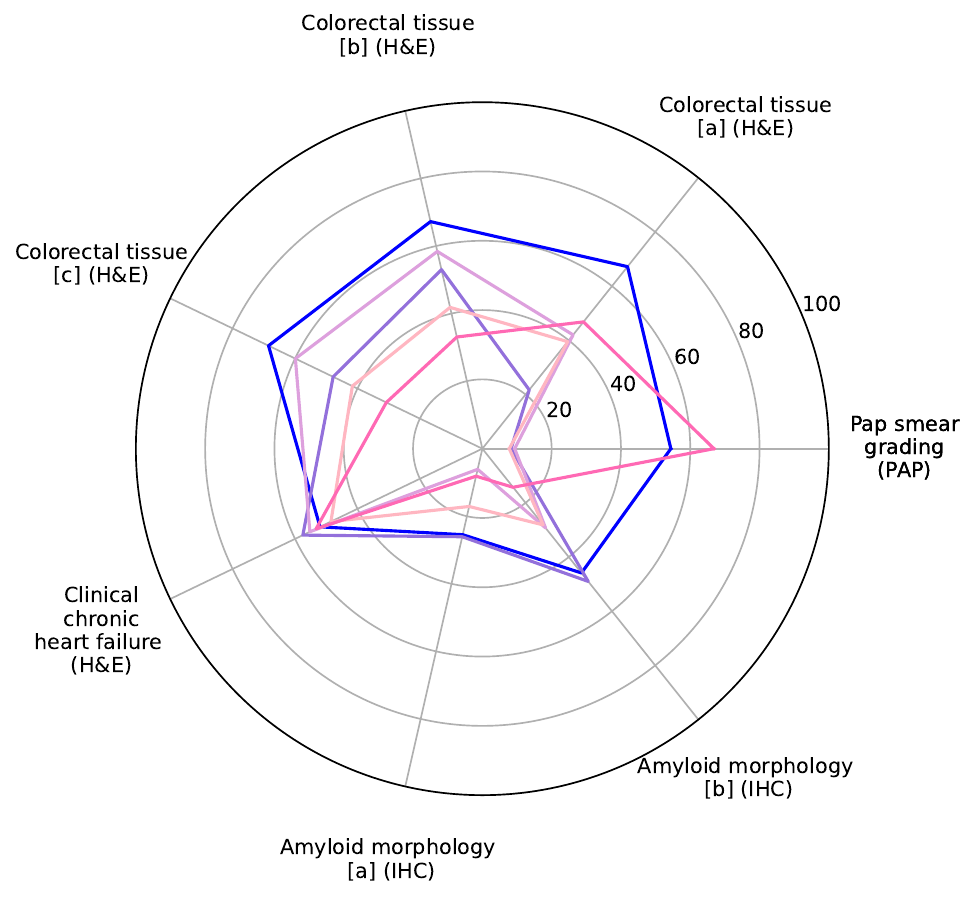}
    }\label{fig5c1_duplicate} \hspace{-3mm}
\caption{Performance comparison of top general domain auto-regressive models, contrastive models, and biomedical contrastive models across all axis of the \dataset\ [perception]}
\label{fig:main_results}
\label{confidence_histogram}
\end{figure*}

\section{Additional Benchmarking details}

\subsection{Model Details}
Table  \cref{tab:model_comparison} provides a breakdown of model parameters and training data (with dataset size), specialist models include their base model.

\subsection{Computing confidence intervals}
\label{sec:cofidence_intervals} 
Error bars represent 95\% confidence intervals (CI) computed via nonparametric bootstrapping using the SciPy $stats\text{.}bootstrap$ function with 1000 resamplings and default settings. No data were excluded from the analyses.

\subsection{Zero-shots results broken down by task}
\begin{figure}
    \centering
    \includegraphics[width=1\linewidth]{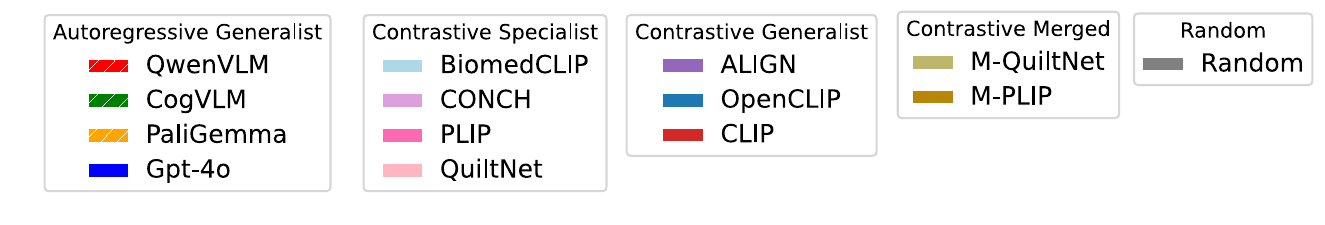} \label{legend}

    \includegraphics[width=0.8\linewidth]{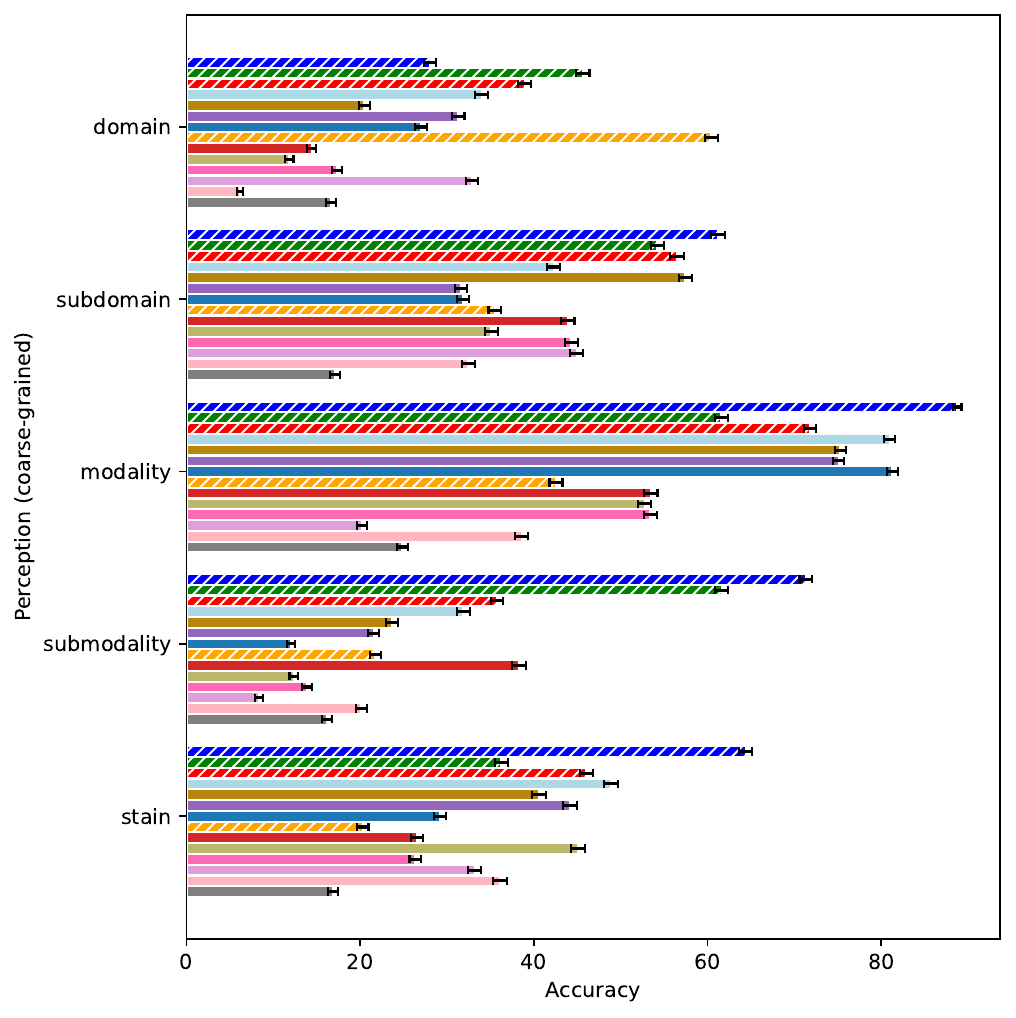}
    \caption{Perception (coarse-grained) results broken down by task}
    \label{fig:pc_}
\end{figure}

Figure \ref{fig:pc_} presents a breakdown of perception coarse-grained results by task. It reveals that autoregressive generalist models perform well across all tasks, as indicated by the overall averages. Notably, while GPT-4o dominates most tasks, PaliGemma excels in domain identification, achieving the best performance in that specific area.

\begin{figure}
    \centering
      \includegraphics[width=1\linewidth]{images/all_labels_1.pdf} \label{legend}

    \includegraphics[width=0.5\linewidth]{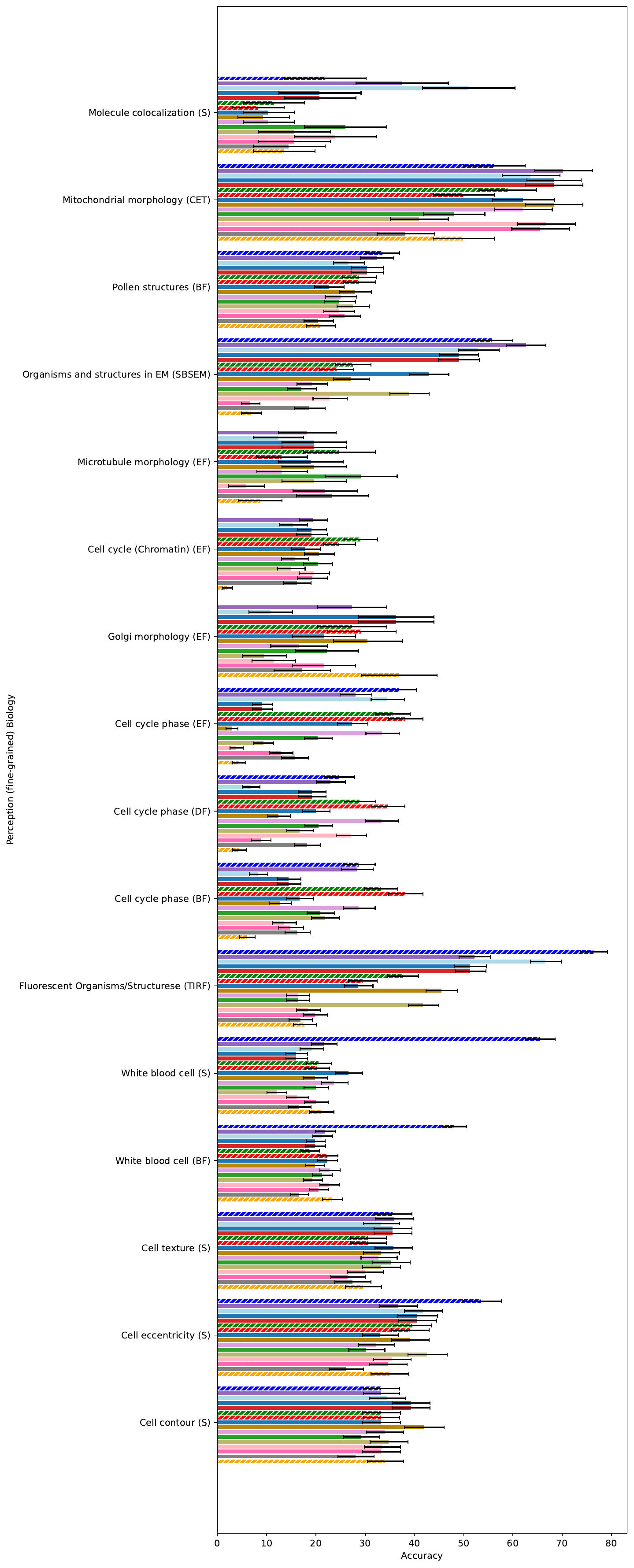}
    \caption{Perception (fine-grained) results broken down by task in biology}
    \label{fig:pf_b}
\end{figure}

Figure \ref{fig:pf_b} shows a breakdown of biology-specific perception fine-grained results by task. While GPT-4o dominates in some tasks, other models excel in specific tasks. For instance, ALIGN outperforms all models in molecular colocalization, while BiomedCLIP has the best performance in mitochondrial morphology classification.

\begin{figure}
    \centering
      \includegraphics[width=1\linewidth]{images/all_labels_1.pdf} \label{legend}

    \includegraphics[width=0.8\linewidth]{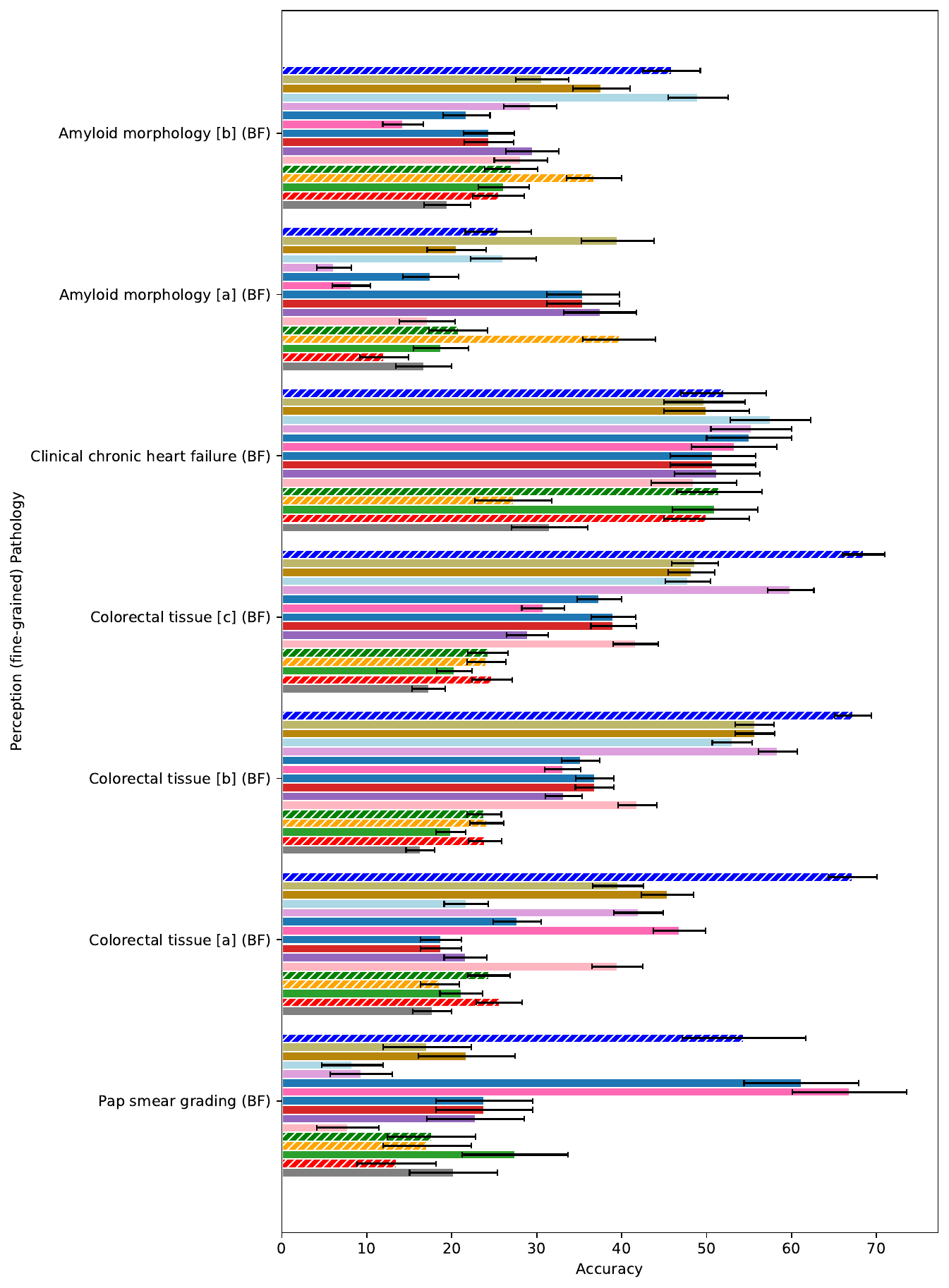}
    \caption{Perception (fine-grained) results broken down by task in pathology}
    \label{fig:pf_p}
\end{figure}

Figure \ref{fig:pf_p} shows a breakdown of pathology-specific perception fine-grained results by task. This breakdown reveals that while GPT-4o has the best performance in three tasks, specialist models still outperform it in amyloid morphology [a] and Pap smear grading.

%Some other table in \cref{fig:combined}

% \subsection{Analyzing model confidence}
% In these figures \cref{confidence_histogram} and \cref{confidence_histogram2} and \cref{confidence_histogram3} (there's actually two with the same label in a row).

\section{Computing resources}
One benefit of \dataset\ is that it is amenable to use by academic labs of all sizes, even those with limited resources. All evaluation tasks could be run on one NVIDIA a6000 (48GB VRAM), except for QWenVL, where we used one A100 (80GB VRAM). Inference required approximately 3.5 for all of \dataset\, running one dataset at a time (requiring one GPU at a time). The computations were performed on-premises university compute environment, with 1024 CPU cores that are AMD EPYC 9334, 2.70GHz.

\newpage
\FloatBarrier
% %%%%%%%%%%%%%%%%%%%%%
% %%%%%% TABLE %%%%%%%%
% %%%%%%%%%%%%%%%%%%%%%
% \begin{table}[h]
% \centering
% \caption{Instance identification breakdown of average correctness (↑) across question length (denoted by the number of characters). Bold indicates the best performance across all models within each quartile. Underlined values indicate the quartile in which a given model performs the best.}
% \input{tables/modality_submodality_domain_subdomain_stain}
% \label{provenance}
% \end{table}

%%%%%%%%%%%%%%%%%%%%%
%%%%%% TABLE %%%%%%%%
%%%%%%%%%%%%%%%%%%%%%
% \begin{table}[h]
% \centering
% \caption{Instance classification breakdown of average correctness (↑) across question length (denoted by the number of characters). Bold indicates the best performance across all models within each quartile. Underlined values indicate the quartile in which a given model performs the best.}
% \input{tables/classification_0.tex}
% \label{provenance}
% \end{table}

% % another table
% \input{tables/images_per_dataset}

%%%%%%%%%%%%%%%%%%%%%
%%%%%% TABLE %%%%%%%%
%%%%%%%%%%%%%%%%%%%%%
% \begin{table}[h]
% \centering
% \caption{Provenance of benchmark broken down by dataset, subdomain, submodality}
% \label{provenance}
% \end{table}

%%%%%%%%%%%%%%%%%%%%%
%%%%%% FIGURE %%%%%%%%
%%%%%%%%%%%%%%%%%%%%%
\begin{figure}
    \centering
    \includegraphics[width=1\linewidth]{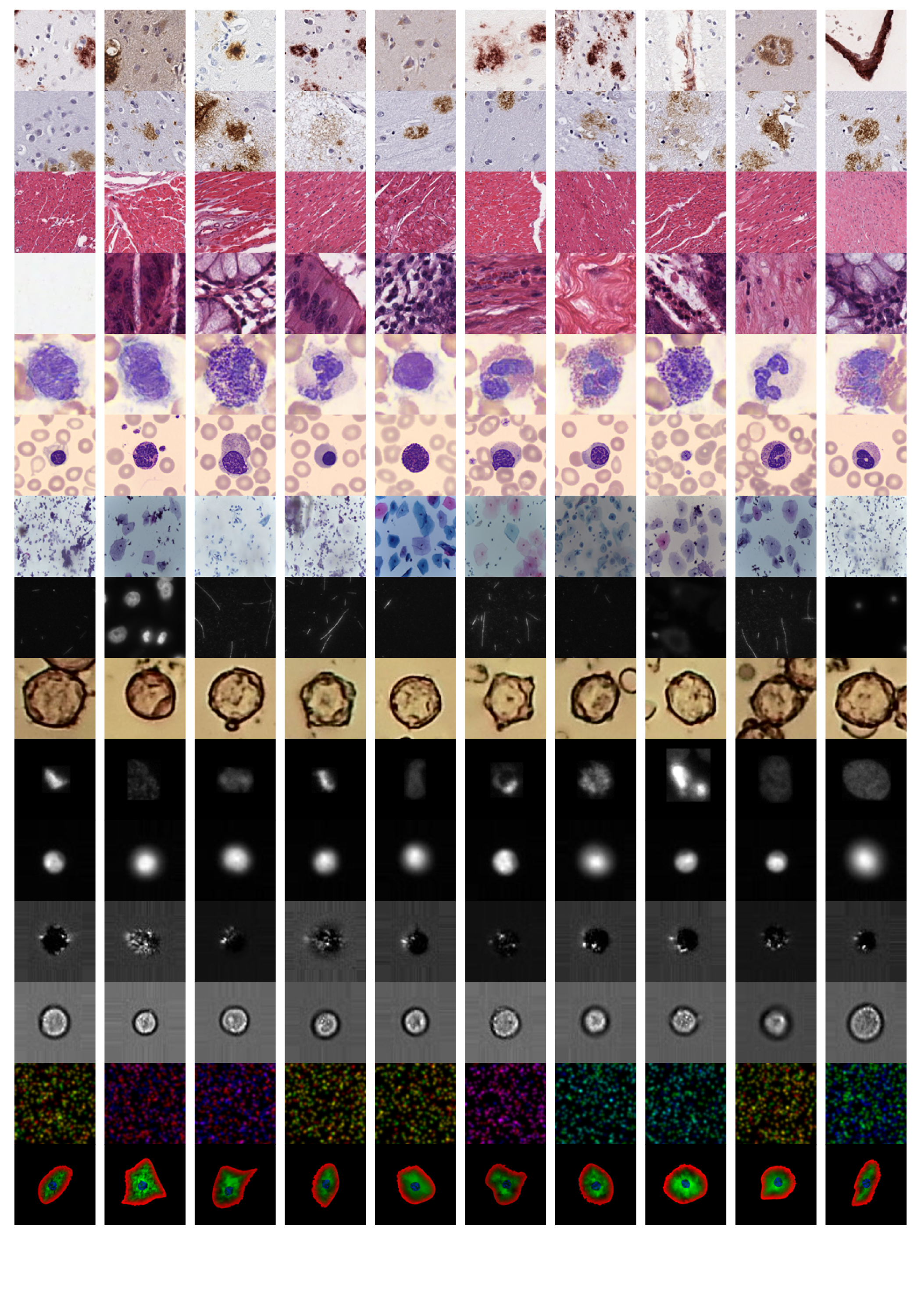}
    \caption{Ten random image samples from each dataset in  \dataset\ [perception].
     }    
     \label{repalebystaining}
\end{figure}

% lit review
\begin{table}[h]
\centering
\begin{adjustbox}{center}
\begin{tabular}{p{3cm}p{1cm}p{4cm}p{1cm}p{1cm}p{1.5cm}p{1.6cm}p{0.5cm}p{0.5cm}p{0.5cm}p{0.5cm}p{0.5cm}}
\toprule
Dataset & Domains & Source & Images  &  QA Pairs & Creation  & VQA Answer Type& VQA & SEG & CAP &CLS   & OD \\
\midrule
\textbf{Radiology} \\

VQA-Med-2018 \cite{hasan2018overview} & 1 & PubMed Central\textsuperscript{\textregistered} & 2,866 &  6,413 &  Automated & Open/Close & \checkmark \\

VQA-Med-2019  \cite{abacha2019vqa} & 1 & MedPix\textsuperscript{\textregistered}  & 4,200 &  15,292 & Automated  & Open/Close &  \checkmark \\

VQA-Med-2020  \cite{ben2021overview} & 1 & MedPix\textsuperscript{\textregistered}  & 5,000 &  5,000 & Automated  & Open/Close &  \checkmark \\

VQA-Med-2021  \cite{sitara2021ssn} & 1 & MedPix\textsuperscript{\textregistered}  & 5000 & 5000 & Automated  & Open/Close &  \checkmark \\

VQA-RAD \cite{lau2018dataset} & 1 & MedPix\textsuperscript{\textregistered} & 315 & 3,515 & Manual  & Open/Close & \checkmark \\

RadVisDial (S) \cite{kovaleva2020towards}  & 1 &  MIMIC-CXR & 91,060  & 455,300 & Automated & Close & \checkmark\\
RadVisDial (G) \cite{lau2018dataset} & 1 &  MIMIC-CXR & 100 & 500 & Manual & Close & \checkmark \\

OVQA \cite{huang2022ovqa} & 1 & EMRs &  2,001 & 19,020 &Automated&  Open/Close  &  \checkmark\\ 

SLAKE \cite{liu2021slake}& 1 & Decathlon,NIH Chest X-ray, CHAOS & 642 & 15,00 & Manual & Open/Close & \checkmark   & \checkmark\\\\

\textbf{Pathology} \\ 
PathVQA \cite{he2020pathvqa}  & 1 & PEIR Digital Library & 4,998 &32,799 & Automated  & Open &  \checkmark  \\
OpenPath \cite{huang2023visual}  & 1 & Twitter (Now X)&  208,414 &  208,414 & Automated  &  &&& \checkmark \\
\\ \\ 

\textbf{Biomedical} \\ 
PMC-VQA  \cite{zhang2023pmc}& 5+\textsuperscript{*} & PubMed Central\textsuperscript{\textregistered} & 149,000 & 227,000 & Automated  & Close  & \checkmark \\ \\

\textbf{Biology } \\
Multimodality Cell Segmentation Challenge \cite{ma2024multimodality} & 1    &  20 Laboratories& 1,500& -  && &&\checkmark\\

\midrule
\dataset\ (ours) & 3 \textsuperscript{*} &
Curated Datasets & 17,356 &  & Expert guided &   Open/Close &\checkmark   & \checkmark  & \checkmark  & \checkmark & \checkmark   \\
\bottomrule
\end{tabular}
\end{adjustbox}
\caption{Comparison of \dataset\ to existing \textbf{composite} medical and biomedical datasets. Only publicly available datasets were considered. 5+*: Mostly Radiology, Pathology, Microscopy, Signals and Generic biomedical illustrations.
}
\label{tab:lit_rev}
\end{table}

% table of model params
\begin{table}[h]
\centering
\caption{VLM Model Breakdown:}
\label{tab:model_comparison}
\begin{adjustbox}{center,max width=\textwidth}
\begin{tabular}{lp{1.9cm}llp{1.5cm}lp{1.5cm}ll}
\toprule
\textbf{Model} & \textbf{Total Params} & \textbf{Base VLM} & \textbf{Vision Encoder} & \textbf{VE Params} & \textbf{Text Encoder} & \textbf{TE Params} & \textbf{Training Data} &  \textbf{Size}\\
\midrule
\textbf{Contrastive VLMs} \\
CLIP & 151.2M & -  & ViT-B/32 & 86M &&& DataComp-1B &13B \\
ALIGN \cite{jia2021scaling} & 172.1M & - &  EfficientNet & 62.1M&  BERTbase & 109.4M& Internet&1.8B\\
CoCa \cite{yu2022coca}  & 383M & - &  ViT-B/16 & 86M&  & 297M& Internet&4.8B\\
OpenCLIP & 223.7M & -  & ViT-B/32 & 86M &&& DataComp-1B &13B \\ 
BLIP \cite{li2022blip} & 223.7M & -& ViT-B/16&86M& BERTbase&137.2M & Internet& 14M \\ 
\\
PLIP \cite{huang2023visual} & 151.2M &CLIP*&ViT-B/32&86M&&&OpenPath (X)&208.4K \\
QuiltNet \cite{ikezogwo2024quilt}  & 151.2M &CLIP*&ViT-B/16 & 86M& GPT2 (77CL)&&Quilt& 1M\\
BiomedCLIP & 195.M & OpenCLIP & ViT-B/16 & 86M & BioMedBERT \cite{chakraborty2020biomedbert} & 110M & PMC-15M& 15M \\
CONCH \cite{lu2023towards}  & 395.2M & CoCa & ViT-B/16 & 86M&& 1.17M\\
%OwlVIT2 & 154.9M \\ 
\\
\textbf{Auto-regressive VLM} \\
% dBLIP2 & 3.7B \\
CogVLM  \cite{wang2023cogvlm} & 17.6B &-& EVA2-CLIP-E && Vicuna-1.5-7B & 7B & Multiple & 1.5B \\
Qwen-VL \cite{bai2023qwen} & 9.6B &-& ViT-bigG & 1.9B & QwenLM & 7.7B & Multiple &1.4B \\
\bottomrule
\end{tabular}
\end{adjustbox}
\end{table}

\begin{table}[h]
    \centering
    \caption{\dataset\ Perception composition by imaging domain and subdomain}
    \begin{tabular}{lr}
    \toprule
    
    \textbf{Domains} & \textbf{Images}\\
    \midrule
    
    Biology & 8,637  \\
    Pathology & 8,678  \\
    \midrule
    
    \textbf{Subdomains}  & \textbf{Images}\\
    \midrule
    
    Cell Biology & 4979  \\
    Gastrointestinal and Liver Pathology & 4194 \\
    Hemato-pathology & 2600 \\
    Molecular Biology & 2229 \\
    Neuro-pathology & 1291 \\
    Botany & 726 \\
    Cardiovascular Pathology & 400 \\
    Neurobiology & 256 \\
    Gynecologic Cytology & 193 \\
    Parasitology & 192 \\
    Developmental Biology & 159 \\
    Biophysics & 96\\

    \bottomrule
    \end{tabular}
\label{tbl:uBenchDomainStatistics}
\end{table}

\begin{table}[h]
    \centering
    \caption{\dataset\ Perception composition by imaging modality and submodality}
    \begin{tabular}{lr}
    \toprule

    \textbf{Modalities} & \textbf{Images}\\
    \midrule
    
    Light Microscopy & 10864 \\
    Fluorescence Microscopy & 5618 \\
    Electron Microscopy & 833 \\
    \midrule

    \textbf{Submodalities} & \textbf{Images}  \\
    \midrule
    
    Brightfield Microscopy & 10121 \\
    Epifluorescence Microscopy & 2217 \\
    Synthetic & 1800 \\
    Confocal Microscopy & 1201 \\
    Darkfield Microscopy & 743 \\
    Serial Blockface Scanning Electron Microscopy & 577 \\
    Total Internal Reflection Fluorescence Microscopy & 400 \\
    Cryo-electron Tomography & 256 \\  

    \bottomrule
    \end{tabular}
\label{tbl:uBenchModalityStatistics}
\end{table}

\begin{table}[h]
    \centering
    \caption{\dataset\ Perception Fine-grained tasks}
    \begin{tabular}{lr}
    \toprule

    \textbf{Classification Tasks} & \textbf{Images} \\
    \midrule
    
    Cell cycle phase & 3169 \\
    Normal and abnormal tissues in colorectal adenocarcinoma or normal gastrointestinal tissue & 3114 \\
    White blood cell & 2600 \\
    Cell phenotypes in synthetic images & 1800 \\
    Amyloid beta pathology & 1291 \\
    Subcellular structures & 1105 \\
    Texture in colorectal cancer & 1000 \\
    Organisms and structures in fluorescence microscopy images & 934 \\
    Organisms and structures in electron microscopy images & 833 \\
    Normal and abnormal pollen grains & 700 \\
    Heart failure using cardiac histopathology images & 400 \\
    Pre-cancerous and cervical cancer lesions in liquid-based cytology Pap smear images & 193 \\
    Colocalization patterns & 96 \\
    \midrule

    \textbf{Segmentation and Object Detection Tasks} & \textbf{Images} \\
    \midrule

    Segmentation of white blood cells & 1600 \\
    Segmentation of subcellular structures & 1105 \\
    Cell segmentation in synthetic images & 600 \\
    Mitochondria segmentation in CryoET images & 256 \\
    Gland segmentation in benign and malignant colon histology images & 80 \\

    \bottomrule
    \end{tabular}
\label{tbl:uBenchFineGrainedTaskStatistics}
\end{table}

\begin{table}[h]
    \centering
    \caption{\dataset\ Perception composition by imaging stain}
    \begin{tabular}{lr}
    \toprule

    \textbf{Stains} & \textbf{Images} \\
    \midrule
    
    H\&E & 4594 \\
    Giemsa & 2600 \\
    Synthetic & 1916 \\
    No Stain & 1742 \\
    DAPI & 1105 \\
    H2B-mCherry & 783 \\
    Propidium Iodide & 743 \\
    Basic Fuchsin & 700 \\
    IHC(DAB) & 610 \\
    Uranyl Acetate & 577 \\
    IHC(HDab) & 491 \\
    AlexaFluor-tubulin & 400 \\
    Papanicolaou & 193 \\
    IHC(Red) & 190 \\
    Hoechst 33342 & 183 \\
    GalT–EGFP & 157 \\
    CellMask & 125 \\
    Tetraspeck Beads & 51 \\
    Lysotracker & 46 \\
    Fluorescent Beads & 26 \\
    Phal & 25 \\
    GalT-GFP & 24 \\
    Soluble GFP & 21 \\
    Uniform Test Slide & 11 \\
    LifeAct & 2 \\

    \bottomrule
    \end{tabular}
\label{tbl:uBenchStainStatistics}
\end{table}

\begin{table}[h]
    \centering
    \caption{\dataset\ Cognition composition by imaging domain and subdomain}
    \begin{tabular}{lr}
    \toprule
    
    \textbf{Domains} & \textbf{Question}\\
    \midrule
    
    Biology & 113  \\
    Pathology & 8  \\
    \midrule
    
    \textbf{Subdomains}  & \textbf{Question}\\
    \midrule

    Cell Biology & 45  \\
    Cell and molecular biology & 28 \\
    Neurobiology & 17 \\
    Developmental biology & 8 \\
    Immunology & 6 \\
    Neuropathology & 6 \\
    Gastrointestinal pathology & 2 \\ 
    Virology & 2 \\
    Botany & 2  \\
    Genetics &  2 \\
    \bottomrule
    \end{tabular}
\label{tbl:uBenchCognitionDomainStatistics}
\end{table}

\begin{table}[h]
    \centering
    \caption{\dataset\ Cognition composition by imaging modality and submodality}
    \begin{tabular}{lr}
    \toprule

    \textbf{Modalities} & \textbf{Questions}\\
    \midrule
    Fluorescence Microscopy & 76 \\
    Light Microscopy & 26 \\
    Electron Microscopy & 17 \\
    \midrule

    \textbf{Submodalities} & \textbf{Questions}  \\
    \midrule

    Epifluorescence microscopy & 37 \\
    Confocal microscopy  & 36 \\
    Brightfield microscopy &  21 \\
    Cryo-electron tomography &  8 \\
    Scanning electron microscopy &  4 \\
    Transmission electron microscopy &  4 \\
    Differential interference contrast microscopy &  4 \\
    Mixed &  3 \\
    Transmission electron microscopy (TEM) &  1 \\
    Lattice light sheet &  1 \\
    Synthetic & 1 \\
    Lattice light-sheet microscopy & 1 \\
    \bottomrule
    \end{tabular}
\label{tbl:uBenchCogntionModalityStatistics}
\end{table}

\begin{table}[]
 \caption{\dataset\ Perception Coarse-Grained  Question and Caption templates}
\begin{adjustbox}{center,max width=\textwidth}
\begin{tabular}{@{}lp{8cm}p{8cm}@{}}
\toprule
\textbf{Type} & \textbf{Question}                                                            & \textbf{Caption}                    \\ \midrule
\textbf{Modality}      & What is the most likely microscopy modality used to acquire this image?      & A microscopy image obtained through \{modality\}.            \\
\textbf{Submodality}    & What is the most likely microscopy submodality used to acquire this image?   & A microscopy image obtained through \{submodality\}.           \\
\textbf{Domain}        & What is the most likely field of study this micrograph would be used for?    & A microscopy image frequently studied in \{domain\}.   \\
\textbf{Subdomain}     & What is the most likely subfield of study this micrograph would be used for? & A microscopy image frequently studied in \{subdomain\}.  \\
\textbf{Stain}         & What is the most likely technique used to stain this micrograph?             & A microscopy image stained with \{stain\}.                     \\ \bottomrule
\end{tabular}
\end{adjustbox}
\label{tbl:template_coarse_grained}
\end{table}

\begin{table}[]
\caption{\dataset\ Perception Fine-grained: Question and caption templates for biology tasks}
\begin{adjustbox}{center,max width=\textwidth}

\begin{tabular}{p{3cm}p{8cm}p{8cm}}
\hline
\textbf{Dataset}                                                              & \textbf{Question}                                                                                                                                                                                                                & \textbf{Caption}                                                                                                                                                                            \\ \hline

PCST-Contour                           & A synthetic fluorescence micrograph is displayed. What is the most likely description for the cells contour irregularities?                                                                                                     & A synthetic fluorescence microscopy image of a cell with \{class\} contours.                                                                                                                \\
PCST-Eccentricity           & A synthetic fluorescence micrograph displayed. What is the most likely description for the cells eccentricity phenotype?                                                                                                        & A synthetic fluorescence microscopy image of a cell with \{class\} eccentricity.                                                                                                            \\
PCST-Texture                           & A synthetic fluorescence micrograph displayed. What is the most likely description for the cells cytoplasm texture?                                                                                                             & A synthetic fluorescence microscopy image of the cell cytoplasm with \{class\} texture.                                                                                                     \\
Colocalization benchmark               & A synthetic confocal fluorescence micrograph of small points in two different channels with different levels of colocalization. Given the image provided, what is the most accurate description for the colocalization patterns? & A synthetic confocal fluorescence microscopy image of small points in two different channels with different levels of colocalization. The image displays \{class\} colocalization patterns. \\
EMPIAR SBF-SEM                        & An electron micrograph is shown. Based on the image, what is the most likely structure on the field of view?                                                                                                                     & A Serial blockface scanning electron microscopy image shows \{class\}                                                                                                                       \\
BBBC048 (Brightfield)     & A brightfield micrograph of jurkat cells acquired using flow cytometry (single cell). Based on the micrograph, what is the most likely cell phase?                                                                               & Brightfield microscopy imaging flow cytometry is used to visualize single Jurkat cells at different cell cycle phases. The image displays a cell in \{class\} stage of the cell cycle.      \\
BBBC048 (Darkfield)       & A darkfield micrograph of jurkat cells acquired using flow cytometry (single cell). Based on the micrograph, what is the most likely cell phase?                                                                                 & A darkfield microscopy imaging flow cytometry is used to visualize single jurkat cells at different cell cycle phases. The image displays a cell in \{class\}.                              \\
BBBC048 (Epifluorescence)  & A propidium iodide stained fluorescence micrograph of Jurkat cells acquired using flow cytometry (single cell). Based on the micrograph, what is the most likely cell phase?                                                     & Epifluorescence microscopy imaging (flow cytometry) shows single Jurkat cells stained with propidium iodide at different cell cycle phases. The image displays a cell in \{class\}.         \\
CellCognition (Golgi)    & A fluorescence micrograph of Hela Kyoto cells stably expressing GalT-eGFP to label the Golgi apparatus. Based on the image what is the most likely the Golgi apparatus morphology?                                               & Fluorescence microscopy image of human Hela Kyoto cells stably expressing galactosyltransferase (GalT-eGFP) to label the Golgi apparatus showing \{class\} morphology.                      \\
CellCognition (H2B)                        & A fluorescence microscopy image of human Hela Kyoto cells with stable chromatin marker expression. Based on the image, what is the most likely cell cycle stage?                                                                 & Fluorescence microscopy image of human Hela Kyoto cells with stable chromatin marker expression. The micrograph displays a cell in \{class\} stage of the cell cycle.                       \\
CellCognition (Mt)                       & A fluorescence micrograph of Hela Kyoto cells stably expressing eGFP-labeled tubulin to label microtubules is shown. Based on the micrograph what is the most likely microtubule morphology?                                     & Fluorescence microscopy image of human Hela Kyoto cells with stable chromatin marker expression (eGFP) displays microtubules showing \{class\} morphology.                                  \\
ICPR2020 Pollen              & Basic fuchsin stained light micrograph of pollen grains. Based on the image, what is the most likely pollen class?                                                                                                               & A brightfield microscopy image of pollen grains shows \{class\} structures.                                                                                                                 \\

Wu et al 2023                              & A cryo-electron tomography of mitochondria in neurons cultured in vitro is shown. Based on the image what is the most likely mitochondrial morphology?                                                                           & A cryo-electron tomography of mitochondria in neurons cultured in vitro shows \{class\}.                                                                                                    \\
Fluorescence Cells \& Structures                                              & A fluorescence micrograph is shown. Based on the image, what is the most likely structure?                                                                                                                                       & A photomicrograph shows a fluorescence microscopy \{class\}.                                                                                                                                \\ \bottomrule    
\end{tabular}
\end{adjustbox}
\label{tbl:template_coarse_grained}
\end{table}

\begin{table}[]
\caption{\dataset\ Perception Fine-grained: Question and caption templates for pathology tasks}
\begin{adjustbox}{center,max width=\textwidth}
\begin{tabular}{p{3cm}p{8cm}p{8cm}}
\hline
\textbf{Dataset}                                                              & \textbf{Question}                                                                                                                                                                                                                & \textbf{Caption}                                                                                                                                                                            \\ \hline
Acevedo et al 2020                   & A Giemsa-stained light micrograph displaying human peripheral blood cells. As a blood cell recognition system, identify the correct cell type:                                                                                   & A brightfield microscopy image of a peripheral blood smear stained with giemsa displaying \{article\} \{class\}.                                                                            \\

Jung et al 2022                        & Synthetically generated Giemsa-stained light micrograph of human peripheral blood cell. As a blood cell recognition system, identify the correct cell type:                                                                      & A synthetic microscopy image of a peripheral blood smear stained with giemsa, displays \{article\} \{class\}.                                                                               \\
Kather et al 2016                      & H\&E stained light micrograph of human colorectal tissue. Based on the image, what is the most likely texture class?                                                                                                             & H\&E stained light microscopy image of human colorectal tissue with \{class\}.                                                                                                          \\
Kather et al 2018                        & H\&E stained light micrograph of human colorectal tissue. Based on the image, what is the most likely texture class?                                                                                                         & H\&E stained light microscopy image of human colorectal tissue with \{class\}.                                                                                                          \\
Kather et al 2018 Val7K            & H\&E stained light micrograph of human colorectal tissue. Based on the image, what is the most likely texture class?                                                                                                         & H\&E stained light microscopy image of human colorectal tissue with \{class\}.                                                                                                          \\
Nirschl et al 2018           & H\&E stained light micrograph of human cardiac tissue. Based on the image, what is the most likely clinical chronic heart diagnosis?                                                                                         & H\&E stained light microscopy image of human cardiac tissue with \{class\} texture.                                                                                                                                                                         \\
Tang et al 2019                           & IHC stained light micrograph of extracellular amyloid-beta deposition in the human brain tissue. Based on the image, what is the most likely amyloid beta morphology pattern?                                                    & Human brain tissue is stained with immunohistochemistry for amyloid-beta and imaged using brightfield microscopy. The micrograph displays \{class\} morphology.                               \\
Wong et al 2022                       & IHC stained light micrograph of extracellular amyloid-beta deposition in the human brain tissue. Based on the image, what is the most likely amyloid beta morphology pattern?                                                    & Human brain tissue is stained with immunohistochemistry for amyloid-beta and imaged using brightfield microscopy. The micrograph displays \{class\} morphology.                               \\ \bottomrule  

\end{tabular}
\end{adjustbox}
\label{tbl:template_coarse_grained_pathology}
\end{table}

\begin{figure*}[htb]
\begin{mdframed}[backgroundcolor=lightblue2]
\begin{verbatim}

light microscopy:
    - brightfield microscopy
    - phase contrast microscopy
    - differential interference contrast microscopy
    - darkfield microscopy
    - polarized light microscopy
    - mixed
    - synthetic
fluorescence microscopy:
    - confocal microscopy
    - epifluorescence microscopy
    - single-molecule localization microscopy (SMLM)
    - stimulated emission depletion microscopy (STED)
    - total internal reflection fluorescence microscopy (TIRF)
    - fluorescence recovery after photobleaching (FRAP)
    - fluorescence resonance energy transfer (FRET)
    - fluorescence in situ hybridization (FISH)
    - fluorescence correlation spectroscopy (FCS)
    - mixed
    - synthetic
electron microscopy:
    - atomic force microscopy
    - scanning electron microscopy
    - serial blockface scanning electron microscopy
    - cryo-electron microscopy
    - cryo-electron tomography
    - immuno-electron microscopy
    - mixed
    - synthetic
\end{verbatim}
\end{mdframed}
\caption{Modality with submodality YAML  file.}
\label{fig:metadata_microscopy}
\end{figure*}

\begin{figure*}[htb]
\begin{mdframed}[backgroundcolor=lightblue2]
\tiny{
\begin{verbatim}

biology:
    - anatomy
    - biochemistry
    - biophysics
    - biotechnology
    - botany
    - cell and molecular biology
    - cell biology
    - cell cycle
    - conservation biology
    - developmental biology
    - ecology
    - evolutionary biology
    - genetics
    - immunology
    - marine biology
    - microbiology
    - neurobiology
    - parasitology
    - pharmacology
    - physiology
    - structural biology
    - systems biology
    - virology
    - zoology
dermatology:
    - infectious dermatology
    - medical dermatology
    - neoplastic dermatology
ophthalmology:
    - cornea and external eye ophthalmology
    - retinal surgery
    - diabetic retinopathy
    - neuro-ophthalmology
    - pediatric ophthalmology
    - neoplastic ophthalmology
    - medical ophthalmology
cytology:
    - gynecologic cytology
    - non-gynecologic cytology
    - fine needle aspiration cytology
pathology:
    - autopsy pathology
    - blood banking and transfusion medicine
    - bone and soft tissue pathology
    - breast pathology
    - cardiovascular pathology
    - clinical pathology
    - dermatopathology
    - endocrine pathology
    - forensic pathology
    - gastrointestinal pathology
    - genitourinary pathology
    - gynecologic pathology
    - head and neck pathology
    - hematopathology
    - hepatobiliary pathology
    - infectious disease pathology
    - molecular pathology
    - nephropathology
    - neuropathology
    - oral pathology
    - ophthalmic pathology
    - pancreatic pathology
    - pediatric pathology
    - pulmonary and pleural pathology
    - renal and medical kidney pathology
    - surgical pathology
radiology:
    - abdominal radiology
    - breast imaging
    - cardiothoracic radiology
    - emergency radiology
    - gastrointestinal radiology
    - genitourinary radiology
    - head and neck radiology
    - interventional radiology
    - musculoskeletal radiology
    - neuroradiology
    - nuclear radiology
    - pediatric radiology
    - vascular and interventional radiology
    
\end{verbatim}
}
\end{mdframed}
\caption{Domain with subdomain YAML  file.}
\label{fig:metadata_domain}
\end{figure*}

\begin{figure*}[htb]
\begin{mdframed}[backgroundcolor=lightblue2]
\begin{verbatim}

light microscopy:
    - H&E
    - IHC(HDab)
    - IHC(Red)
    - Giemsa
    - PAS
    - Papanicolaou
    - Masson Trichrome
    - Toluidine Blue
    - Wright-Giemsa
    - Ziehl-Neelsen
    - Gram
    - Congo Red
    - Alcian Blue
    - Basic fuchsin
    - None
    - synthetic
fluorescence microscopy:
    - DAPI   
    - Hoechst
    - propidium iodide
    - SYTOX
    - Alexa Fluor 350 # blue
    - Alexa Fluor 405
    - GFP     
    - FITC
    - Cy2
    - Alexa Fluor 488
    - RFP    
    - Cy3
    - H2B-mCherry
    - Texas Red
    - Alexa Fluor 555
    - Alexa Fluor 568
    - Cy5    
    - Alexa Fluor 647
    - Alexa Fluor 660
    - AlexaFluor-tubulin
    - synthetic
    - GalT–EGFP d
electron microscopy:
    - uranyl acetate
    - osmium tetroxide
    - lead citrate
    - phosphotungstic acid
    - tannic acid
    - sodium silicotungstate
    - sodium phosphotungstate
    - sodium metaperiodate
    - synthetic
    - None

\end{verbatim}
\end{mdframed}
\caption{Modality with submodality YAML  file.}
\label{fig:metadata_stain}
\end{figure*}

\begin{figure*}[htb]
\begin{mdframed}[backgroundcolor=lightblue]
\begin{verbatim}

I'm creating a dataset to evaluate VLM understanding on biomedical images. 
Could you convert this user input question into a multi-choice question 
with 6 answer choices? One choice should be "None of the above", 
and this choice should have a 1/6 chance of being correct. 
Output a JSON format:
{"question": str, "choices": list, "answer": int (start from 0)}.


Context: {question["CONTEXT"]}
Input Question: {question["INPUT"]}
Correct Answer: {answer}

\end{verbatim}
\end{mdframed}
\caption{Prompt used to convert open VQA to closed  VQA}
\label{fig:prompy_OVQA_CVQA}
\end{figure*}

\begin{figure*}[htb]
\begin{mdframed}[backgroundcolor=lightblue]
\begin{verbatim}

Given this question, can you help me annotate the following fields?

## Modality and Submodality
{Modalities YAML}

## Domain and Subdomain
{Domains YAML}

## Scale (nano/subcellular, micro/cellular, macro/tissues)
{Scales Table}

## Content
+ Gene pathways
+ Metabolic pathways
+ Cell signalling and signal transduction
+ Cell physiology/function
+ Protein-protein interactions
+ Cell-cell interaction
+ Unique properties of the cell of origin/cell type in the image
+ Cytoskeleton and cell structure/morphology
+ Drug or small molecule mechanism of action
+ Other

## Relevant biological keywords 
For example, brain, HeLa, mitochondria, GFP, etc

Output a JSON with {"modality": str, 
                    "submodality": str, 
                    "domain": str, 
                    "subdomain": str, 
                    "scale": str, 
                    "content": str, 
                    "keywords": list[str]}
                    
\end{verbatim}
\end{mdframed}
\caption{Prompt use to classify questions post-hoc}
\label{fig:prompt_classify}
\end{figure*}

\begin{figure*}[htb]
\begin{mdframed}[backgroundcolor=lightblue]
\tiny{
\begin{verbatim}

metadata:
  height: 250
  width: 250
  name: 01145_6de79663_33375_split-test_chronic-heart-failure.png
  format: .png
  createdAt: '2024-05-27T17:39:08.866Z'
  updatedAt: '2024-05-27T17:39:08.866Z'
comments: []
custom_metadata:
  age: 47.0
  classes_to_idx:
    not_chronic_heart_failure: 0
    chronic_heart_failure: 1
  cvdo_id:
  - CVDO_0000569
  cvdo_name:
  - cardiomyopathy
  dataset_name: nirschl_et_al_2018
  dataset_slug: nirschl_et_al_2018
  domain: pathology
  ethnicity: Caucasian
  filename: 01145_6de79663_33375_split-test_chronic-heart-failure.png
  file_size: 141080
  institution:
  - upenn
  image_id: 6de79663-0392-4f20-b2ea-16ddf4e3b4e4
  image_md5: 97ee369881fcbfd18fb4fb8dcfe2ca17
  label: 1
  label_name: chronic_heart_failure
  label_subname: cardiomyopathy
  label_task: classification of heart failure using cardiac histopathology images
  last_updated: '2024-05-27T17:39:08.868Z'
  license: CC-BY-4.0
  microns_per_pixel: 2.0
  modality: light microscopy
  ncbitaxon_id:
  - NCBITaxon_9606
  ncbitaxon_name:
  - Homo sapiens
  ncit_id:
  - NCIT_C50577
  normal_or_abnormal: abnormal
  original_filename: 33375_0_fal_20_0.png
  patient_id: '33375'
  pato_id:
  - PATO_0000384
  sex: male
  snomedct_id:
  - SCTID_48447003
  split: test
  stain: H&E
  subdomain: cardiovascular pathology
  submodality: brightfield microscopy
  supported_tasks:
  - multi_class
  uberon_id:
  - UBERON_0000948
  uberon_name:
  - heart
  questions: null
tags:
- CVDO_0000569
- H&E
- Homo sapiens
- NCBITaxon_9606
- NCIT_C50577
- PATO_0000384
- SCTID_48447003
- UBERON_0000948
- brightfield microscopy
- cardiomyopathy
- cardiovascular pathology
- heart
- light microscopy
- pathology

\end{verbatim}}
\end{mdframed}
\caption{\dataset\ Perception: Example of densely annotated metadata for a single data point. Metadata is collected and reviewed by an expert. }
\label{fig:data_point}
\end{figure*}

\begin{figure*}[htb]
\begin{mdframed}[backgroundcolor=lightblue3]

\begin{verbatim}

Answer with a single letter, no extra details.
Question: {question}
{options}

\end{verbatim}
\end{mdframed}
\caption{Prompts used to run inference with auto-regressive models }
\label{fig:autoregressive_promtps}
\end{figure*}

\begin{figure*}[htb]
\begin{mdframed}[backgroundcolor=lightblue3]

\begin{verbatim}
You are invited to an alpha-testing phase of a vision-language chat app 
for biologists. The application is free of charge, poses no risks that 
would not be present with general internet usage, and you may stop use 
at any time. You may use it for your daily research or however you wish. 

Website:
    Create an account at: ###
    Acknowledge terms of service
    Registration and use of the app is consent to use the submitted  
    image/text for model training and testing purposes. The user's field 
    of study and training level will be recorded. However, all personal
    information will remain confidential. The users will retain the 
    copyright and ownership of the input image data. However, the terms 
    of service allow permission to use and redistribute the image under 
    a CC-BY-SA 4.0 license. The raw and/or curated image-text data may be 
    used to create a public benchmark of real-world biology user-AI 
    assistant instruction tuning dataset to benefit the biomedical 
    computer vision community.
    
Main interface
1. Upload a biology or biomedical image.
    Describe the image as context for the model. For example: 
    "Actin (orange) and mitochondria (cyan) in a micropatterned HeLa cell. 
    This is a still image from time-lapse live cell imaging. Wild-type 
    genotype and no drug treatment."
    
2. Prompt/question:
    For example: How would you describe the pattern of the 
    organelle in cyan? 
    What is the most likely organelle? What antibody marker or dye 
    specifically labels this organelle for cell biology experiments?
    Feel free to challenge the model with difficult questions requiring 
    complex reasoning. There is no need to limit questions to simple 
    perceptual tasks such as classification ("what is in the image?") 
    or visual question-answering. Have it reason and interpret 
    images in a way that would challenge a new biology graduate student.
    Ask the model to do complex image-based reasoning about biological 
    processes/pathways:
    
    "Given the gene knockout cell image provided, what biological pathway 
    is most likely disrupted, if any?"
    
    "Are there any small molecules that reverse this phenotype?"
    
    "What diseases are associated with gene <my favorite gene>?"
    
    Determine whether the VLM can understand true biological signals vs. 
    artifacts that confound analysis, See how the VLM handles diverse 
    modalities and experiments (EM, fluorescence, brightfield, CLEM etc) 
    as well as different cell types and tissues. 
    Ask the model to generate new hypotheses based on an image or connect 
    to relevant literature. 
    
    After you have a response to your question, we encourage you to 
    provide feedback on the VLM answer. Please give details on why the 
    answer is correct/helpful or incorrect/not helpful. As needed, 
    provide additional details as to whether the VLM
      1) understood the  question
      2) identified the biological feature(s) in the image
      3) provided a correct biological interpretation/answer. 


\end{verbatim}
\end{mdframed}
\caption{Email Invitation: Invitation for collaboration send to submitters. }
\label{fig:invitation}
\end{figure*}

\newpage

%\subsection{\dataset\ Perception (Coarse-grained) examples}
% cognition data samples

% perception data samples

\FloatBarrier
\section{\dataset\ Perception (Coarse-Grained) Closed VQA Data Samples}
\label{perception_coarse_grained_data_examples}

\begin{table}[ht]
\centering
\begin{tabular}{|p{0.6\linewidth}|p{0.3\linewidth}|}
\hline
\multicolumn{2}{|c|}{\textbf{QUESTION TYPE:  Modality}} \\ 
\hline
\raggedright \textbf{Question:} What is the most likely microscopy modality used to acquire this image? \\
\textbf{Options:} \\
A. electron microscopy\newline  B. fluorescence microscopy\newline \textcolor{orange}{C. light microscopy}\newline D. none of the above\newline  &
\vspace{3em}\includegraphics[width=\linewidth]{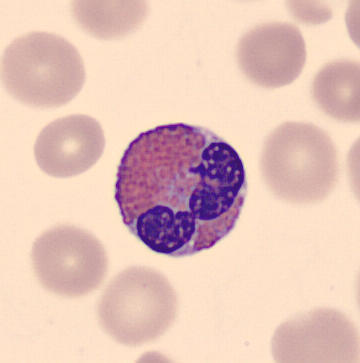} \\
\hline
\end{tabular}
\end{table}

\begin{table}[ht]
\centering
\begin{tabular}{|p{0.6\linewidth}|p{0.3\linewidth}|}
\hline
\multicolumn{2}{|c|}{\textbf{QUESTION TYPE: Modality}} \\ 
\hline
\raggedright \textbf{Question:} What is the most likely microscopy modality used to acquire this image? \\
\textbf{Options:} \\
A. light microscopy\newline  
B. electron microscopy\newline 
\textcolor{orange}{C. fluorescence microscopy}\newline 
D. none of the above\newline  &
\vspace{3em}\includegraphics[width=\linewidth]{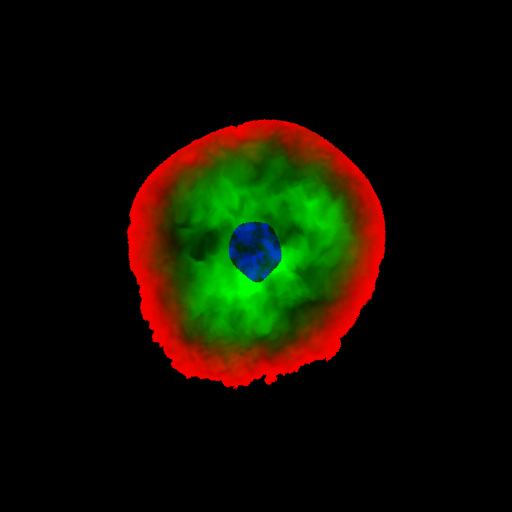} \\
\hline
\end{tabular}
\end{table}

\begin{table}[ht]
\centering
\begin{tabular}{|p{0.6\linewidth}|p{0.3\linewidth}|}
\hline
\multicolumn{2}{|c|}{\textbf{QUESTION TYPE: Submodality}} \\ 
\hline
\raggedright \textbf{Question:} What is the most likely microscopy submodality used to acquire this image? \\
\textbf{Options:} \\
A. fluorescence resonance energy transfer (FRET)\newline  
B. total internal reflection fluorescence microscopy (TIRF)\newline  
C. mixed\newline  
D. stimulated emission depletion microscopy (STED)\newline 
\textcolor{orange}{E. confocal microscopy}\newline 
F. none of the above\newline  &
\vspace{3em}\includegraphics[width=\linewidth]{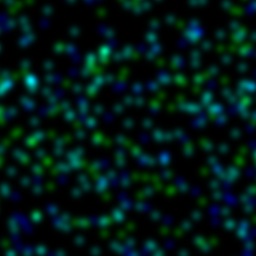} \\
\hline
\end{tabular}
\end{table}

\begin{table}[ht]
\centering
\begin{tabular}{|p{0.6\linewidth}|p{0.3\linewidth}|}
\hline
\multicolumn{2}{|c|}{\textbf{QUESTION TYPE: Submodality}} \\ 
\hline
\raggedright \textbf{Question:} What is the most likely microscopy submodality used to acquire this image? \\
\textbf{Options:} \\
\textcolor{orange}{A. serial blockface scanning electron microscopy}\newline  
B. atomic force microscopy\newline  
C. synthetic\newline  
D. cryo-electron microscopy\newline  
E. mixed\newline  
F. none of the above\newline  &
\vspace{3em}\includegraphics[width=\linewidth]{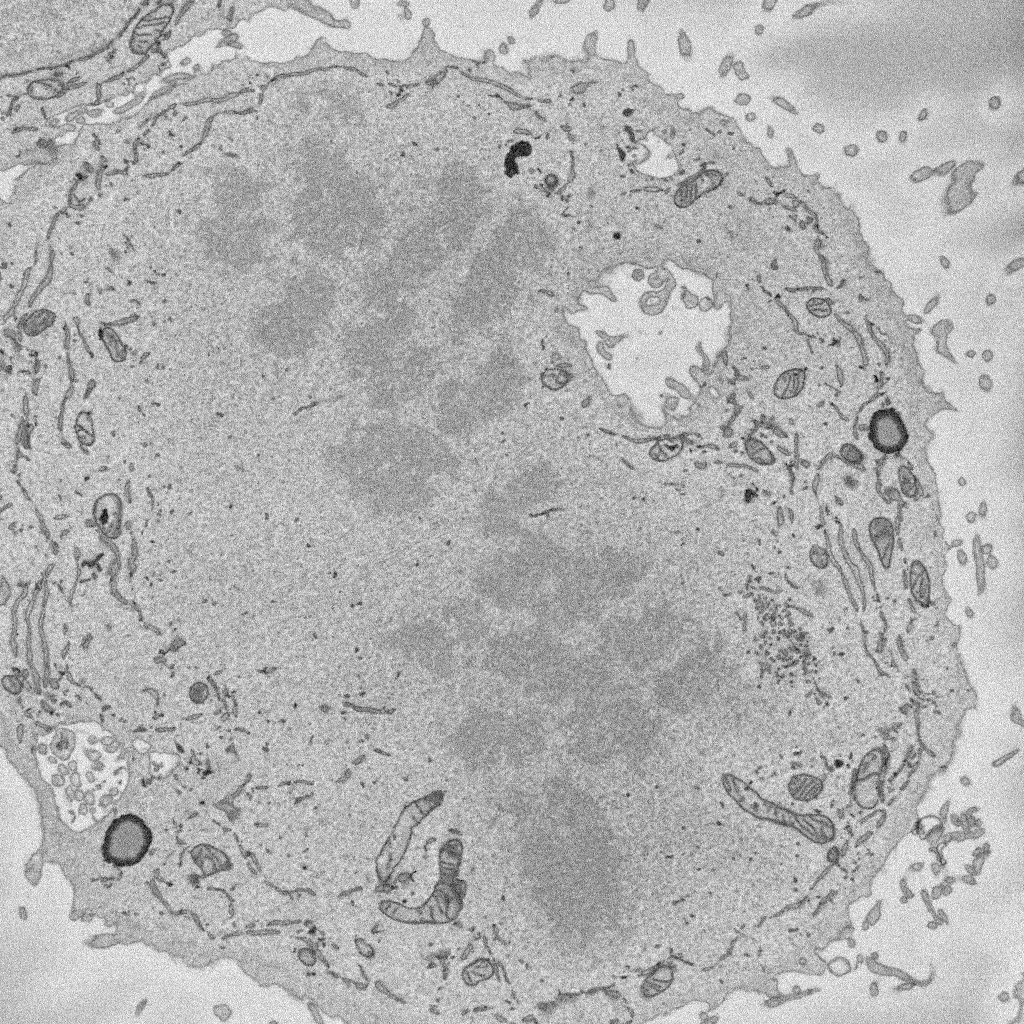} \\
\hline
\end{tabular}
\end{table}
\begin{table}[ht]
\centering
\begin{tabular}{|p{0.6\linewidth}|p{0.3\linewidth}|}
\hline
\multicolumn{2}{|c|}{\textbf{QUESTION TYPE: Domain}} \\ 
\hline
\raggedright \textbf{Question:} What is the most likely field of study this micrograph would be used for? \\
\textbf{Options:} \\
A. ophthalmology\newline 
\textcolor{orange}{B. biology}\newline 
C. radiology\newline  
D. pathology\newline  
E. cytology\newline  
F. none of the above\newline  &
\vspace{3em}\includegraphics[width=\linewidth]{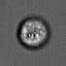} \\
\hline
\end{tabular}
\end{table}

\begin{table}[ht]
\centering
\begin{tabular}{|p{0.6\linewidth}|p{0.3\linewidth}|}
\hline
\multicolumn{2}{|c|}{\textbf{QUESTION TYPE: Domain}} \\ 
\hline
\raggedright \textbf{Question:} What is the most likely field of study this micrograph would be used for? \\
\textbf{Options:} \\
A. cytology\newline  
B. radiology\newline  
C. ophthalmology\newline  
D. pathology\newline 
\textcolor{orange}{E. biology}\newline 
F. none of the above\newline  &
\vspace{3em}\includegraphics[width=\linewidth]{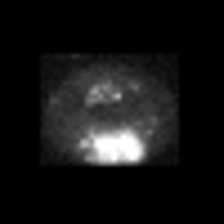} \\
\hline
\end{tabular}
\end{table}

\begin{table}[ht]
\centering
\begin{tabular}{|p{0.6\linewidth}|p{0.3\linewidth}|}
\hline
\multicolumn{2}{|c|}{\textbf{QUESTION TYPE: Subdomain}} \\ 
\hline
\raggedright \textbf{Question:} What is the most likely subfield of study this micrograph would be used for? \\
\textbf{Options:} \\
A. fine needle aspiration cytology\newline 
\textcolor{orange}{B. gynecologic cytology}\newline 
C. non-gynecologic cytology\newline  
D. none of the above\newline  &
\vspace{3em}\includegraphics[width=\linewidth]{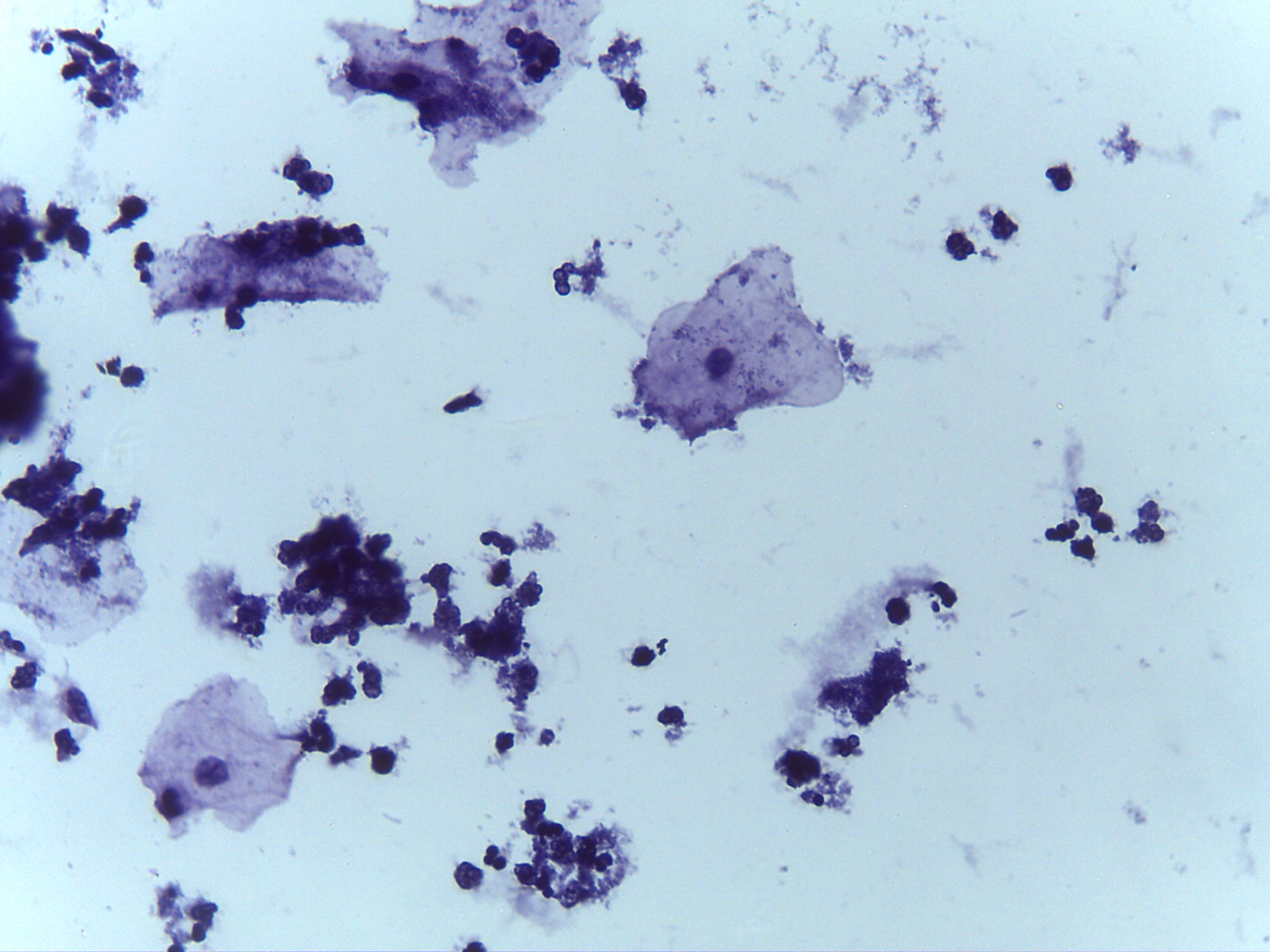} \\
\hline
\end{tabular}
\end{table}

\begin{table}[ht]
\centering
\begin{tabular}{|p{0.6\linewidth}|p{0.3\linewidth}|}
\hline
\multicolumn{2}{|c|}{\textbf{QUESTION TYPE: Subdomain}} \\ 
\hline
\raggedright \textbf{Question:} What is the most likely subfield of study this micrograph would be used for? \\
\textbf{Options:} \\
A. biotechnology.\newline 
\textcolor{orange}{B. botany.}\newline 
C.  neurobiology.\newline  
D.  physiology.\newline  
E. biology.\newline  
F. none of the above\newline  &
\vspace{3em}\includegraphics[width=\linewidth]{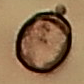} \\
\hline
\end{tabular}
\end{table}

\begin{table}[ht]
\centering
\begin{tabular}{|p{0.6\linewidth}|p{0.3\linewidth}|}
\hline
\multicolumn{2}{|c|}{\textbf{QUESTION TYPE: Stain}} \\ 
\hline
\raggedright \textbf{Question:} What is the most likely technique used to stain this micrograph? \\
\textbf{Options:} \\
A. Papanicolaou\newline  
B. H\&E\newline  
C. PAS\newline  
D. IHC(HDab)\newline 
\textcolor{orange}{E. Giemsa}\newline 
F. none of the above\newline  &
\vspace{3em}\includegraphics[width=\linewidth]{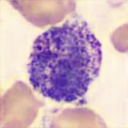} \\
\hline
\end{tabular}
\end{table}

\begin{table}[ht]
\centering
\begin{tabular}{|p{0.6\linewidth}|p{0.3\linewidth}|}
\hline
\multicolumn{2}{|c|}{\textbf{QUESTION TYPE: Stain}} \\ 
\hline
\raggedright \textbf{Question:} What is the most likely technique used to stain this micrograph? \\
\textbf{Options:} \\
A. Giemsa\newline  
B. Ziehl-Neelsen\newline  
C. IHC(Red)\newline 
\textcolor{orange}{D. H\&E}\newline 
E. Alcian Blue\newline  
F. none of the above\newline   &
\vspace{3em}\includegraphics[width=\linewidth]{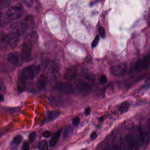} \\
\hline
\end{tabular}
\end{table}

\FloatBarrier
\FloatBarrier
\section{\dataset\ Perception (Fine-Grained) Closed VQA Data Samples}
\label{perception_fine_grained_data_examples}

\begin{table}[ht]
\centering
\begin{tabular}{|p{0.6\linewidth}|p{0.3\linewidth}|}
\hline
\multicolumn{2}{|c|}{\textbf{TASK:  White blood cell classification}} \\ 
\hline
\raggedright \textbf{Prompt:} A Giemsa-stained light micrograph displaying human peripheral blood cells. As a blood cell recognition system, identify the correct cell type: \\
\textbf{Options:} \\
\textcolor{orange}{A. eosinophil}\newline B. neutrophil\newline  C. immature granulocyte\newline  D. platelet\newline  E. erythroblast\newline  F. none of the above\newline &
\vspace{3em}\includegraphics[width=\linewidth]{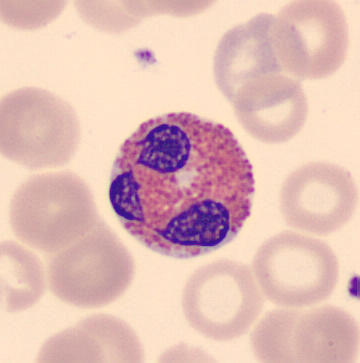} \\
\hline
\end{tabular}
\end{table}

\begin{table}[ht]
\centering
\begin{tabular}{|p{0.6\linewidth}|p{0.3\linewidth}|}
\hline
\multicolumn{2}{|c|}{\textbf{TASK: Classification of cell contour irregularity phenotypes in synthetic images}} \\ 
\hline
\raggedright \textbf{Prompt:} A synthetic fluorescence micrograph is displayed. What is the most likely description for the cell's contour irregularities? \\
\textbf{Options:} \\
A. irregular\newline \textcolor{orange}{B. intermediate}\newline C. smooth\newline D. none of the above\newline &
\vspace{2em}\includegraphics[width=\linewidth]{tables/pf1.png} \\
\hline
\end{tabular}
\end{table}

\begin{table}[ht]
\centering
\begin{tabular}{|p{0.6\linewidth}|p{0.3\linewidth}|}
\hline
\multicolumn{2}{|c|}{\textbf{TASK: Classification of cell contour irregularity phenotypes in synthetic images}} \\ 
\hline
\raggedright \textbf{Prompt:} An electron micrograph is shown. Based on the image, what is the most likely structure on the field of view? \\
\textbf{Options:} \\
A. a Zebrafish retina\newline B. Leishmania haptomonad\newline C. a HeLa cell in metaphase\newline \textcolor{orange}{D. Cardiac muscle}\newline E. a Tobacco leaf chloroplast\newline F. none of the above\newline &
\vspace{2em}\includegraphics[width=\linewidth]{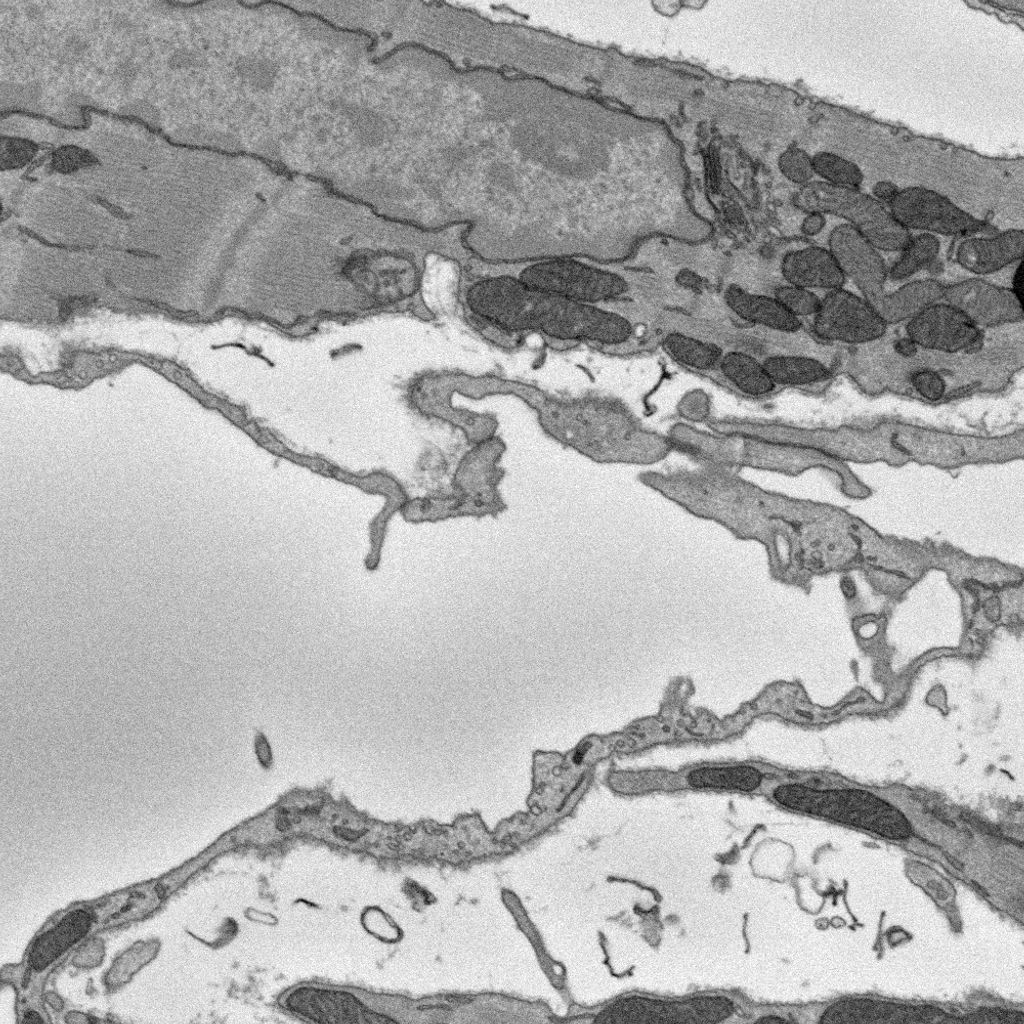} \\
\hline
\end{tabular}
\end{table}

\begin{table}[ht]
\centering
\begin{tabular}{|p{0.6\linewidth}|p{0.3\linewidth}|}
\hline
\multicolumn{2}{|c|}{\textbf{TASK: Classification of cell cycle phase}} \\ 
\hline
\raggedright \textbf{Prompt:} A brightfield micrograph of Jurkat cells acquired using flow cytometry (single cell). Based on the micrograph, what is the most likely cell phase? \\
\textbf{Options:} \\
A. interphase (G2)\newline \textcolor{orange}{B. interphase (G1) phase}\newline C. Telophase\newline D. Anaphase\newline E. Synthesis\newline F. none of the above\newline &
\vspace{2em}\includegraphics[width=\linewidth]{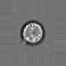} \\
\hline
\end{tabular}
\end{table}

\begin{table}[ht]
\centering
\begin{tabular}{|p{0.6\linewidth}|p{0.3\linewidth}|}
\hline
\multicolumn{2}{|c|}{\textbf{TASK: Classification of cell cycle stages in live cell imaging data}} \\ 
\hline
\raggedright \textbf{Prompt:} A fluorescence micrograph of Hela Kyoto cells stably expressing GalT-eGFP to label the Golgi apparatus. Based on the image what is the most likely the Golgi apparatus morphology? \\
\textbf{Options:} \\
A. Anaphase\newline B. Diffuse\newline \textcolor{orange}{C. Interphase}\newline D. Golgi twin\newline E. Partly disassembled\newline F. none of the above\newline &
\vspace{2em}\includegraphics[width=\linewidth]{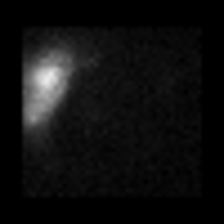} \\
\hline
\end{tabular}
\end{table}

\begin{table}[ht]
\centering
\begin{tabular}{|p{0.6\linewidth}|p{0.3\linewidth}|}
\hline
\multicolumn{2}{|c|}{\textbf{TASK: Classification of pre-cancerous and cervical cancer lesions in liquid-based cytology Pap smear images}} \\ 
\hline
\raggedright \textbf{Prompt:} Liquid-based cytology pap smear of human pre-cancerous or cervical cancer lesions. Based on the cytogram, what is the most likely finding? \\
\textbf{Options:} \\
A. Squamous cell carcinoma (SCC)\newline B. Low-grade (LSIL)\newline \textcolor{orange}{C. Negative (NILM)}\newline D. High-grade (HSIL)\newline E. none of the above\newline &
\vspace{2em}\includegraphics[width=\linewidth]{tables/pf5.png} \\
\hline
\end{tabular}
\end{table}

\begin{table}[ht]
\centering
\begin{tabular}{|p{0.6\linewidth}|p{0.3\linewidth}|}
\hline
\multicolumn{2}{|c|}{\textbf{TASK: Classification of normal and abnormal pollen grains}} \\ 
\hline
\raggedright \textbf{Prompt:} Basic fuchsin stained light micrograph of pollen grains. Based on the image, what is the most likely pollen class? \\
\textbf{Options:} \\
A. abnormal Corylus avellana\newline \textcolor{orange}{B. Non-pollen}\newline C. normal Alnus\newline D. normal Corylus avellana\newline E. none of the above\newline &
\vspace{2em}\includegraphics[width=\linewidth]{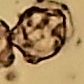} \\
\hline
\end{tabular}
\end{table}

\begin{table}[ht]
\centering
\begin{tabular}{|p{0.6\linewidth}|p{0.3\linewidth}|}
\hline
\multicolumn{2}{|c|}{\textbf{TASK: White blood cell classification}} \\ 
\hline
\raggedright \textbf{Prompt:} Synthetically generated Giemsa-stained light micrograph of human peripheral blood cell. As a blood cell recognition system, identify the correct cell type: \\
\textbf{Options:} \\
\textcolor{orange}{A. Neutrophil}\newline B. Basophil\newline C. Lymphocyte\newline D. Eosinophil\newline E. Monocyte\newline F. none of the above\newline &
\vspace{2em}\includegraphics[width=\linewidth]{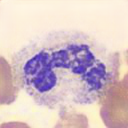} \\
\hline
\end{tabular}
\end{table}

\begin{table}[ht]
\centering
\begin{tabular}{|p{0.6\linewidth}|p{0.3\linewidth}|}
\hline
\multicolumn{2}{|c|}{\textbf{TASK: Texture classification in colorectal cancer}} \\ 
\hline
\raggedright \textbf{Prompt:} H\&E stained light micrograph of human colorectal tissue. Based on the image, what is the most likely texture class? \\
\textbf{Options:} \\
A. chronic heart failure\newline \textcolor{orange}{B. not chronic heart failure}\newline C. none of the above\newline &
\vspace{2em}\includegraphics[width=\linewidth]{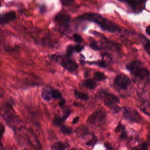} \\
\hline
\end{tabular}
\end{table}

\begin{table}[ht]
\centering
\begin{tabular}{|p{0.6\linewidth}|p{0.3\linewidth}|}
\hline
\multicolumn{2}{|c|}{\textbf{TASK: Classification of heart failure using cardiac histopathology images}} \\ 
\hline
\raggedright \textbf{Prompt:} H\&E stained light micrograph of human cardiac tissue. Based on the image, what is the most likely clinical chronic heart diagnosis? \\
\textbf{Options:} \\
A. chronic heart failure\newline \textcolor{orange}{B. not chronic heart failure}\newline C. none of the above\newline &
\vspace{2em}\includegraphics[width=\linewidth]{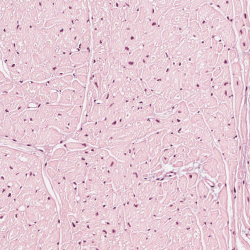} \\
\hline
\end{tabular}
\end{table}

\begin{table}[ht]
\centering
\begin{tabular}{|p{0.6\linewidth}|p{0.3\linewidth}|}
\hline
\multicolumn{2}{|c|}{\textbf{TASK: Amyloid beta pathology classification}} \\ 
\hline
\raggedright \textbf{Prompt:} IHC stained light micrograph of extracellular amyloid-beta deposition in the human brain tissue. Based on the image, what is the most likely amyloid beta morphology pattern? \\
\textbf{Options:} \\
A. Caa\newline B. Cored\newline \textcolor{orange}{C. Negative}\newline D. Diffuse\newline E. none of the above\newline &
\vspace{2em}\includegraphics[width=\linewidth]{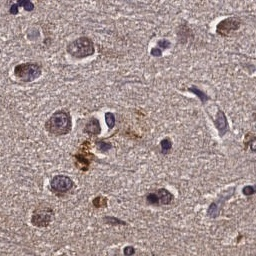} \\
\hline
\end{tabular}
\end{table}

\begin{table}[ht]
\centering
\begin{tabular}{|p{0.6\linewidth}|p{0.3\linewidth}|}
\hline
\multicolumn{2}{|c|}{\textbf{TASK: Classification of mitochondrial morphology in CryoET images}} \\ 
\hline
\raggedright \textbf{Prompt:} A cryo-electron tomography of mitochondria in neurons cultured in vitro is shown. Based on the image what is the most likely mitochondrial morphology? \\
\textbf{Options:} \\
A. abnormal mitochondrial morphology\newline \textcolor{orange}{B. normal mitochondrial morphology}\newline C. none of the above\newline &
\vspace{2em}\includegraphics[width=\linewidth]{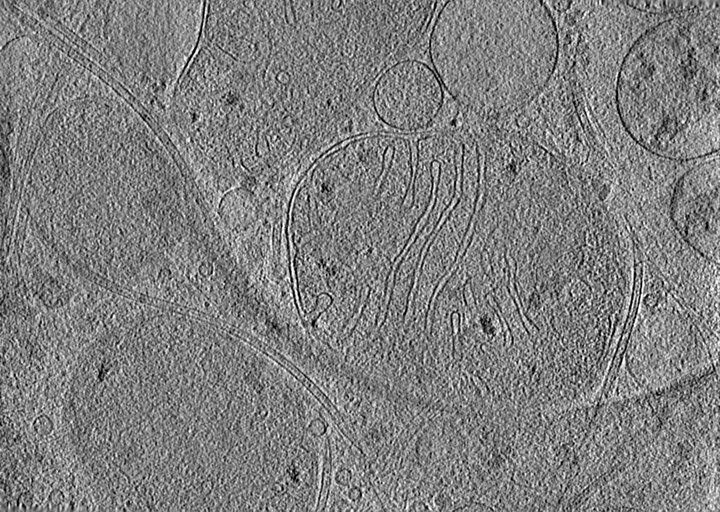} \\
\hline
\end{tabular}
\end{table}

\FloatBarrier
\FloatBarrier
\section{\dataset\ Cognition Closed VQA Data Samples}
\label{cognition_data_examples}

\begin{table}[ht]
\centering
\begin{tabular}{|p{0.6\linewidth}|p{0.3\linewidth}|}
\hline
\multicolumn{2}{|c|}{\textbf{Cognition Question Example}} \\ 
\hline
\raggedright \textbf{Question:} What type of cells are labelled green? \\
\textbf{Options:} \\
\textcolor{orange}{A. Cholinergic neurons} \newline B. Glial cells \newline C. Muscle fibers \newline D. Epithelial cells \newline F. Fibroblasts \newline G. None of the above \newline   &
\vspace{3em}\includegraphics[width=\linewidth]{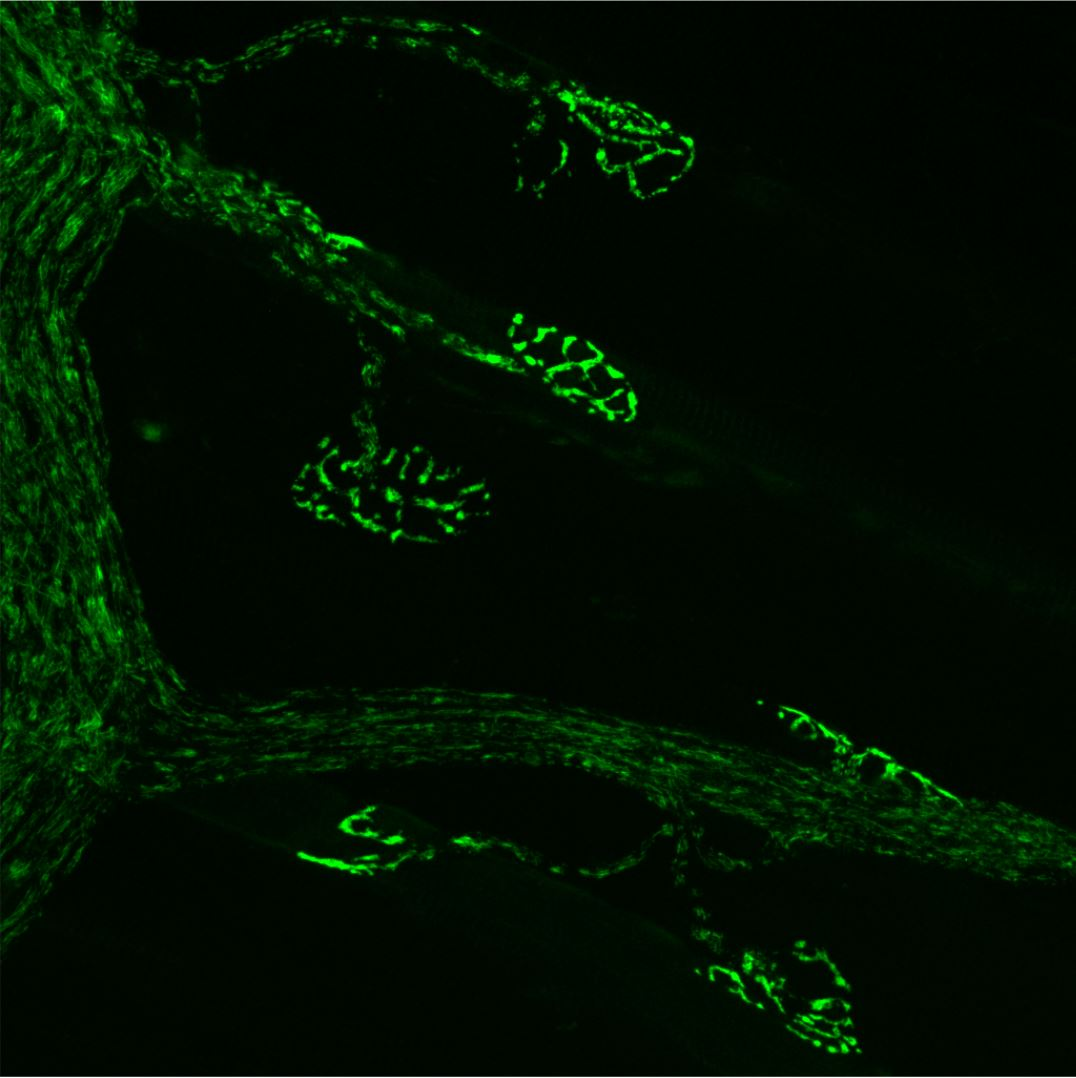} \\
\hline
\end{tabular}
\end{table}
\begin{table}[ht]
\centering
\begin{tabular}{|p{0.6\linewidth}|p{0.3\linewidth}|}
\hline
\multicolumn{2}{|c|}{\textbf{Cognition Question Example}} \\ 
\hline
\raggedright \textbf{Question:} In the provided fluorescence microscopy image of a section of a human pancreas stained with DAPI (dark blue), anti-insulin (light blue), anti-glucagon (red), and anti-somatostatin (green), are there more green-labeled or red-labeled features visible? \\
\textbf{Options:} \\
A. There are more green-labeled features (anti-somatostatin). \newline \textcolor{orange}{B. There are more red-labeled features (anti-glucagon).} \newline C. The number of green-labeled features (anti-somatostatin) and red-labeled features (anti-glucagon) are equal. \newline D. There are no green-labeled features (anti-somatostatin) visible. \newline E. There are no red-labeled features (anti-glucagon) visible. \newline F. None of the above \newline   &
\vspace{3em}\includegraphics[width=\linewidth]{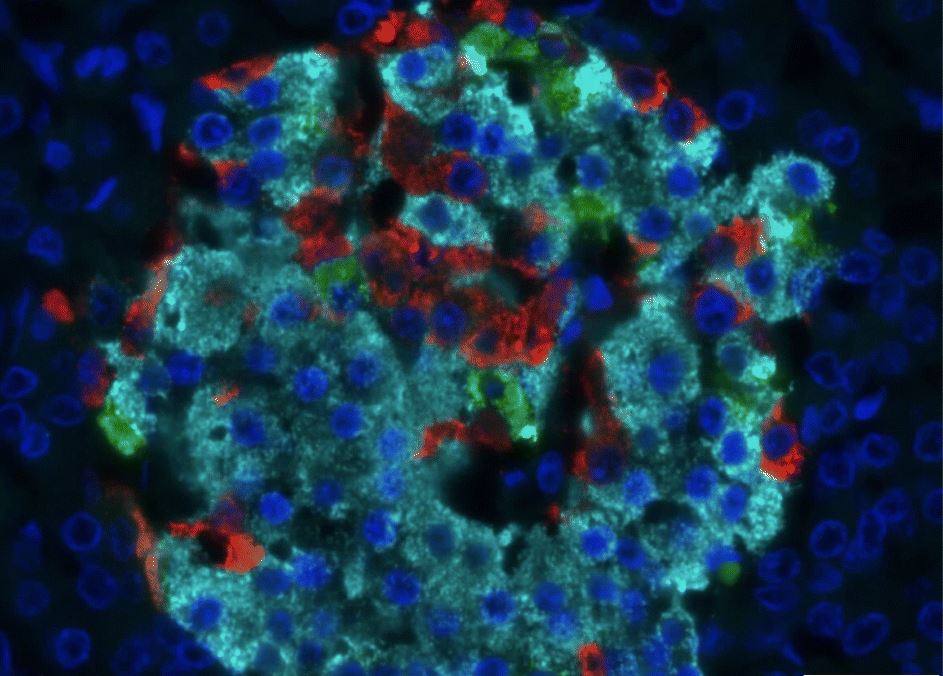} \\
\hline
\end{tabular}
\end{table}

\begin{table}[ht]
\centering
\begin{tabular}{|p{0.6\linewidth}|p{0.3\linewidth}|}
\hline
\multicolumn{2}{|c|}{\textbf{Cognition Question Example}} \\ 
\hline
\raggedright \textbf{Question:} What are the dark structures in the image? How does it relate to the biological process demonstrated in the image? \\
\textbf{Options:} \\
A. The dark structures are mitochondria, involved in energy production during cell metabolism. \newline B. The dark structures are lysosomes, which digest cellular waste as the cell divides. \newline C. The dark structures are chloroplasts, which harvest light energy during photosynthesis in plant cells. \newline D. The dark structures are vesicles, which transport materials to the cleavage furrow during cytokinesis. \newline \textcolor{orange}{E. The dark structures are homologous chromosomes, separated into each daughter cell during mitosis.} \newline F. None of the above \newline   &
\vspace{3em}\includegraphics[width=\linewidth]{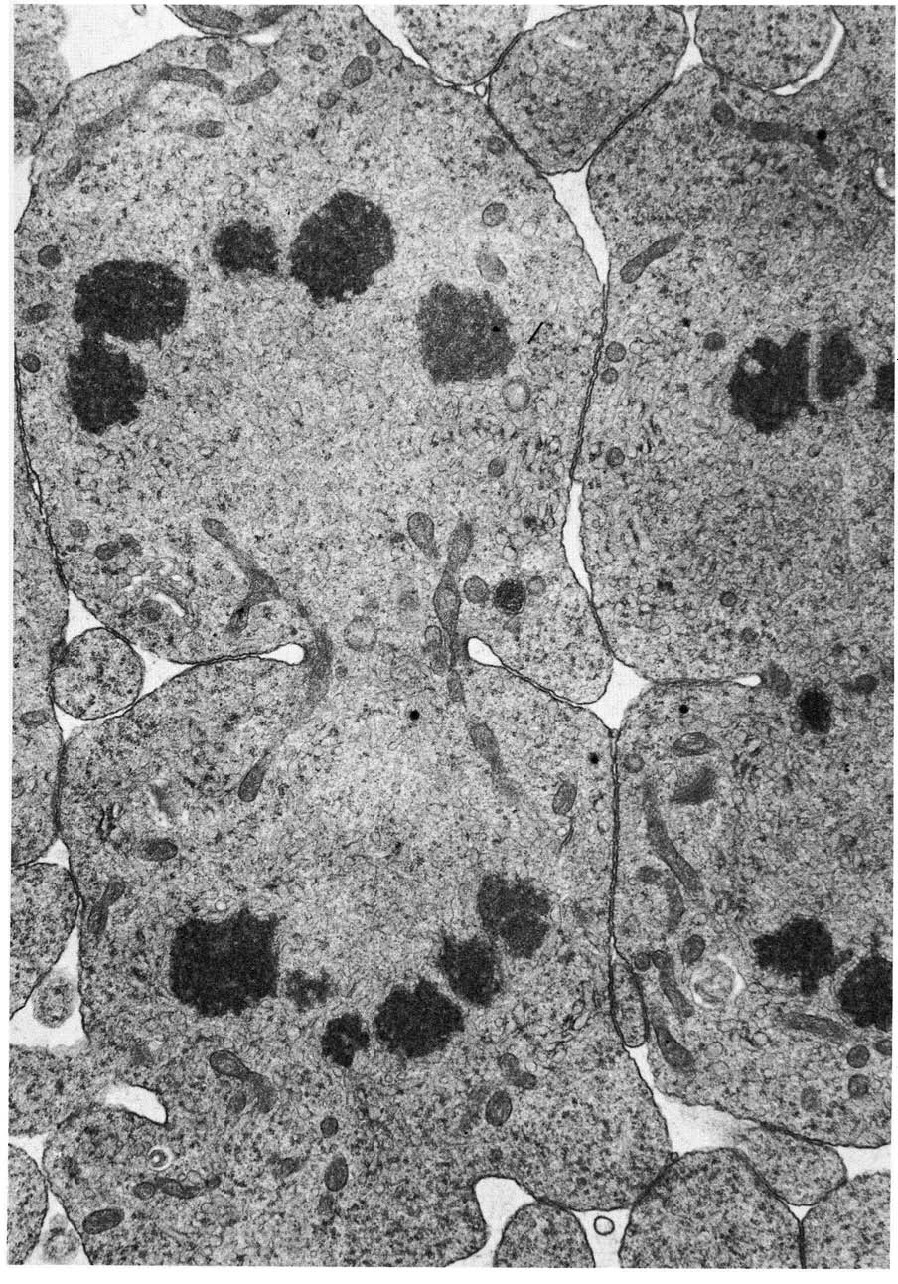} \\
\hline
\end{tabular}
\end{table}

\begin{table}[ht]
\centering
\begin{tabular}{|p{0.6\linewidth}|p{0.3\linewidth}|}
\hline
\multicolumn{2}{|c|}{\textbf{Cognition Question Example}} \\ 
\hline
\raggedright \textbf{Question:} What is the most likely structure labeled by the bright dots along the neurons? \\
\textbf{Options:} \\
A. Synapse parts \newline B. N-cadherin complexes \newline C. V-glut 1 and 2 transporters \newline D. Postsynaptic NMDA receptors \newline \textcolor{orange}{E. Excitatory synapses} \newline F. None of the above \newline   &
\vspace{3em}\includegraphics[width=\linewidth]{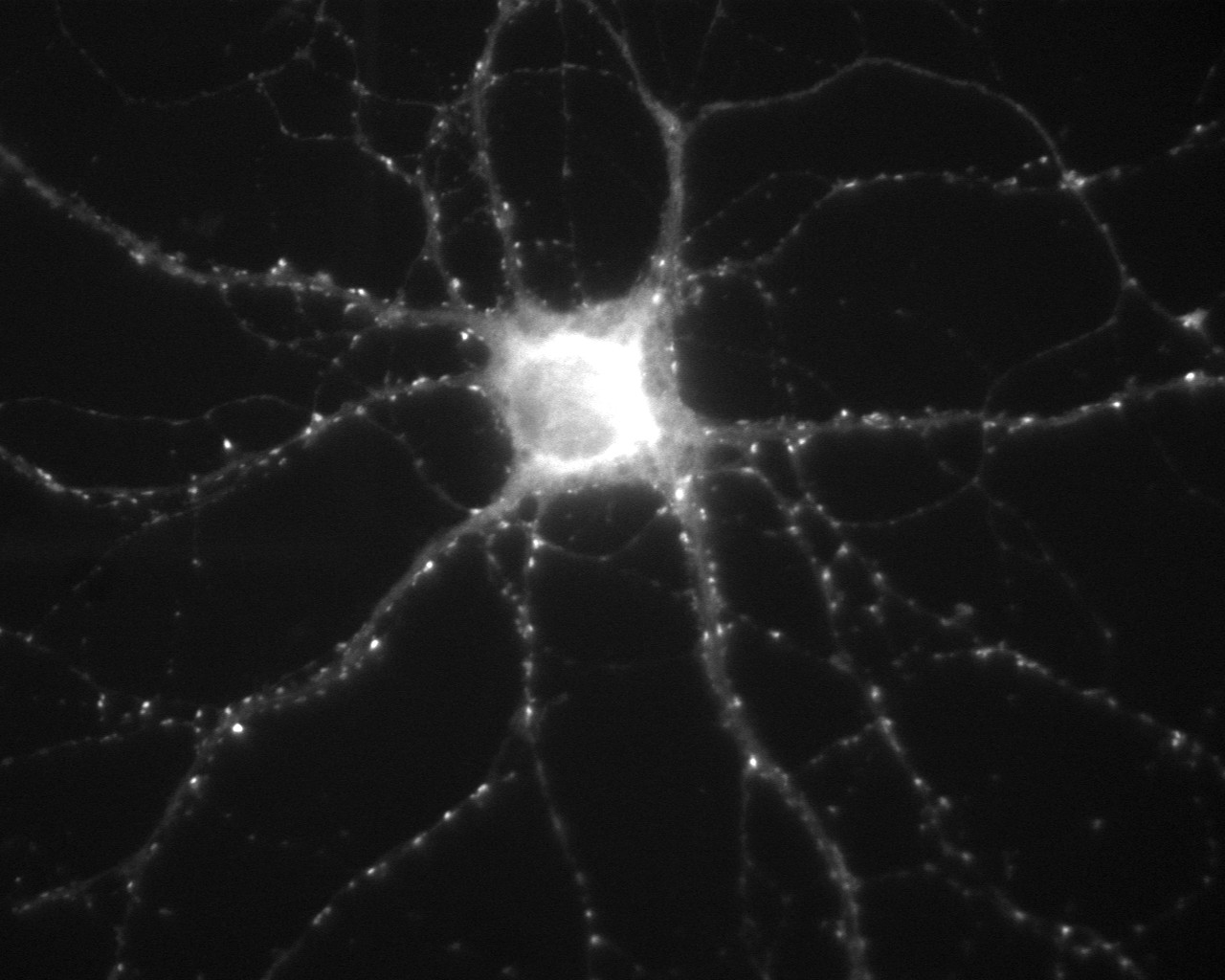} \\
\hline
\end{tabular}
\end{table}

\begin{table}[ht]
\centering
\begin{tabular}{|p{0.6\linewidth}|p{0.3\linewidth}|}
\hline
\multicolumn{2}{|c|}{\textbf{Cognition Question Example}} \\ 
\hline
\raggedright \textbf{Question:} What type of imaging is this? \\
\textbf{Options:} \\
\textcolor{orange}{A. This is an image of HT55 cancer cells and T cells.} \newline B. This is an image of neuronal cells. \newline C. This is an image of bacterial colonies. \newline D. This is an image of red blood cells. \newline E. This is an image of muscle tissue. \newline F. None of the above. \newline   &
\vspace{3em}\includegraphics[width=\linewidth]{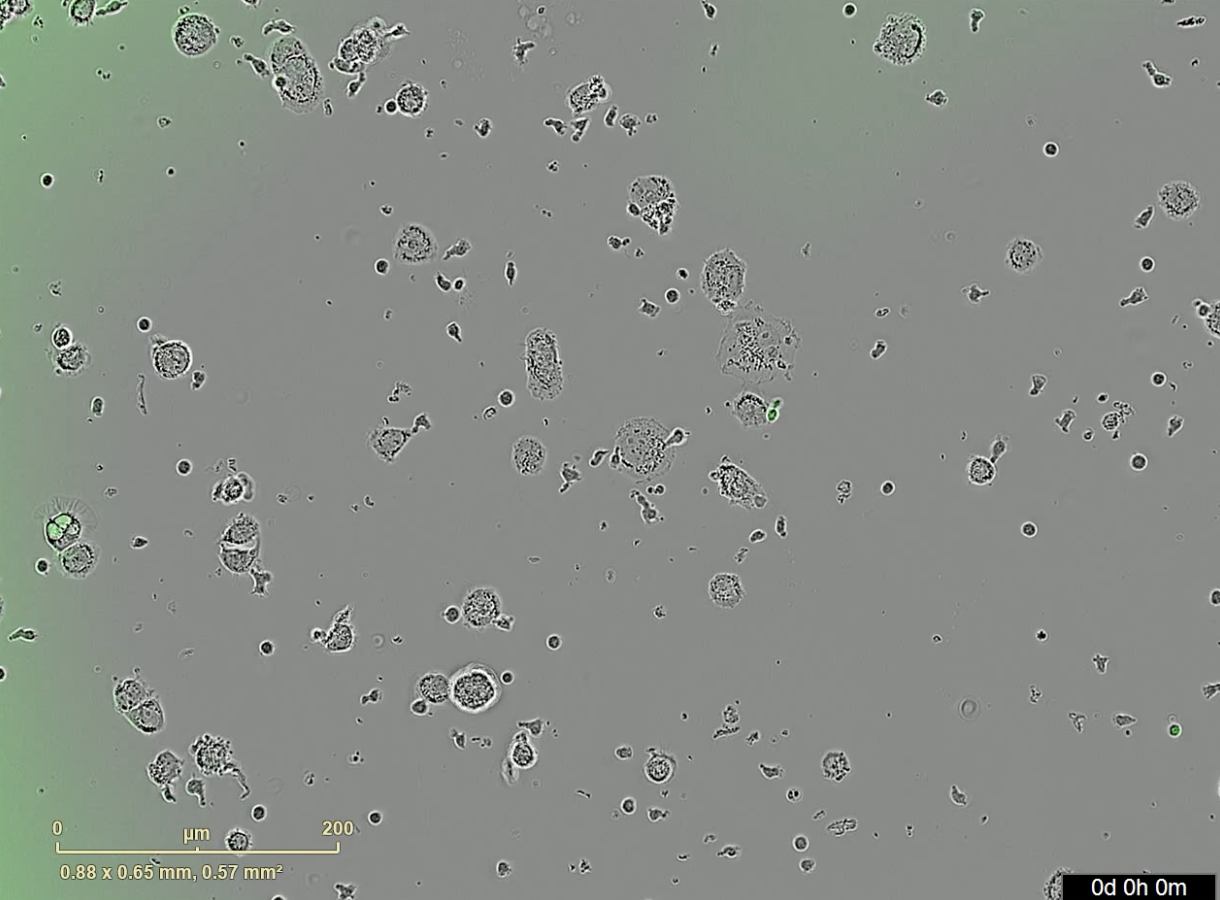} \\
\hline
\end{tabular}
\end{table}

\begin{table}[ht]
\centering
\begin{tabular}{|p{0.6\linewidth}|p{0.3\linewidth}|}
\hline
\multicolumn{2}{|c|}{\textbf{Cognition Question Example}} \\ 
\hline
\raggedright \textbf{Question:} What percentage of this tumor is composed of calcium? \\
\textbf{Options:} \\
A. 5\% \newline B. 10\% \newline C. 20\% \newline D. 15\% \newline \textcolor{orange}{E. 1\%} \newline F. None of the above \newline   &
\vspace{3em}\includegraphics[width=\linewidth]{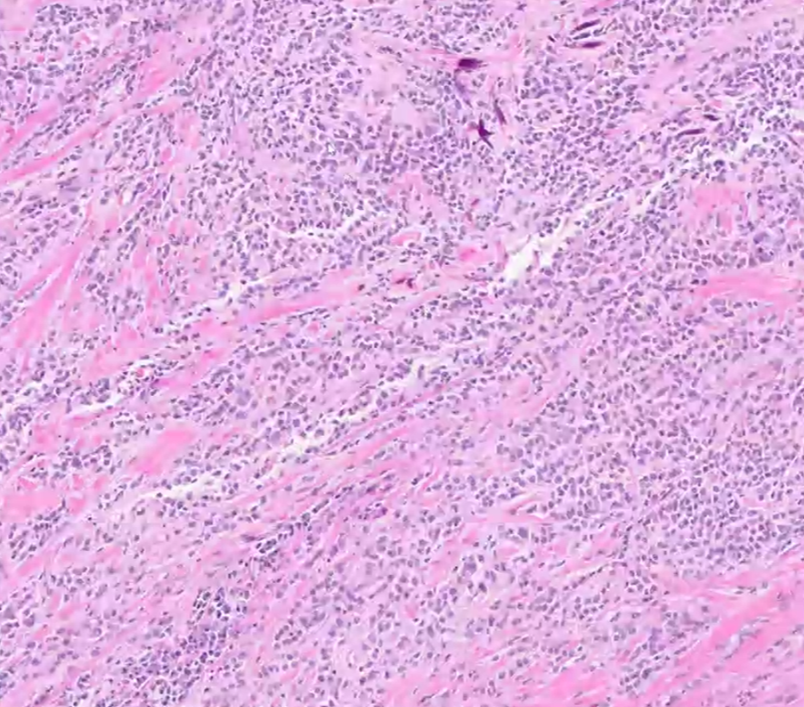} \\
\hline
\end{tabular}
\end{table}

\begin{table}[ht]
\centering
\begin{tabular}{|p{0.6\linewidth}|p{0.3\linewidth}|}
\hline
\multicolumn{2}{|c|}{\textbf{Cognition Question Example}} \\ 
\hline
\raggedright \textbf{Question:} What are the vesicular structures seen at the left and bottom borders? What is their function? \\
\textbf{Options:} \\
A. Early endosomes which are involved in sorting of endocytosed material. \newline B. Electron opaque cross-sectioned kinetodesmal fibers for structural support. \newline C. Granulo-fibrillar material involved in cellular structure. \newline \textcolor{orange}{D. Mitochondria which provide energy to the cell.} \newline E. Axosome which gives rise to microtubules. \newline F. None of the above. \newline   &
\vspace{3em}\includegraphics[width=\linewidth]{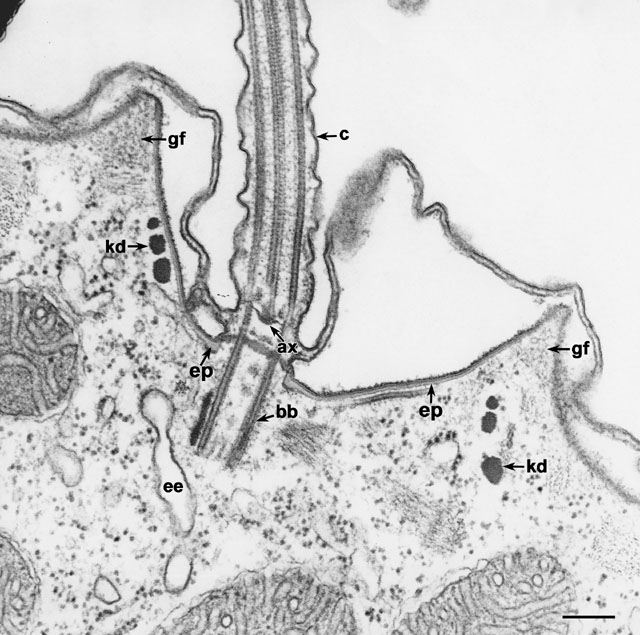} \\
\hline
\end{tabular}
\end{table}

\begin{table}[ht]
\centering
\begin{tabular}{|p{0.6\linewidth}|p{0.3\linewidth}|}
\hline
\multicolumn{2}{|c|}{\textbf{Cognition Question Example}} \\ 
\hline
\raggedright \textbf{Question:} Are synapses visible in this image? \\
\textbf{Options:} \\
A. Yes, synapses are clearly visible as distinct structures. \newline B. No, synapses are not visible as the image lacks specific synaptic markers. \newline C. Yes, synapses are visible where dendrites and axons come into close proximity. \newline \textcolor{orange}{D. No, synapses cannot be identified due to the resolution limitation of fluorescence microscopy.} \newline E. Yes, the orange and gray colored areas clearly show synapses. \newline F. None of the above. \newline   &
\vspace{3em}\includegraphics[width=\linewidth]{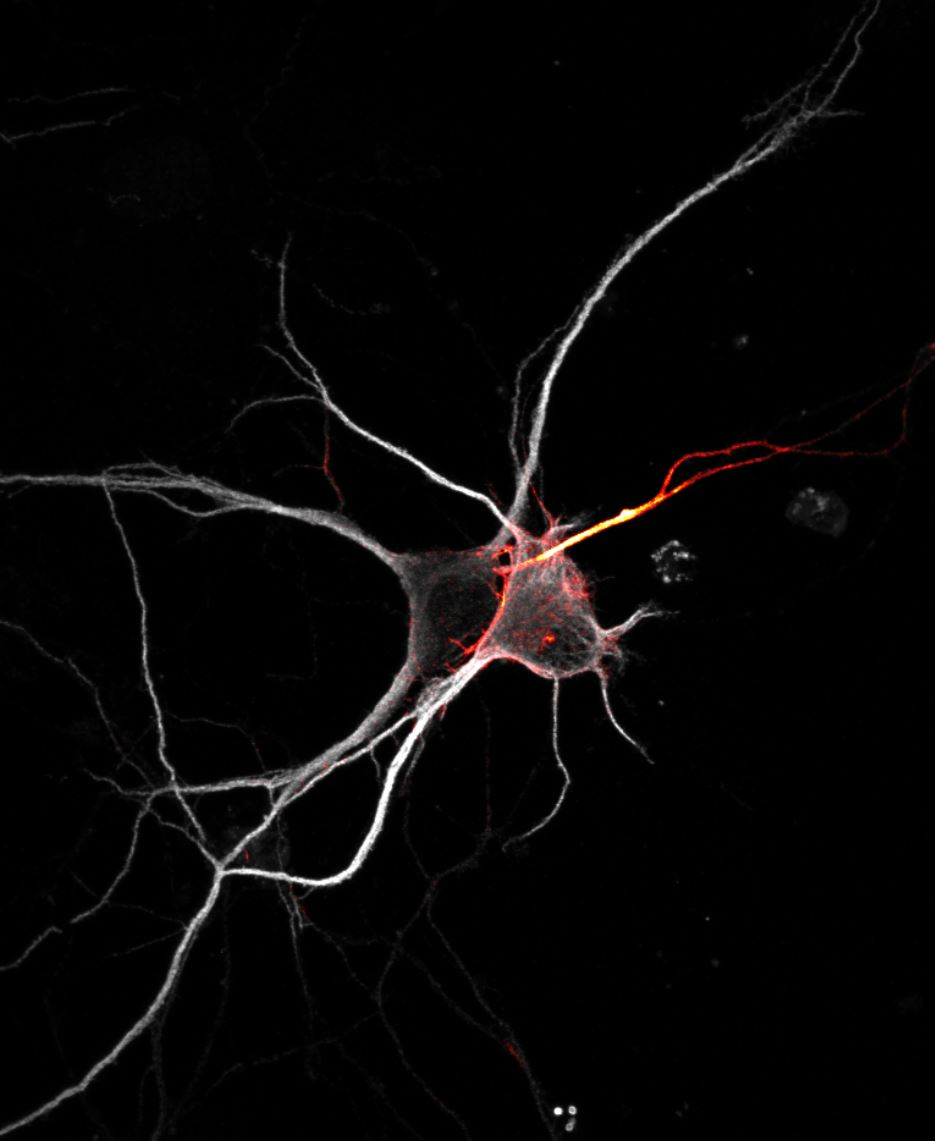} \\
\hline
\end{tabular}
\end{table}

\begin{table}[ht]
\centering
\begin{tabular}{|p{0.6\linewidth}|p{0.3\linewidth}|}
\hline
\multicolumn{2}{|c|}{\textbf{Cognition Question Example}} \\ 
\hline
\raggedright \textbf{Question:} Describe the spatial relationship between organelles in the cell. \\
\textbf{Options:} \\
A. Lysosomes are in two populations, one at the cell center and one at the periphery. \newline B. The Golgi apparatus surrounds the mitochondria while the ER is dispersed in the cytosol. \newline C. ER and lipid droplets cluster together at the cell periphery. \newline D. Mitochondria are positioned mainly in the cell periphery, while lysosomes are located centrally. \newline E. Lysosomes are located close to the nucleus while peroxisomes are scattered throughout the cytoplasm. \newline F. None of the above. \newline   &
\vspace{3em}\includegraphics[width=\linewidth]{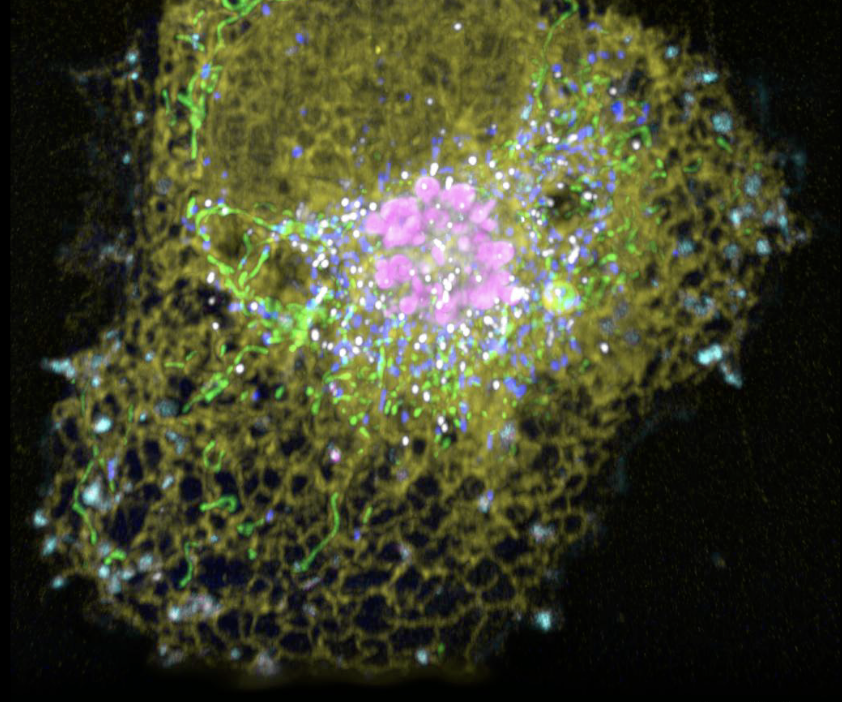} \\
\hline
\end{tabular}
\end{table}

\begin{table}[ht]
\centering
\begin{tabular}{|p{0.6\linewidth}|p{0.3\linewidth}|}
\hline
\multicolumn{2}{|c|}{\textbf{Cognition Question Example}} \\ 
\hline
\raggedright \textbf{Question:} Count the number of red cells in this image two days and four hours after initial co-culture in a 96 well plate. \\
\textbf{Options:} \\
A. 7500 cells \newline B. None of the above \newline C. 8500 cells \newline D. 9000 cells \newline E. 9500 cells \newline \textcolor{orange}{F. 8000 cells} \newline   &
\vspace{3em}\includegraphics[width=\linewidth]{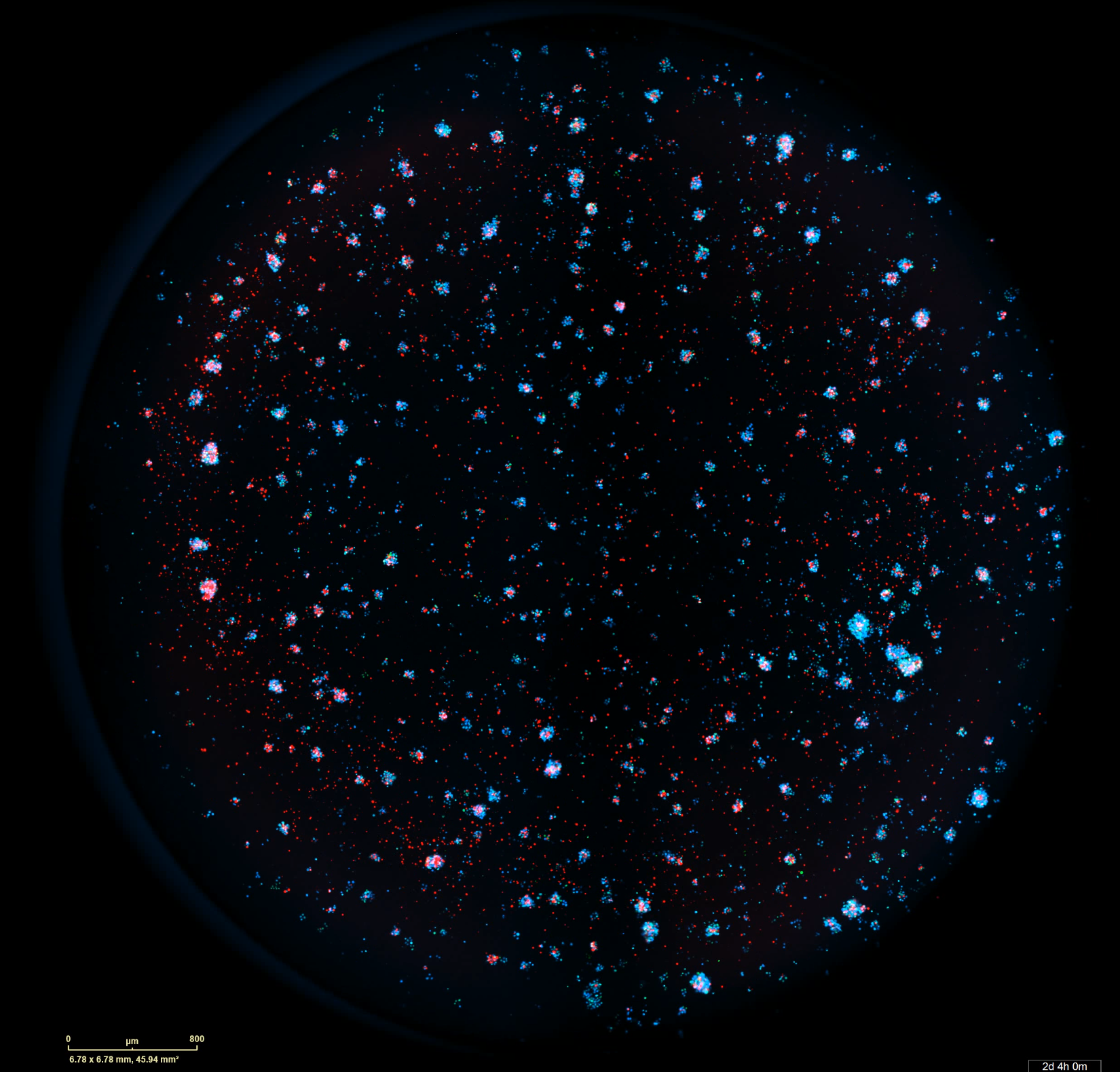} \\
\hline
\end{tabular}
\end{table}

\FloatBarrier

\end{document}